\documentclass[runningheads]{llncs}

\usepackage[mobile]{eccv}

\usepackage{eccvabbrv}

\usepackage{graphicx}
\usepackage{booktabs}
\usepackage{placeins}

\usepackage[accsupp]{axessibility}  % Improves PDF readability for those with disabilities.

\usepackage{hyperref}

\usepackage{orcidlink}
\usepackage{makecell}
\usepackage{multirow}
\usepackage{pifont}
\newcommand{\cmark}{\ding{51}}  % ✅
\newcommand{\xmark}{\ding{55}}  % ❌
\usepackage{tcolorbox}
\tcbuselibrary{breakable}
\usepackage{wrapfig}
\usepackage[table]{xcolor} 
\definecolor{teaserred}{RGB}{180,10,56}

\definecolor{teaserblue}{RGB}{0,15,139}

\definecolor{editblue}{RGB}{16, 82, 186}

\definecolor{uclagold}{RGB}{254,180,167}

\definecolor{grayred}{RGB}{232,237,205}

\usepackage{color, colortbl}

\definecolor{TealBlue}{rgb}{1.0, 0.97, 0.8}

\definecolor{iccvblue}{RGB}{189,227,195}
\newtcolorbox{ttcolorbox}[1][]{
  colframe=iccvblue,
  colback=iccvblue!5!white,
  title=#1,                
  fonttitle=\bfseries\sffamily,
  breakable
}

\begin{document}

\title{MultiHaystack: Benchmarking Multimodal Retrieval and Reasoning over 40K Images, Videos, and Documents} 


% Second Author\inst{2,3}\orcidlink{1111-2222-3333-4444} \and
% Third Author\inst{3}\orcidlink{2222--3333-4444-5555}}
\author{Dannong Xu\inst{1}\textsuperscript{*} \and
Zhongyu Yang\inst{2}\textsuperscript{*} \and
Jun Chen\inst{3} \and Yingfang Yuan\inst{4} \and Ming Hu\inst{5} \and \\ Lei Sun\inst{1} \and Luc Van Gool\inst{1} \and Danda Pani Paudel\inst{1} \and 
Chun-Mei Feng\inst{6}\textsuperscript{†} }

\authorrunning{D.~Xu et al.}

\institute{INSAIT, Sofia University “St. Kliment Ohridski” \and
Lanzhou University \and
King Abdullah University of Science and Technology \and
Heriot-Watt University \and
Monash University \and
University College Dublin \\
(*) equal contribution ; (†) corresponding author\\
\email{dannong.xu@insait.ai, chunmei.feng@ucd.ie}}

\maketitle
\begin{abstract}

Multimodal large language models (MLLMs) achieve strong performance on benchmarks that evaluate text, image, or video understanding separately. However, these settings do not assess a critical real-world requirement, which involves retrieving relevant evidence from large, heterogeneous multimodal corpora prior to reasoning.
Most existing benchmarks restrict retrieval to small, single-modality candidate sets, substantially simplifying the search space and overstating end-to-end reliability.
To address this gap, we introduce \textbf{MultiHaystack}, the first benchmark designed to evaluate both retrieval and reasoning under large-scale, cross-modal conditions.
MultiHaystack comprises over 46{,}000 multimodal retrieval candidates across documents, images, and videos, along with 747 open yet verifiable questions. Each question is grounded in a unique validated evidence item within the retrieval pool, requiring evidence localization across modalities and fine-grained reasoning.
In our study, we find that models perform competitively when provided with the corresponding evidence, but their performance drops sharply when required to retrieve that evidence from the full corpus.
Additionally, even the strongest retriever, E5-V, achieves only 40.8\% Recall@1, while state-of-the-art MLLMs such as GPT-5 experience a significant drop in reasoning accuracy from 80.86\% when provided with the corresponding evidence to 51.4\% under top-5 retrieval.
These results indicate that multimodal retrieval over heterogeneous pools remains a primary bottleneck for MLLMs, positioning MultiHaystack as a valuable testbed that highlights underlying limitations obscured by small-scale evaluations and promotes retrieval-centric advances in multimodal systems. Our code and benchmark are available at \url{https://danielxu0208.github.io/MultiHaystack.github.io/}.
\end{abstract}

\begin{figure}[ht]
    \centering
    \includegraphics[width=\linewidth]{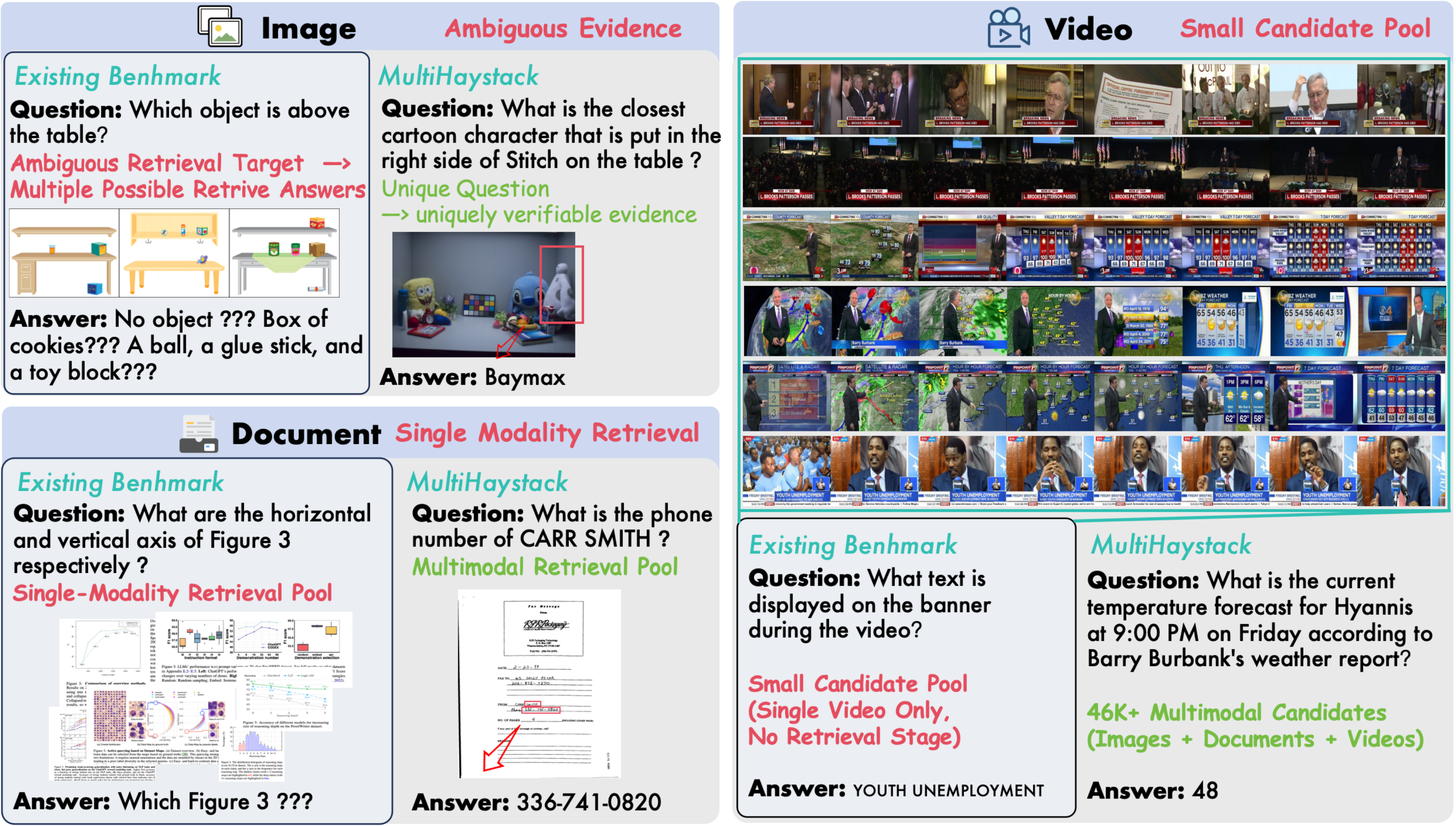}
\caption{\textbf{Comparison with existing visual question answering benchmarks.}
Existing benchmarks often suffer from three key limitations: 
(\textit{i}) ambiguous evidence that leads to multiple possible answers, 
(\textit{ii}) retrieval restricted to a single modality, and 
(\textit{iii}) small candidate pools (often limited to a single image, document, or video). 
In contrast, \textbf{MultiHaystack} provides questions grounded in uniquely verifiable evidence over a large-scale multimodal corpus of 46K+ items spanning documents, images, and videos, requiring both modality selection and fine-grained reasoning.}
    \label{fig:compare}
\end{figure}
\section{Introduction}
\label{sec:intro}
Multimodal large language models (MLLMs) \cite{chen2023minigpt, liu2025ola, zhang2025openmmreasoner} have driven substantial advances in AI, providing multimodal understanding and reasoning capabilities across a wide range of tasks \cite{mermaid,evermemos,Zhu2025InternVL3EA, inex, yang2025longvt, wikiautogen}. However, real-world applications of MLLMs typically require retrieval prior to reasoning to address practical challenges such as hallucination and the need for domain-specific knowledge. Meanwhile, these real-world scenarios often involve large-scale and inherently multimodal candidate pools. These considerations motivate research on developing MLLMs with integrated retrieval and reasoning capabilities over massive multimodal datasets. For example, consider a user uploading an image of a complex mechanical part and asking, ``\texttt{Which exact step in the video manual demonstrates the replacement of this component?}''. To answer this query, MLLMs must first retrieve the single relevant video segment from a large, multimodal instructional pool based on the visual query, and then perform fine-grained reasoning to pinpoint the answer.

To properly evaluate MLLMs on such retrieval--reasoning tasks, assessments must explicitly measure both stages. However, most existing benchmarks focus primarily on the reasoning phase and do not enable this decoupled, step-wise evaluation. Evaluating both stages provides sharper diagnostic insight into error sources and clarifies the critical interplay between retrieval quality and downstream reasoning performance.

Additionally, although recent retrieval-oriented datasets have made progress, they remain insufficient in three critical aspects. First, many benchmarks adopt an \textbf{unrealistic scale}. They include only hundreds or thousands of candidates, which makes the retrieval task relatively trivial and inflates reported accuracy~\cite{meng2025mmiu}. Second, \textbf{modality coverage} is often limited. Many datasets are restricted to a single modality and therefore fail to adequately evaluate cross-modal performance, particularly in settings that require retrieving and integrating evidence across text, images, and videos~\cite{chen2024document, wang2024mmneedle}. Third, question--evidence design is often \textbf{ambiguous} (as shown in \Cref{fig:compare}). Some benchmarks do not explicitly assign a unique retrieval target to each question, instead permitting vague answers linked to multiple possible targets. This ambiguity hinders reproducibility and obscures true model weaknesses~\cite{Mathew2020DocVQAAD}.

\begin{wrapfigure}[15]{r}{0.43\textwidth}
    \centering
\vspace{-8mm}    
\includegraphics[width=0.43\textwidth]{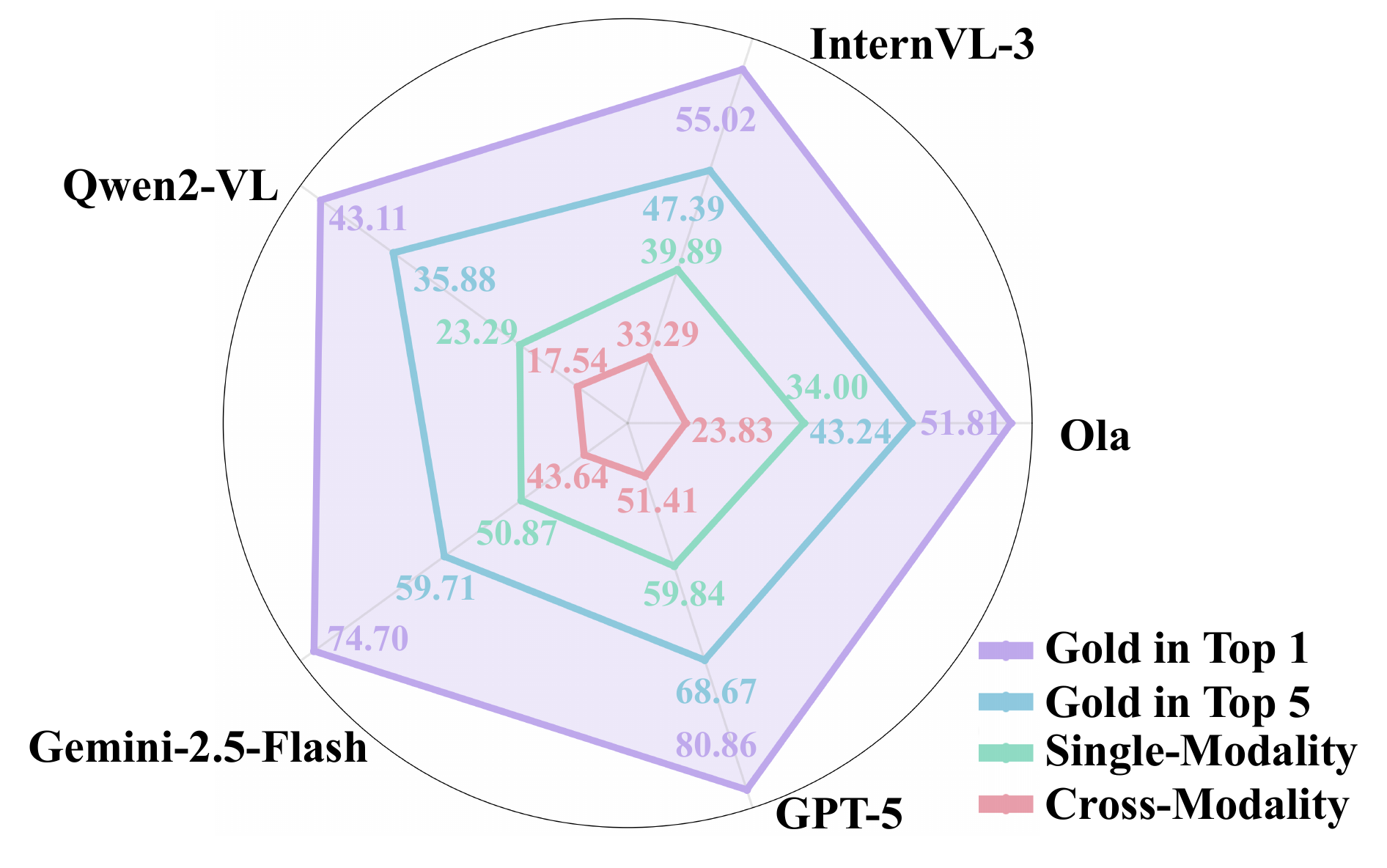}
    \caption{\textbf{Performance on MultiHaystack.} ``Gold in Top-1/5'' directly provides answer-containing files; ``Single-Modality'' and ``Cross-Modality'' require retrieval within one or across multiple modalities. }
    \label{fig:performance_compare}
\end{wrapfigure}

To address these gaps, we introduce MultiHaystack, the first large-scale benchmark designed for realistic cross-modal retrieval and reasoning. MultiHaystack comprises 46{,}260 documents, images, and videos in total, paired with 747 evidence-grounded questions. Crucially, each question is anchored to a unique retrieval target that supports a verifiable answer, thereby ensuring clarity and evaluation reproducibility. As illustrated in Table~\ref{tab:compare}, each question requires retrieving a single relevant item from a multimodal pool of up to 46K candidates, followed by fine-grained cross-modal reasoning.

To date, few MLLMs are natively equipped to handle end-to-end retrieval and reasoning at this massive scale. Our benchmark, therefore, provides a rigorous foundation for developing and diagnosing such models. To assess current MLLMs, we decompose the task and evaluate a range of state-of-the-art multimodal retrievers and reasoning models. The results reveal that reasoning performance remains high when the exact gold evidence is provided, but declines sharply when evidence must be retrieved from a large multimodal pool, particularly in cross-modal scenarios. For example, as shown in \Cref{fig:performance_compare}, GPT-5 experiences a substantial drop in reasoning accuracy from 80.86\% (when provided with the corresponding evidence) to 51.4\% under top-5 retrieval. Furthermore, even strong retrievers such as E5-V achieve a 72.42\% Recall@1 on a 1K pool, which drastically degrades to 40.83\% as the candidate pool expands to 46K. Together, these findings identify retrieval as a central bottleneck, demonstrating how small-scale, single-modality evaluations have masked this limitation.

\newpage
In summary, our contributions are as follows:

\begin{itemize}
\setlength\itemsep{0pt}
\setlength\parsep{0pt}
\setlength\topsep{0pt}
    
\item We introduce \textbf{MultiHaystack}, the first large-scale benchmark for cross-modal retrieval and reasoning, spanning 46K+ documents, images, and videos.  

\item We strictly ground each question in a single piece of evidence, where each query necessitates precise evidence retrieval across modalities, enabling fine-grained, step-wise evaluation.

\item We conduct comprehensive experiments on diverse MLLMs, exposing the severe performance degradation at scale and highlighting multimodal retrieval over heterogeneous pools as the primary frontier for MLLM reasoning.  
\end{itemize}

\begin{table*}[t]
\centering
\caption{\textbf{Comparison of Benchmarks.} 
MultiHaystack is a large-scale multimodal benchmark with 46K+ items, 
supporting multimodal retrieval, unique evidence, and six task types, 
while prior benchmarks remain limited.}
\vspace{-2mm}
\renewcommand{\arraystretch}{1.5}
\resizebox{\textwidth}{!}{
\begin{tabular}{lccccc}
\toprule
\textbf{Benchmark} & 
\textbf{Modality} & 
\makecell{\textbf{Retrieval Candidates}\\\textbf{per QA}} & 
\textbf{Data Types}  & 
\makecell{\textbf{Unique}\\\textbf{Evidence}} & 
\makecell{\textbf{Task}\\\textbf{Types}} \\
\midrule
WebQA \cite{webvqa2023}              & \includegraphics[height=1.2em]{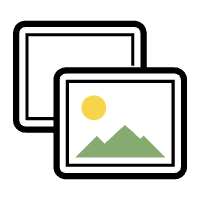}     & 1--5        & Web images                & \textcolor{green}{\cmark} & 1 \\
RetVQA \cite{retvqa}                 & \includegraphics[height=1.2em]{fig/image.png}     & 20--30      & Natural images           & \textcolor{green}{\cmark} & 2 \\
MM-NIAH \cite{wang2024needle}        & \includegraphics[height=1.2em]{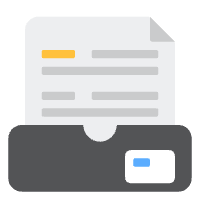} \includegraphics[height=1.2em]{fig/image.png}   & 10--70+     & Mixed text-image         & \textcolor{red}{\xmark}  & 3 \\
MMNeedle \cite{wang2024mmneedle}     & \includegraphics[height=1.2em]{fig/image.png}     & 10--160     & Image patch              & \textcolor{green}{\cmark}  & 1 \\
DocHaystack \cite{chen2024document}  & \includegraphics[height=1.2em]{fig/doc.png}        & 100--1000   & Document images           & \textcolor{green}{\cmark} & 2 \\
\textbf{MultiHaystack}               & \includegraphics[height=1.2em]{fig/doc.png} \includegraphics[height=1.2em]{fig/image.png} \includegraphics[height=1.2em]{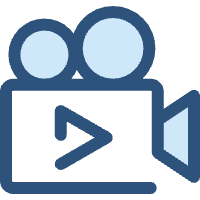} & $\mathbf{46K+}$ & \textbf{Multimedia items} & \textcolor{green}{\cmark} &  \textbf{6} \\
\bottomrule
\end{tabular}
}
\vspace{-5mm}
\label{tab:compare}
\end{table*}

\section{Related Work}

\noindent \textbf{Visual Question Answering (VQA) Benchmarks.} 
Early Visual Question Answering benchmarks evaluate perception over isolated images or documents~\cite{lin2015microsoftcococommonobjects,plummer2016flickr30kentitiescollectingregiontophrase,DBLP:journals/ijcv/GoyalKASBP19/VQA-Matter,evermembench,ging2024openendedvqabenchmarkingvisionlanguage,chang2025wearvqavisualquestionanswering,marino2019okvqavisualquestionanswering,piergiovanni2022videoquestionansweringiterative,Mathew2020DocVQAAD, XR,yang2025longvt,mangalam2023egoschemadiagnosticbenchmarklongform,yang2022learninganswervisualquestions,wu2024longvideobenchbenchmarklongcontextinterleaved}. 
Subsequent extensions incorporate external knowledge and broader modalities such as video and audio~\cite{schwenk2022okvqa,singh2019towards,Mathew2020DocVQAAD,liu2024mmbench,li2024mvbenchcomprehensivemultimodalvideo,li2025omnibenchfutureuniversalomnilanguage}. 
While advancing multimodal reasoning, these evaluations persistently assume a single, bounded context where the answer-containing content is already provided \cite{chen2023pretrainedvisionlanguagemodels,jiang2026pixelsfactspix2factbenchmarking}. 
Consequently, they excel at probing \textit{intra-instance} understanding but fundamentally bypass the critical preliminary step of open-corpus retrieval, which is essential for real-world reliability.

\noindent \textbf{Retrieval-Centric Benchmarks.} 
To address this gap, recent benchmarks adopt needle-in-a-haystack retrieval settings~\cite{webvqa2023,retvqa,chen2024document,wang2024needle,wang2024mmneedle}. 
However, they remain limited in several respects. 
First, candidate pools are often small, which underestimates retrieval difficulty and its impact on end-to-end performance~\cite{meng2025mmiu,luo2025globalretrievalaugmentedgeneration}. 
Second, many benchmarks focus on a single dominant modality, leaving cross-modal retrieval across documents, images, and videos less explored~\cite{chen2024document,wang2024mmneedle,peng2026unidocbenchunifiedbenchmarkdocumentcentric}. 
Third, loosely grounded question--evidence designs can lead to ambiguous or non-unique retrieval targets, making it difficult to separate retrieval errors from downstream reasoning failures~\cite{Mathew2020DocVQAAD,shen2025rightwayassessingdocument,ying2026retrievalinfusedreasoningsandboxbenchmark}. 

MultiHaystack addresses these limitations through three design choices. 
First, it scales the candidate pool to 46K+ multimodal items, introducing realistic retrieval difficulty that affects end-to-end performance. 
Second, it integrates documents, images, and videos within a unified benchmark, requiring models to perform cross-modal selection and integration. 
Third, it enforces uniquely verifiable evidence grounding, enabling retrieval errors to be separated from downstream reasoning failures. 
Collectively, these choices transform MultiHaystack into a high-fidelity diagnostic tool that complements prior benchmarks in evaluating the reliability of contemporary multimodal RAG systems.
\begin{figure}[t]
    \centering
    \includegraphics[width=\linewidth]{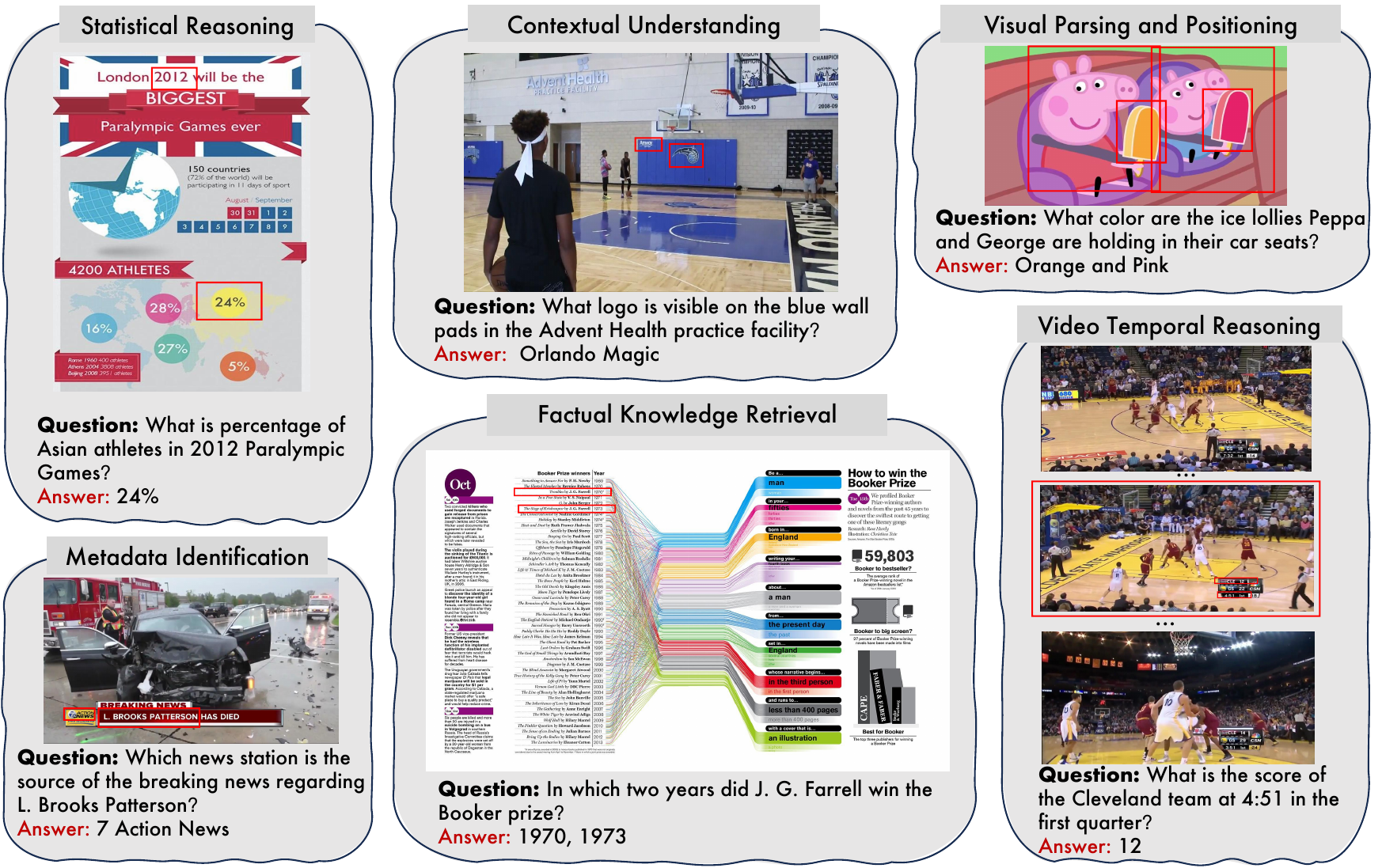}
    \caption{\textbf{Examples of six tasks in MultiHaystack:} Visual Parsing \& Positioning (spatial layouts), Contextual Understanding (embedded text), Video Temporal Reasoning (motion/order), Statistical Reasoning (charts/tables), Metadata Identification (affiliations/timestamps), and Factual Knowledge Retrieval (corpus-grounded facts).}
    \label{fig:task-examples}
    \vspace{-2mm}
\end{figure}

\section{MultiHaystack}
In this section, we introduce \textbf{MultiHaystack}, a large-scale benchmark for evaluating MLLMs on large-corpus heterogeneous cross-modal retrieval and grounded reasoning. 
Unlike instance-level VQA benchmarks, MultiHaystack requires open-domain evidence selection from a unified multimodal candidate pool.
The benchmark is built through four stages: data collection, question generation, multi-step filtering, and data enrichment, where hard negatives are added to scale the candidate pool and increase retrieval difficulty.
We formalize the heterogeneous corpus as 
$\mathcal{D} = \{d_1, \ldots, d_N\}$,
where each candidate item $d_i$ is an image, video, or document (see \cref{fig:data_pipeline}), jointly indexed in a shared retrieval space. 
For each question, exactly one $d_i \in \mathcal{D}$ provides the uniquely supporting evidence. 
This unique-evidence constraint ensures explicit grounding and unambiguous evaluation while preventing shortcut solutions.

\subsection{Task Definition} 
Given a question $q$ and a heterogeneous corpus $\mathcal{D}$, the model must retrieve the uniquely paired supporting evidence $d_i \in \mathcal{D}$ and generate the answer based on $d_i$. 
This retrieval-then-reasoning formulation requires both evidence selection and grounded inference to succeed. The step-wise design enables the separate evaluation of cross-modal retrieval and reasoning.

% Retrieval errors directly lead to incorrect answers, thereby jointly evaluating cross-modal retrieval and reasoning.

\subsection{Data Statistics}

\noindent\textbf{Comparison.} 
Most existing benchmarks either focus on a single modality or assume that relevant evidence is confined to a small pre-selected context (\cref{fig:compare}), thereby under-testing large-corpus evidence selection. 
In contrast, MultiHaystack operates over a heterogeneous corpus of documents, images, and videos, where each query is grounded in exactly one supporting item within the full candidate pool. 
This unique-evidence design requires models to first retrieve the correct cross-modal evidence and then perform reasoning conditioned on it. 
As summarized in Table~\ref{tab:compare}, MultiHaystack therefore jointly evaluates large-scale cross-modal retrieval and grounded reasoning in a unified setting.

\noindent\textbf{Task Distribution.} 
Conditioned on the retrieved evidence, we categorize questions into six task types that capture diverse reasoning demands in cross-modal settings (as shown in \cref{fig:task-examples}):

\begin{itemize}
\setlength\itemsep{0pt}
\setlength\parsep{0pt}
\setlength\topsep{0pt}

\item \textbf{Visual Parsing and Positioning (VPP)} 
requires precise spatial grounding of objects and their relative layout within images or frames.

\item \textbf{Contextual Understanding (CU)} 
requires integrating embedded visual text or symbols with the surrounding context for semantic interpretation.

\item \textbf{Video Temporal Reasoning (VTR)} 
requires modeling cross-frame dynamics to infer motion, temporal order, and state transitions.

\item \textbf{Statistical Reasoning (SR)} 
requires extracting and reasoning over quantitative patterns in structured visual data.

\item \textbf{Metadata Identification (MI)} 
requires identifying and grounding structured metadata such as affiliations and timestamps.

\item \textbf{Factual Knowledge Retrieval (FKR)} 
requires retrieving and synthesizing corpus-grounded evidence for factual answering.

\end{itemize}

Based on this design, MultiHaystack contains 33 visual parsing tasks, 30 contextual understanding tasks, 44 video temporal reasoning tasks, 321 statistical reasoning tasks, 285 metadata identification tasks, and 34 factual knowledge retrieval tasks.
Each question is derived from a controlled ``needle'' extracted from a haystack of documents, images, and videos. 
The needle provides the unique evidence required to answer the question, ensuring explicit semantic grounding and non-trivial retrieval difficulty. 
Unlike prior benchmarks with potentially ambiguous targets~\cite{chen2024document, wang2024mmneedle}, 
\textbf{MultiHaystack} constrains each question to be grounded in a single, specific piece of evidence.

\begin{figure*}[t]
    \centering
    \includegraphics[width=\linewidth]{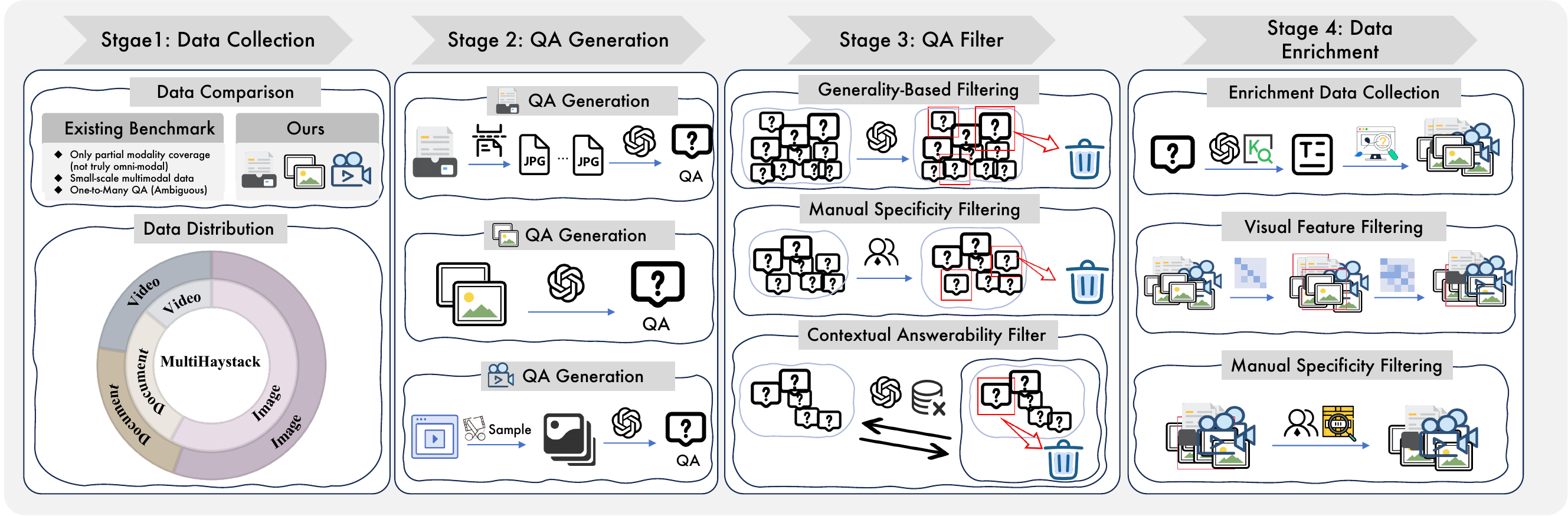}
    \caption{\textbf{Benchmark construction pipeline.} MultiHaystack is built in four stages: collecting diverse multimodal sources, generating specific QA pairs, filtering for unique and grounded answers, and enriching with data. This design ensures coverage across six tasks (\cref{fig:task-examples}) and overcomes the unimodal, small-scale, or ambiguous limitations of prior benchmarks.}
    \label{fig:data_pipeline}
\end{figure*}

\subsection{MultiHaystack Construction}

To ensure broad modality coverage and unambiguous, verifiable answers, MultiHaystack is constructed via a four-stage pipeline: data collection, question generation, filtering, and enrichment (\cref{fig:data_pipeline}).

\noindent\textbf{Stage~1: Data Collection.}  
We construct $\mathcal{D}$ by combining data from three modalities: images (DocHaystack~\cite{chen2024document}, MMIU~\cite{meng2025mmiu}, A-OKVQA~\cite{schwenk2022okvqa}), videos (VideoVista~\cite{li2024videovista}, MMBench-Video~\cite{fang2024mmbenchvideolongformmultishotbenchmark}, FineVideo~\cite{Farre2024FineVideo}, MVBench~\cite{li2024mvbenchcomprehensivemultimodalvideo}), and documents (MINT1T~\cite{mint1t}), spanning diverse cross-modal sources.

\noindent\textbf{Stage~2: Question generation.} 
Each item $d_i$ is normalized into an image-based representation: PDF pages are rendered page-wise, standalone images are used directly, and videos are uniformly sampled into eight frames.
Based on these images, GPT-4o generates a set of QA pairs $Q_i$. 
On average, each item contributes around 30 candidate questions, forming the raw QA pool $\mathcal{Q} = \bigcup_i Q_i$.

\noindent\textbf{Stage~3: Question filtering.}  
We apply a three-step process to ensure specificity and evidence grounding. 
(1) GPT-4o and Gemini-2.5-Flash remove ambiguous questions with multiple valid answers.  
(2) Manual review discards questions without explicit anchors (e.g., objects, locations, timestamps).
(3) A retrieval-independence test removes questions solvable without retrieving the supporting item.  
The resulting set $\mathcal{Q}^{\star}$ contains 747 questions; their supporting items comprise 433 images, 105 videos, and 209 documents (282 items total), with an approximately balanced modality mix for evaluation.

\noindent\textbf{Stage~4: Data enrichment.} 
To faithfully model real-world retrieval settings, where relevant evidence is hidden among many irrelevant items, we construct distraction candidates $\mathcal{D}^{-}$ for each question $q \in \mathcal{Q}^{\star}$. $\mathcal{D}^{-}$ is constructed using the keywords of each query generated by GPT-4o, followed by keyword-based web scraping, and is then filtered by CLIP similarity and \texttt{vidore/colqwen2-v0.1} score to ensure semantic plausibility without redundancy.
Manual verification further confirms that $\mathcal{D}^{-}$ never contains the correct answer, yielding a challenging yet unambiguous dataset of about 46K+ items.

\noindent\textbf{Data profile.} The final benchmark includes 747 questions and 46,260 items: 25,652 images, 10,419 videos, and 10,189 documents.  
Further examples and detailed analyses are provided in Appendix~\ref{sec:appendix_example}.

% \noindent\textbf{Manual verification.}  
% Each query is validated with retrieval models and human checks to confirm that answers come only from the designated supporting item.  
% Although retrieval targets are defined at the page or frame level, evaluation is conducted at the item level, since models process whole documents or videos.  
% This dual validation guarantees reliability and prevents shortcuts.

To sum up, MultiHaystack pairs rigorously filtered QA pairs with large-scale distractors to simulate real-world environments. 
Ground-truth evidence is annotated at the fine-grained page or frame level, while evaluation is conducted at the item level (e.g., entire documents/videos), enabling joint assessment of precise retrieval and complex reasoning in long-context scenarios.
% While ground-truth evidence is defined at the page or frame level, evaluation occurs at the full item level (entire documents/videos) to jointly assess retrieval and reasoning in long contexts.

\section{Experiments}
\label{sec:exper}
\subsection{Experimental Setup}

\noindent \textbf{Methods.}  
To evaluate models on our benchmark, we follow standard retrieval-augmented pipelines~\cite{chen2024document} and unify input modalities for retrieval: images are used directly, videos are represented by 8 uniformly sampled frames over the full duration, and documents are rendered page-by-page into sequential images~\cite{ma2024mmlongbenchdoc}. 
QA pairs are annotated at the \emph{page/frame level} but evaluated at the \emph{item level}: retrieval is correct if the item containing the annotated page or frame appears in the top-$k$ results. 
For answer generation, we provide the full retrieved item as evidence, consistent with standard retrieval-augmented generation. 
We use a vision--language encoder (e.g., SigLIP~\cite{zhai2023sigmoid}) to score and rank corpus candidates, then pass the top-$k$ evidence with the question to the multimodal model. 
When models impose input restrictions, such as maximum context length or incomplete video support, we apply a unified preprocessing policy (e.g., truncating long items by prioritizing content around the matched page/frame) to ensure compatibility while keeping comparisons fair.

\begin{table*}[!t]
\centering
\caption{\textbf{Retrieval performance in cross-modality vs.\ single-modality settings.} 
Cross-modality results are shown in black, while single-modality results are shown in {\color{gray}gray} for comparison. 
Best values per column are highlighted in bold.}
\label{tab:retrieval_different_modality}
\resizebox{\textwidth}{!}{%
\setlength{\tabcolsep}{5pt}
\renewcommand{\arraystretch}{0.95}
\begin{tabular}{l ccc ccc ccc ccc}
\toprule
\multirow{2}{*}{\textbf{Model}} &
\multicolumn{3}{c}{\textbf{Video}} &
\multicolumn{3}{c}{\textbf{Image}} &
\multicolumn{3}{c}{\textbf{Document}} &
\multicolumn{3}{c}{\textbf{Overall}} \\
\cmidrule(lr){2-4} \cmidrule(lr){5-7} \cmidrule(lr){8-10} \cmidrule(lr){11-13}
& \textbf{R@1} & \textbf{R@3} & \textbf{R@5}
& \textbf{R@1} & \textbf{R@3} & \textbf{R@5}
& \textbf{R@1} & \textbf{R@3} & \textbf{R@5}
& \textbf{R@1} & \textbf{R@3} & \textbf{R@5} \\
\midrule

CLIP & 26.67  {\color{gray}\scriptsize(56.19)} & 40.00 {\color{gray}\scriptsize(78.10)} & 51.43 {\color{gray}\scriptsize(80.00)} & 21.71 {\color{gray}\scriptsize(30.25)} & 31.64 {\color{gray}\scriptsize(40.88)} & 34.87 {\color{gray}\scriptsize(44.34)} & 34.93 {\color{gray}\scriptsize(38.76)} & 46.89 {\color{gray}\scriptsize(51.67)} & 48.80 {\color{gray}\scriptsize(53.59)} & 26.10 {\color{gray}\scriptsize(36.28)} & 37.08 {\color{gray}\scriptsize(49.13)} & 41.10 {\color{gray}\scriptsize(51.94)} \\
% \rowcolor{gray!10} & {\color{gray}56.19} & {\color{gray}78.10} & {\color{gray}80.00} & {\color{gray}30.25} & {\color{gray}40.88} & {\color{gray}44.34} & {\color{gray}38.76} & {\color{gray}51.67} & {\color{gray}53.59} & {\color{gray}36.28} & {\color{gray}49.13} & {\color{gray}51.94} \\

SigLIP2 & \textbf{40.00} {\color{gray}\scriptsize(63.81)}  & \textbf{60.00} {\color{gray}\scriptsize(83.81)}  & \textbf{74.29} {\color{gray}\scriptsize(91.43)}  &  32.10  {\color{gray}\scriptsize(44.11)}  & 40.88   {\color{gray}\scriptsize(53.12)}  & 45.27  {\color{gray}\scriptsize(58.66)}  &  \textbf{59.81}  {\color{gray}\scriptsize(61.72)}  &  68.42 {\color{gray}\scriptsize(70.33)}  &  72.73 {\color{gray}\scriptsize(75.12)}  & \textbf{40.96}  {\color{gray}\scriptsize(51.81)}  &  51.27 {\color{gray}\scriptsize(62.25)}  & 57.03 {\color{gray}\scriptsize(67.87)}  \\
% \rowcolor{gray!10} & {\textbf{\color{gray}63.81}} & {\textbf{\color{gray}83.81}} & {\textbf{\color{gray}91.43}} & {\textbf{\color{gray}44.11}} & {\color{gray}53.12} & {\color{gray}58.66} & {\color{gray}61.72} & {\color{gray}70.33} & {\color{gray}75.12} & {\textbf{\color{gray}51.81}} & {\color{gray}62.25} & {\color{gray}67.87} \\

OpenCLIP & 38.10 
{\color{gray}\scriptsize(60.00)} &  56.19 {\color{gray}\scriptsize(74.29)} & 62.86 {\color{gray}\scriptsize(78.10)} & 19.40 {\color{gray}\scriptsize(25.40)} & 27.94 {\color{gray}\scriptsize(35.80)} & 32.33 {\color{gray}\scriptsize(42.26)} & 28.71 {\color{gray}\scriptsize(32.06)} & 36.84 {\color{gray}\scriptsize(42.58)} & 43.06 {\color{gray}\scriptsize(47.85)} & 24.63 {\color{gray}\scriptsize(32.13)} & 34.40 {\color{gray}\scriptsize(43.11)} & 39.63 {\color{gray}\scriptsize(48.86)} \\
% \rowcolor{gray!10} & {\color{gray}60.00} & {\color{gray}74.29} & {\color{gray}78.10} & {\color{gray}25.40} & {\color{gray}35.80} & {\color{gray}42.26} & {\color{gray}32.06} & {\color{gray}42.58} & {\color{gray}47.85} & {\color{gray}32.13} & {\color{gray}43.11} & {\color{gray}48.86} \\

Jina-Clip-V1 & 21.90 
{\color{gray}\scriptsize(42.86)} & 38.10 {\color{gray}\scriptsize(59.05)} & 47.62 {\color{gray}\scriptsize(67.62)} & 7.39 {\color{gray}\scriptsize(13.39)} & 10.16 {\color{gray}\scriptsize(19.17)} & 12.93 {\color{gray}\scriptsize(22.40)} &  16.75 {\color{gray}\scriptsize(17.70)} & 21.05 {\color{gray}\scriptsize(22.01)} & 22.49 {\color{gray}\scriptsize(22.97)} & 12.05 {\color{gray}\scriptsize(18.74)} & 17.14 {\color{gray}\scriptsize(25.57)} & 20.48 {\color{gray}\scriptsize(28.92)} \\
% \rowcolor{gray!10} & {\color{gray}42.86} & {\color{gray}59.05} & {\color{gray}67.62} & {\color{gray}13.39} & {\color{gray}19.17} & {\color{gray}22.40} & {\color{gray}17.70} & {\color{gray}22.01} & {\color{gray}22.97} & {\color{gray}18.74} & {\color{gray}25.57} & {\color{gray}28.92} \\

Jina-Clip-V2 & 20.00 
{\color{gray}\scriptsize(36.19)} & 30.48 {\color{gray}\scriptsize(56.19)} & 35.24 {\color{gray}\scriptsize(76.19)} & 11.78 {\color{gray}\scriptsize(27.25)} & 21.02 {\color{gray}\scriptsize(42.73)} & 25.17 {\color{gray}\scriptsize(48.04)} & 40.67 {\color{gray}\scriptsize(41.63)} & 51.67 {\color{gray}\scriptsize(52.63)} & 55.98  {\color{gray}\scriptsize(56.46)} & 21.02 {\color{gray}\scriptsize(32.53)} & 30.92 {\color{gray}\scriptsize(47.39)} & 35.21 {\color{gray}\scriptsize(54.35)} \\
% \rowcolor{gray!10} & {\color{gray}36.19} & {\color{gray}56.19} & {\color{gray}76.19} & {\color{gray}27.25} & {\color{gray}42.73} & {\color{gray}48.04} & {\color{gray}41.63} & {\color{gray}52.63} & {\color{gray}56.46} & {\color{gray}32.53} & {\color{gray}47.39} & {\color{gray}54.35} \\

NEV & 25.71
{\color{gray}\scriptsize(38.10)} & 40.00 {\color{gray}\scriptsize(54.29)} & 42.86 {\color{gray}\scriptsize(60.95)} & 5.31 {\color{gray}\scriptsize(8.78)} & 7.39 {\color{gray}\scriptsize(12.01)} & 8.78 {\color{gray}\scriptsize(13.63)} & 9.09 {\color{gray}\scriptsize(10.53)} & 12.92 {\color{gray}\scriptsize(13.88)} & 13.88 {\color{gray}\scriptsize(16.27)} & 9.24 {\color{gray}\scriptsize(13.39)} & 13.52 {\color{gray}\scriptsize(18.47)} & 14.99 {\color{gray}\scriptsize(21.02)} \\
% \rowcolor{gray!10} & {\color{gray}38.10} & {\color{gray}54.29} & {\color{gray}60.95} & {\color{gray}8.78} & {\color{gray}12.01} & {\color{gray}13.63} & {\color{gray}10.53} & {\color{gray}13.88} & {\color{gray}16.27} & {\color{gray}13.39} & {\color{gray}18.47} & {\color{gray}21.02} \\

E5-V & 34.29 
{\color{gray}\scriptsize(62.86)} & 51.43 {\color{gray}\scriptsize(81.90)} & 60.95 {\color{gray}\scriptsize(83.81)} & \textbf{33.49}  {\color{gray}\scriptsize(43.19)} & \textbf{55.20} {\color{gray}\scriptsize(68.36)} & \textbf{62.82} {\color{gray}\scriptsize(73.44)} & 59.33 {\color{gray}\scriptsize(60.77)} & \textbf{70.33} {\color{gray}\scriptsize(71.29)} & \textbf{75.12} {\color{gray}\scriptsize(76.08)} &  40.83 {\color{gray}\scriptsize(50.87)} & \textbf{58.90} {\color{gray}\scriptsize(71.08)} & \textbf{66.00} {\color{gray}\scriptsize(75.64)} \\
% \rowcolor{gray!10} & {\color{gray}62.86} & {\color{gray}81.90} & {\color{gray}83.81} & {\color{gray}43.19} & {\textbf{\color{gray}68.36}} & {\textbf{\color{gray}73.44}} & {\color{gray}60.77} & {\textbf{\color{gray}71.29}} & {\textbf{\color{gray}76.08}} & {\color{gray}50.87} & {\textbf{\color{gray}71.08}} & {\textbf{\color{gray}75.64}} \\

MM-Embed & 37.14 
{\color{gray}\scriptsize(60.95)} & 47.62 {\color{gray}\scriptsize(80.00)} & 55.24 {\color{gray}\scriptsize(87.62)} & 31.41 {\color{gray}\scriptsize(43.65)} & 43.65 {\color{gray}\scriptsize(64.43)} & 51.27 {\color{gray}\scriptsize(67.21)} & 53.59 {\color{gray}\scriptsize(62.68)} & 62.68 {\color{gray}\scriptsize(67.46)} & 70.81 {\color{gray}\scriptsize(75.60)} & 38.42 {\color{gray}\scriptsize(51.41)} & 49.53 {\color{gray}\scriptsize(67.47)} & 57.30 {\color{gray}\scriptsize(72.42)} \\
% \rowcolor{gray!10} & {\color{gray}60.95} & {\color{gray}80.00} & {\color{gray}87.62} & {\color{gray}43.65} & {\color{gray}64.43} & {\color{gray}67.21} & {\textbf{\color{gray}62.68}} & {\color{gray}67.46} & {\color{gray}75.60} & {\color{gray}51.41} & {\color{gray}67.47} & {\color{gray}72.42} \\

\bottomrule
\end{tabular}}
\end{table*}

\noindent \textbf{Baselines.}  
\noindent We evaluate two categories of baselines: VLMs for multimodal retrieval and MLLMs for multimodal reasoning.

For multimodal retrieval, we benchmark two categories of VLMs: 
\emph{CLIP-based models}, including CLIP~\cite{radford2021learning}, OpenCLIP~\cite{Cherti2022ReproducibleSL}, and Jina-CLIP v1/v2~\cite{2405.20204,koukounas2024jinaclipv2multilingualmultimodalembeddings}; 
and \emph{multimodal embedding models}, including SigLIP2~\cite{tschannen2025siglip2multilingualvisionlanguage}, Nomic-Embed-Vision~\cite{nussbaum2024nomic}, E5-V~\cite{jiang2024e5vuniversalembeddingsmultimodal}, and MM-Embed~\cite{lin2025mmembeduniversalmultimodalretrieval}.

For multimodal reasoning, we evaluate two categories of MLLMs: 
\emph{open-source models}, including Ola-7B~\cite{liu2025ola}, InternVL-3-8B~\cite{Zhu2025InternVL3EA}, and Qwen2-VL-7B~\cite{Qwen2VL}; 
and \emph{proprietary models}, including GPT-5~\cite{openai2025-gpt5} and Gemini-2.5-Flash~\cite{google2024gemini}.

\noindent \textbf{Metrics.}  
\noindent 
Retrieval is measured at the \emph{item level}, reporting Recall@1/3/5 to indicate whether the ground-truth item appears in the top-$k$ results, since real-world applications often require retrieving the entire file rather than a single page or frame. 
While long documents and videos are internally segmented into pages or frames to support downstream reasoning, retrieval always operates on complete items, and evaluation strictly follows this item-level definition.
Reasoning accuracy is evaluated using GPT-4o-mini as an automatic judge under a fixed rubric; we further validate judge reliability via human annotation.  
Details of the rubric are provided in Appendix~\ref{sec:prompt}.

\subsection{Multimodal Retrieval Results}
Table~\ref{tab:retrieval_different_modality} compares both single-modality and cross-modality retrieval to reveal the strengths and weaknesses of current retrieval models. 

\noindent \textbf{Single Modality.}  
When restricted to a single modality, current models achieve strong performance. For instance, SigLIP2 exceeds 90\% Recall@5 on videos, while MM-Embed surpasses 75\% Recall@5 on documents. 
Such results suggest that single-modal retrieval is already well handled by modern VLMs, likely because candidates are modality-homogeneous and free from cross-modal embedding interference. 
Therefore, these benchmarks provide limited diagnostic power for revealing the failures that emerge in realistic heterogeneous environments.

\noindent \textbf{Cross Modalities.}
In contrast, cross-modal retrieval remains highly challenging.  
Even the strongest models, SigLIP2 and E5-V, reach only 40.96\% and 40.83\% R@1—drops of over 40 points from their unimodal results. MM-Embed attains relatively higher recall at R@5 (57.30\%), yet still falls well short of its unimodal performance. Weaker baselines degrade even further, with document retrieval proving the most difficult.
These findings indicate that retrieval over heterogeneous modalities remains the dominant failure mode, motivating MultiHaystack as a diagnostic benchmark for cross-modal grounding at scale.

\begin{table*}[!t]
\centering
\caption{\textbf{Retrieval results across six tasks.} Recall@1/3/5 of different vision-language retrieval models on MultiHaystack across six distinct tasks.}
\label{tab:retrieval_different_task}
\resizebox{\textwidth}{!}{%
  \setlength{\tabcolsep}{6pt}%
  \renewcommand{\arraystretch}{0.9}%
  \begin{tabular}{l ccc ccc ccc ccc ccc ccc }
    \toprule
    \multirow{3}{*}{\textbf{Model}}& \multicolumn{3}{c}{\textbf{VPP}} &
    \multicolumn{3}{c}{\textbf{CU}} &
    \multicolumn{3}{c}{\textbf{VTR}} & \multicolumn{3}{c}{\textbf{FKR}} & \multicolumn{3}{c}{\textbf{SR}} 
    & \multicolumn{3}{c}{\textbf{MI}} \\
    \cmidrule(lr){2-4} \cmidrule(lr){5-7}  \cmidrule(lr){8-10}  \cmidrule(lr){11-13}
    \cmidrule(lr){14-16}  \cmidrule(lr){17-19}
    % \multicolumn{10}{c}{\cellcolor{pink} \textbf{\textit{Perceptual Understanding }}} \\   
    & \textbf{R@1} & \textbf{R@3} & \textbf{R@5}
    & \textbf{R@1} & \textbf{R@3} & \textbf{R@5}
    & \textbf{R@1} & \textbf{R@3} & \textbf{R@5} & \textbf{R@1} & \textbf{R@3} & \textbf{R@5}
    & \textbf{R@1} & \textbf{R@3} & \textbf{R@5}
    & \textbf{R@1} & \textbf{R@3} & \textbf{R@5} \\
\midrule    
    CLIP                   & 33.33 & 39.39 & 42.42 & 16.67 & 30.00 & 43.33 & 29.55 & 40.91 & 50.00  & 20.59 & 23.53 & 29.41 & 25.23 & 35.51 & 38.63 & 27.37 & 40.35 & 43.51 \\
    SigLIP2                & \textbf{42.42} & \textbf{66.67} & \textbf{72.73} & \textbf{53.33} & \textbf{66.67}  & \textbf{80.00} & 29.55 & 52.27 & 65.91 & \textbf{26.47}  & 32.35 & 47.06 & 38.01  & 45.48  & 50.47 & \textbf{46.32} & 56.49 & 60.00 \\
    OpenCLIP               & 36.36 & 51.52 & 54.55 & 30.00 & 46.67 & 50.00 & \textbf{38.64} & \textbf{61.36}  & \textbf{70.45}  & 17.65 & 20.59 & 20.59 & 22.74 & 31.78 & 37.69 & 23.51 & 31.58 & 36.49 \\
    Jina-Clip-V1           & 21.21 & 27.27 & 39.39 & 13.33 & 13.33 & 20.00 & 29.55 & 59.09 & \textbf{70.45} & 2.94 & 8.82 & 8.82 & 9.03 & 13.08 & 15.26 & 12.63 & 15.44 & 17.89 \\
    Jina-Clip-V2           & 33.33 & 42.42 & 48.48 & 10.00 & 16.67 & 23.33 & 11.36 & 20.45 & 25.00 & 11.76 & 20.59 & 23.53 & 17.13 & 28.35 & 31.78 & 27.72 & 36.84  & 41.75 \\
    NEV     & 15.15 & 24.24 & 27.27 & 16.67 & 26.67 & 30.00 & 34.09  & 50.00 & 54.55 &  0.00 &  0.00 & 2.94 & 7.17 & 10.59 & 11.84 &  7.37 &  10.18 &  10.88 \\
    E5-V     & 42.42 & 57.58 & 66.67 & 36.67 & 43.33  & 43.33 & \textbf{38.64} & \textbf{61.36} & \textbf{70.45} & \textbf{26.47}  & \textbf{44.12} & \textbf{58.82}  & \textbf{38.63}  & \textbf{62.31} & \textbf{69.78} & 45.61 & \textbf{58.25} & \textbf{64.21} \\
    MM-Embed     & 36.36 & 45.45 & 54.55 & 30.00 & 36.67 & 40.00 & 27.27  & 56.82 & 59.09  & 20.59 & 29.41 & 44.12 & 37.69 & 53.27 & 64.49 & 44.21  & 48.42 & 52.63 \\
% \cmidrule{1-10}
% \multicolumn{10}{c}{\cellcolor{TealBlue} \textbf{\textit{Analytical Reasoning}}} \\
% \cmidrule{1-10}
%     & \multicolumn{3}{c}{\textbf{Factual Knowledge Retrieval}} & \multicolumn{3}{c}{\textbf{Statistical Reasoning}} 
%     & \multicolumn{3}{c}{\textbf{Metadata Identification}} \\
% \midrule    
    
    % CLIP                   & 20.59 & 23.53 & 29.41 & 25.23 & 35.51 & 38.63 & 27.37 & 40.35 & 43.51 \\
    % SigLIP2                & \textbf{26.47}  & 32.35 & 47.06 & 38.01  & 45.48  & 50.47 & \textbf{46.32} & 56.49 & 60.00 \\
    % OpenCLIP               & 17.65 & 20.59 & 20.59 & 22.74 & 31.78 & 37.69 & 23.51 & 31.58 & 36.49 \\
    % Jina-Clip-V1           & 2.94 & 8.82 & 8.82 & 9.03 & 13.08 & 15.26 & 12.63 & 15.44 & 17.89 \\
    % Jina-Clip-V2           & 11.76 & 20.59 & 23.53 & 17.13 & 28.35 & 31.78 & 27.72 & 36.84  & 41.75 \\
    % Nomic-Embed-Vision     &  0.00 &  0.00 & 2.94 & 7.17 & 10.59 & 11.84 &  7.37 &  10.18 &  10.88 \\
    % E5-V     & \textbf{26.47}  & \textbf{44.12} & \textbf{58.82}  & \textbf{38.63}  & \textbf{62.31} & \textbf{69.78} & 45.61 & \textbf{58.25} & \textbf{64.21} \\
    % MM-Embed     & 20.59 & 29.41 & 44.12 & 37.69 & 53.27 & 64.49 & 44.21  & 48.42 & 52.63 \\
    \bottomrule
  \end{tabular}%
}
\end{table*}

\begin{wraptable}[9]{r}{0.5\textwidth}
\vspace{-45pt}
\renewcommand{\arraystretch}{1}
\begin{center}
  \centering
\caption{\textbf{Multimodal reasoning performance.} Each model answers questions using top-$5$ items retrieved by E5-V
from cross-modality inputs; {\color{gray}gray} numbers show single-modality Recall@5 for reference.}
  \label{tab:vqa_single}
  \resizebox{0.5\textwidth}{!}{%
    \setlength{\tabcolsep}{6pt}%
    \renewcommand{\arraystretch}{1.0}%
    \begin{tabular}{l c c c c}
      \toprule
      \textbf{Model} & \textbf{Video} & \textbf{Image} & \textbf{Document} & \textbf{Overall} \\
      \midrule
      Ola      
      & 14.29 {\color{gray}\scriptsize(22.86)} 
      & 20.09 {\color{gray}\scriptsize(31.41)} 
      & 36.36 {\color{gray}\scriptsize(44.98)} 
      & 23.83 {\color{gray}\scriptsize(34.00)} \\

      InternVL-3 
      & 17.14 {\color{gray}\scriptsize(20.95)} 
      & 29.33 {\color{gray}\scriptsize(38.80)} 
      & 49.28 {\color{gray}\scriptsize(51.67)} 
      & 33.29 {\color{gray}\scriptsize(39.89)} \\

      Qwen2-VL  
      & 16.19 {\color{gray}\scriptsize(18.10)} 
      & 16.86 {\color{gray}\scriptsize(24.94)} 
      & 19.62 {\color{gray}\scriptsize(22.49)} 
      & 17.54 {\color{gray}\scriptsize(23.29)} \\

      Gemini-2.5-Flash
      & 52.38 {\color{gray}\scriptsize(61.90)} 
      & 35.10 {\color{gray}\scriptsize(44.57)} 
      & 56.94 {\color{gray}\scriptsize(58.37)} 
      & 43.64 {\color{gray}\scriptsize(50.87)} \\

      GPT-5   
      & \textbf{60.00} {\color{gray}\scriptsize(67.62)} 
      & \textbf{43.19} {\color{gray}\scriptsize(52.66)} 
      & \textbf{64.11} {\color{gray}\scriptsize(70.81)} 
      & \textbf{51.41} {\color{gray}\scriptsize(59.84)} \\
      \bottomrule
    \end{tabular}%
  }
\end{center}
\end{wraptable}

\newpage

\subsection{Multimodal Reasoning Results}
As shown in Table~\ref{tab:vqa_single}, we evaluate multimodal reasoning accuracy when conditioning on items retrieved from single-modality versus cross-modality search, using the same retriever (E5-V) for all MLLMs to isolate the effect of retrieval quality. 
GPT-5 achieves the highest overall performance, reaching 59.84\% with single-modality retrieval but only 51.41\% under cross-modal retrieval. 
Gemini-2.5-Flash follows a similar pattern, dropping from 50.87\% to 43.64\%. 
In contrast, weaker models such as Ola and Qwen2-VL remain below 25\% overall even with unimodal retrieval, indicating limited grounding ability. 
Overall, retrieval errors propagate directly into reasoning, and improving multimodal reasoning at scale is inseparable from robust cross-modal retrieval.

\begin{figure}[!t]
    \centering
    \includegraphics[width=\linewidth]{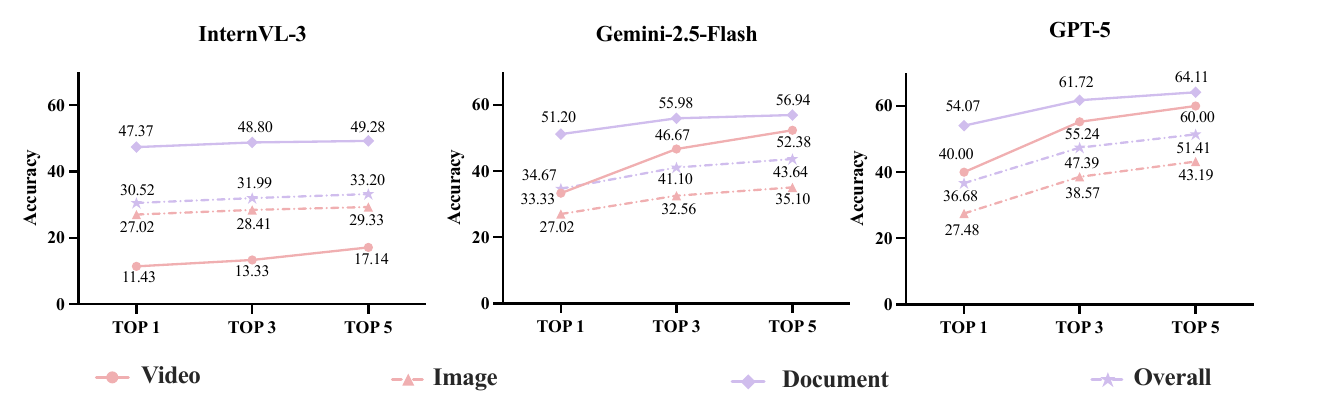}
    \caption{\textbf{Top-$k$ ablation analysis for MLLMs integrated with E5-V.} }
    \label{fig:ablation}
\end{figure}

\subsection{Discussion}

\subsubsection{How does performance vary across tasks?}
We further break down results by task category (Table~\ref{tab:retrieval_different_task} and Table~\ref{tab:vqa_different_task}) to reveal capability gaps that are masked by aggregate ``Overall'' metrics.

\noindent \textbf{Multimodal Retrieval.} 
Table~\ref{tab:retrieval_different_task} shows substantial variation across task categories, exposing distinct retrieval gaps. 
SigLIP2 and E5-V achieve over 60\% Recall@5 on Visual Parsing and Positioning and Video Temporal Reasoning, indicating strength in spatial parsing and temporal alignment. 
However, both drop below 50\% on Factual Knowledge Retrieval and Statistical Reasoning, where the evidence often hinges on fine-grained entities, numbers, or cross-page/frame context rather than global visual similarity, making nearest-neighbor retrieval more brittle. 
MM-Embed is more balanced across tasks, but this does not close the gap on reasoning-intensive categories. 
Overall, these discrepancies show that aggregate retrieval metrics can mask meaningful failure modes, and fine-grained task breakdown is essential for diagnosing where retrieval most often fails.

\begin{wraptable}[8]{r}{0.52\textwidth}
\vspace{-22pt}
\renewcommand{\arraystretch}{1}
\begin{center}
\setlength{\tabcolsep}{2.5pt}
\caption{\textbf{Comparison of MLLMs' reasoning performance integrated with E5-V across six tasks.}}
\label{tab:vqa_different_task}
\resizebox{\linewidth}{!}{%
  \setlength{\tabcolsep}{6pt}%
  \renewcommand{\arraystretch}{0.9}%
\begin{tabular}{l *{6}{c}}
    \toprule
    \textbf{Model} &
    \textbf{VPP} &
    \textbf{CU} &
    \textbf{VTR} &
    \textbf{FKR} &
    \textbf{SR} &
    \textbf{MI} \\
    \midrule
    Ola            & 27.27 & 30.00 & 18.18  & 20.59 & 26.17 & 21.40\\
    InternVL-3     & 42.42 & 23.33 & 11.36 & 23.53 & 29.91 & 41.40\\
    Qwen2-VL       & 18.18 & 23.33 & 6.82 & 14.71 & 18.38 & 17.89 \\
    Gemini-2.5-Flash & 54.55  & 46.67  & 56.82 & 32.35 & 35.51 & 50.53 \\
    GPT-5         & \textbf{57.58} & \textbf{56.67} & \textbf{52.27} & \textbf{50.00} & \textbf{43.61} & \textbf{58.95}\\
    % \cmidrule{1-4}
    % %======================================================
    % \textbf{} &
    % \textbf{FKR} &
    % \textbf{SR} &
    % \textbf{MI} \\
    % \midrule
    % Ola            & 20.59 & 26.17 & 21.40 \\
    % InternVL-3     & 23.53 & 29.91 & 41.40 \\
    % Qwen2-VL       & 14.71 & 18.38 & 17.89 \\
    % Gemini-2.5-Flash & 32.35 & 35.51 & 50.53 \\
    % GPT-5         & \textbf{50.00} & \textbf{43.61} & \textbf{58.95} \\
    \bottomrule
  \end{tabular}%
}
\end{center}
\end{wraptable}

\noindent \textbf{Multimodal Reasoning.} 
Table~\ref{tab:vqa_different_task} shows that reasoning performance varies substantially across task types and often mirrors retrieval difficulty. 
GPT-5 achieves the best overall results, performing strongly on Metadata Identification (58.95\%) and Visual Parsing and Positioning (57.58\%), where answers typically rely on localized visual cues or explicit metadata. 
However, its accuracy drops on Statistical Reasoning (43.61\%), suggesting that numerical aggregation and quantitative reasoning remain challenging even when relevant evidence is retrieved. 
Gemini-2.5-Flash follows a similar pattern, while weaker models such as Ola and Qwen2-VL remain below 30\% on most tasks, indicating limited ability to integrate retrieved evidence into reliable reasoning chains. 
Overall, these results highlight that multimodal reasoning ability is highly task-dependent and reinforce the need for fine-grained evaluation to diagnose where reasoning failures occur.

\begin{figure}[!t]
    \centering
    \includegraphics[width=0.98\linewidth]{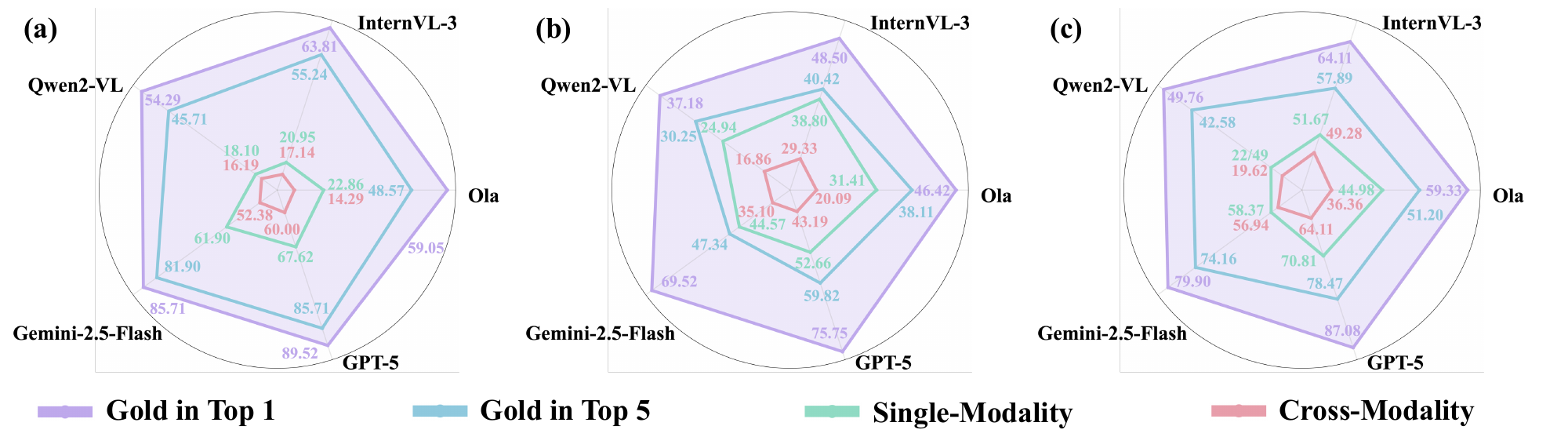}
    \vspace{-1mm}
    \caption{\textbf{Comparison in three distinct modalities.} (a) represents the video modality, (b) represents the image modality, and (c) represents the document modality.}
    \vspace{-3.5mm}
    \label{fig:video_image_doc_radar}
\end{figure}

\begin{wraptable}[6]{r}{0.6\textwidth}
\vspace{-44pt}
\renewcommand{\arraystretch}{1}
\begin{center}
  \centering
   \caption{\textbf{Reliability Matrix for LLM-as-Judge}}
   \vspace{-8pt}
  \label{tab:llm_eval}
  \resizebox{\linewidth}{!}{%
    \setlength{\tabcolsep}{6pt}%
    \renewcommand{\arraystretch}{0.9}%
    \begin{tabular}{l c c c c}
      \toprule
      \textbf{Model}           & \textbf{Cohen's $\kappa$} & \textbf{95\% CI} & \textbf{Accuracy} & \textbf{Pearson $r$} \\
      \midrule
      Ola         & 0.918 & {[0.710, 1.000]} & 0.967 & 0.921  \\
      InternVL-3  & 0.865 & {[0.667, 1.000]} & 0.933 & 0.873 \\
      Qwen2-VL    & 1.000 & {[1.000, 1.000]} & 1.000 & 1.000 \\
      Gemini-2.5-Flash      & 0.932 & {[0.772, 1.000]} & 0.967 & 0.934 \\
      GPT-5       & 1.000 & {[1.000, 1.000]} & 1.000 & 1.000 \\
      \bottomrule
    \end{tabular}%
  }
\end{center}
\end{wraptable}

\subsubsection{Can LLMs serve as reliable judges?}
\label{sec:llm_judge}
We sample 30 non-overlapping QA pairs for each model and ask MTurk workers to label answers as correct or incorrect, then compare their judgments with GPT-4o-mini. 
Table~\ref{tab:llm_eval} shows strong consistency: Cohen’s $\kappa$ exceeds 0.86 and accuracy remains above 93\% across all models, with Qwen2-VL and GPT-5 reaching perfect agreement. 
These results suggest that GPT-4o-mini provides human-level reliability for answer verification, enabling rigorous and cost-efficient evaluation at scale.

\subsubsection{How does Top-k affect retrieval and reasoning?}
\Cref{fig:ablation} presents a top-$k$ ablation analysis for three MLLMs integrated with E5-V. 
As expected, reasoning accuracy improves from Top-1 to Top-5 retrieved items, confirming that retrieval coverage is a key bottleneck. 
GPT-5 benefits the most, reaching 64.11\% overall accuracy at Top-5, while Gemini-2.5-Flash shows moderate gains, and InternVL-3 remains consistently weaker. 
Across modalities, documents yield the largest improvements, likely because relevant evidence is more dispersed across pages, making higher $k$ more effective for coverage. 
However, even with five retrieved items, substantial gaps remain across modalities and models, highlighting that retrieval improvements must be complemented by stronger reasoning robustness.

\subsubsection{What is the gap between Gold, Single-Modality, and Cross-Modality settings?}

\begin{figure}[!t]
    \centering
    \includegraphics[width=\linewidth]{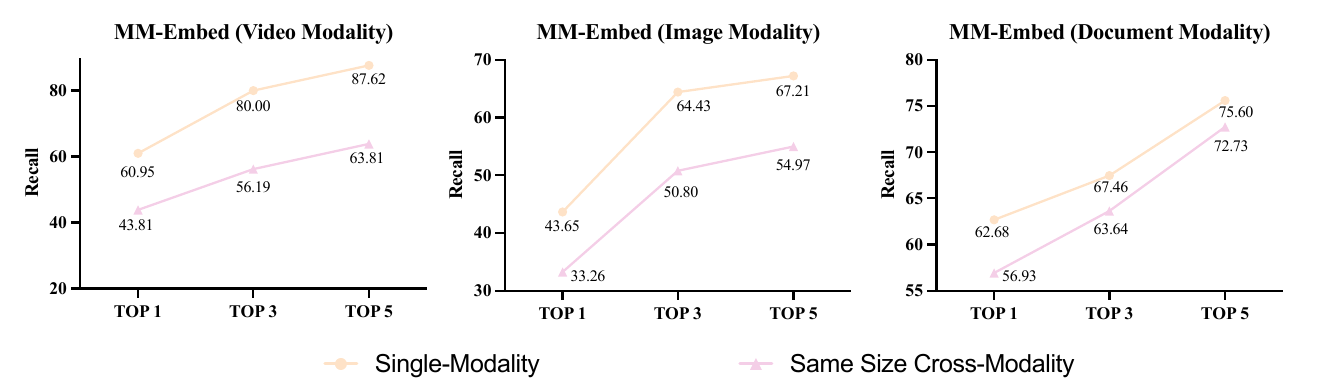}
    \vspace{-3mm}
    \caption{\textbf{Pool-size controlled comparison.} Recall of MM-Embed under single-modality retrieval and mixed-modality retrieval with an identical total pool size. Performance remains substantially lower in the mixed-modality condition, indicating that the cross-modal gap arises from modality heterogeneity rather than pool size.}
    \label{fig:poolsize}
    \vspace{-3mm}
\end{figure}

\Cref{fig:video_image_doc_radar} compares model performance under Gold, single-modality, and cross-modality settings across video, image, and document tasks. 
Gold Top-1/5 provides an upper bound by directly supplying the answer-containing item. 
Single-modality retrieval approaches this level across all modalities, suggesting that models perform reliably when evidence remains within a single modality. 
In contrast, cross-modality retrieval leads to substantial declines, with the largest drop in videos, followed by documents and images, reflecting the difficulty of aligning heterogeneous modalities under temporal and semantic variability. 
Even GPT-5 shows pronounced degradation in this setting, indicating that unimodal evaluations can conceal realistic failure modes.

To understand the source of this degradation, \Cref{fig:poolsize} compares single-modality and mixed-modality retrieval under the same total pool size. 
Performance remains consistently lower in the mixed-modality condition, showing that the cross-modal gap is not driven by pool size but by modality heterogeneity, including embedding mismatch and semantic interference. 
Together, these results reinforce the benchmark’s motivation: reasoning becomes relatively reliable once correct evidence is surfaced, but identifying that evidence in large heterogeneous environments remains the dominant bottleneck.

\subsubsection{Why is data enrichment essential for large-scale evaluation?}
\begin{figure}[!t]
    \centering
    \includegraphics[width=\linewidth]{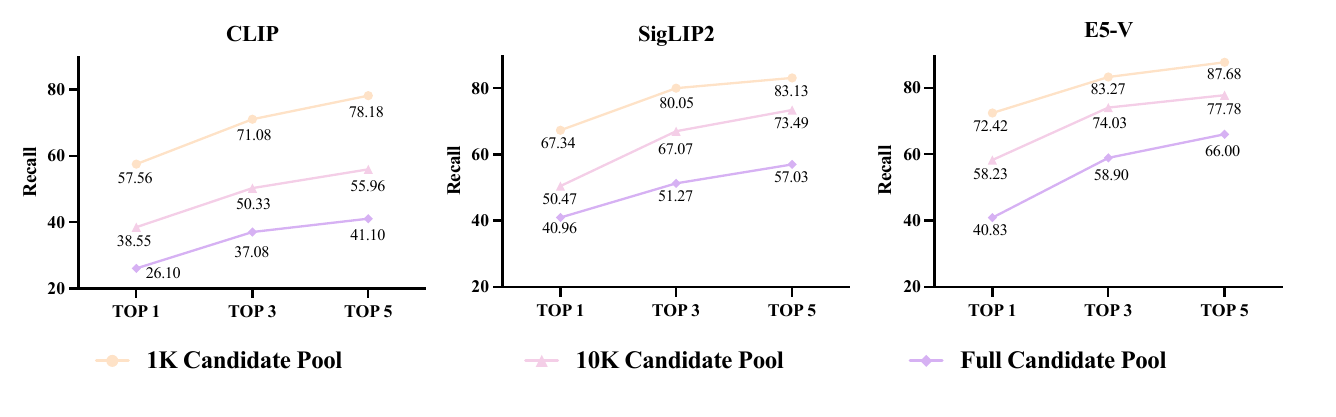}
    \vspace{-3mm}
    \caption{\textbf{Effect of data enrichment under varying candidate pool sizes,} showing that recall consistently drops as the pool expands.}
    \vspace{-3mm}
    \label{fig:distractor}
\end{figure}

We simulate realistic retrieval by progressively enlarging the candidate pool: all positives are retained, and distractors are added until reaching 1K, 10K, and the full corpus. 
As shown in \Cref{fig:distractor}, recall consistently drops as the pool expands, and the degradation rate reflects model robustness at scale. 
CLIP degrades sharply, which is consistent with a stronger reliance on surface-level similarity, whereas SigLIP2 and E5-V degrade more gradually, indicating better discrimination under large heterogeneous pools. 
These robustness differences are often obscured in small candidate sets, underscoring the necessity of large-scale enrichment for faithfully evaluating real-world retrieval.
\begin{figure}[!t]
    \centering
    \includegraphics[width=0.98\linewidth]{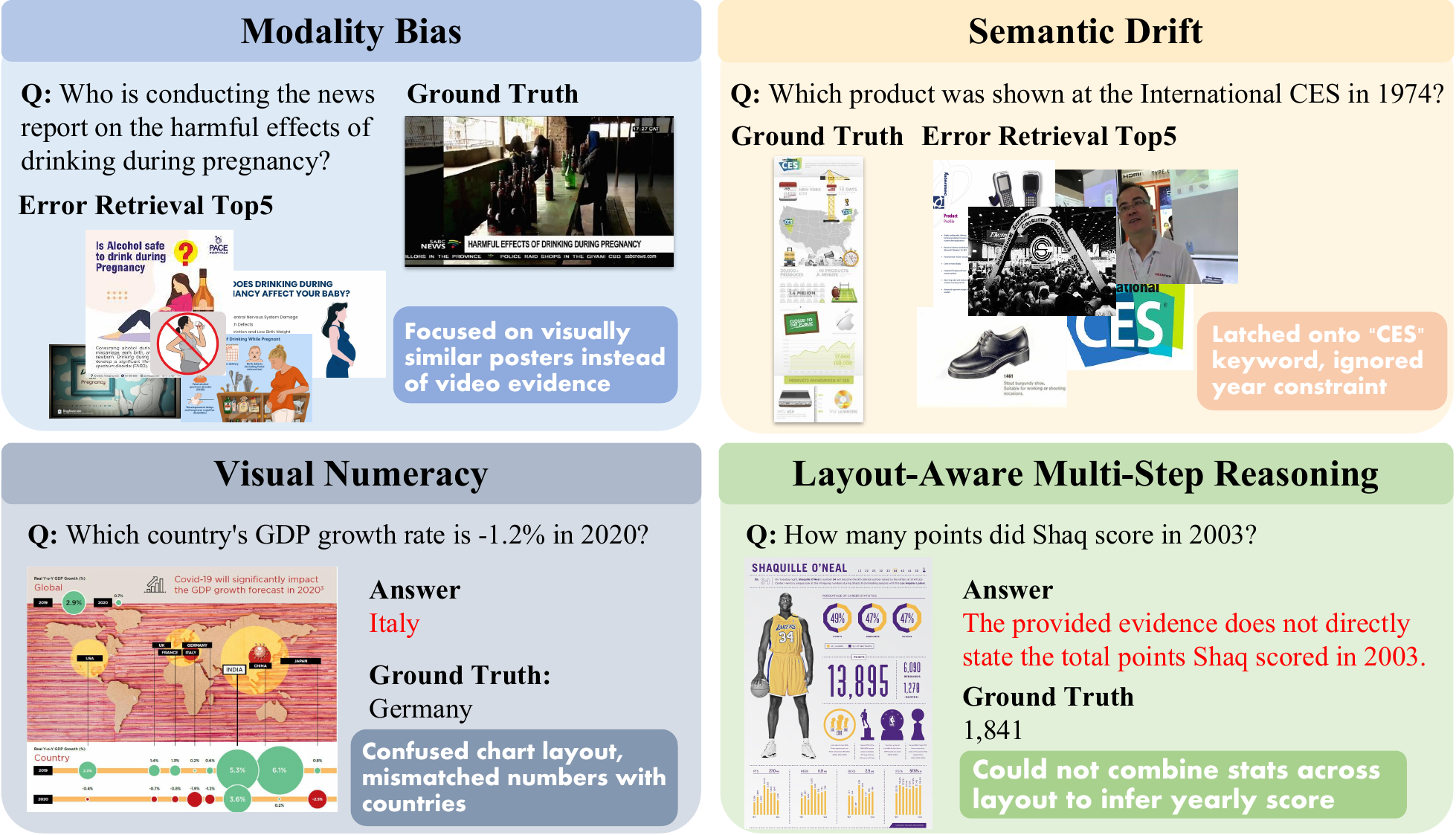}
    \vspace{-1mm}
    \caption{\textbf{Error cases} illustrating retrieval errors, including \emph{modality bias} (retrieving images instead of video evidence) and \emph{semantic drift} (violating temporal constraints), and reasoning errors, including \emph{visual numeracy} (misreading numbers in charts) and \emph{layout-aware multi-step reasoning} (failing to integrate structured cues across layouts).}
    \label{fig:error_analysis}
    \vspace{-3mm}
\end{figure}

\section{Error Analysis}
This section provides a detailed error analysis (\cref{fig:error_analysis}), distinguishing between retrieval failures (locating relevant evidence) and reasoning failures (interpreting the retrieved evidence).
More detailed analysis is provided in Appendix~\ref{sec:error}.

\noindent \textbf{Retrieval errors.}  
VLMs frequently exhibit modality bias, retrieving visually salient images instead of the correct video evidence, as well as semantic drift, where models favor frequent entities or global visual similarity while overlooking temporal or contextual constraints. 
These behaviors often lead to failures in temporal queries and factual retrieval tasks, particularly when the correct evidence requires aligning events across frames or identifying specific document regions. 
Such errors indicate a misalignment between retrieval objectives and query intent, highlighting the need for retrieval models that better account for modality-specific structure and constraint-aware matching.

\noindent \textbf{Reasoning errors.}  
Even when the correct evidence is retrieved, MLLMs still struggle with visual numeracy, such as mismatching chart values and axis labels, and with layout-aware multi-step reasoning, where information distributed across different regions of a document or frame must be integrated to reach the correct conclusion. 
These failures often arise when models must combine multiple cues, such as spatial layout, numerical values, and textual context, within a single reasoning chain. 
Overall, these patterns reveal persistent gaps in fine-grained perceptual grounding and compositional reasoning, suggesting the need for stronger numeracy capabilities and architectures that better exploit layout and structural cues.
\begin{table*}[!t]
\centering
\caption{\textbf{Preliminary studies with advanced retrieval pipelines.}}
\vspace{-2mm}
\label{tab:future}
\resizebox{\textwidth}{!}{%
\setlength{\tabcolsep}{5pt}
\renewcommand{\arraystretch}{0.95}
\begin{tabular}{l ccc ccc ccc ccc}
\toprule
\multirow{3}{*}{\textbf{Methods}} &
\multicolumn{3}{c}{\textbf{Video}} &
\multicolumn{3}{c}{\textbf{Image}} &
\multicolumn{3}{c}{\textbf{Document}} &
\multicolumn{3}{c}{\textbf{Overall}} \\
\cmidrule(lr){2-4} \cmidrule(lr){5-7} \cmidrule(lr){8-10} \cmidrule(lr){11-13}
& \textbf{R@1} & \textbf{R@3} & \textbf{R@5}
& \textbf{R@1} & \textbf{R@3} & \textbf{R@5}
& \textbf{R@1} & \textbf{R@3} & \textbf{R@5}
& \textbf{R@1} & \textbf{R@3} & \textbf{R@5} \\
\midrule
\rowcolor{gray!20}
\multicolumn{13}{c}{\textbf{\textit{Single-Modality}}} \\
\cmidrule{1-13}
E5-V     & 62.86 & 81.90 & 83.81 & 43.19 & 68.36 & 73.44 & 60.77 & 71.29 & 76.08 & 50.87 & 71.08 & 75.64 \\
\midrule
\rowcolor{gray!20}
\multicolumn{13}{c}{\textbf{\textit{Cross-Modality}}} \\
\cmidrule{1-13}
E5-V     & 34.29 & 51.43 & 60.95 & 33.49 & 55.20 & 62.82 & 59.33 & 70.33 & 75.12 & 40.83 & 58.90 & 66.00 \\
E5-V + Refined Query    & 41.90 & 59.05 & 64.76 & 36.26 & 56.58 & 66.74 & 60.29 & 70.81 & 75.12 & 43.78 & 60.91 & 68.81 \\
E5-V + MMSearch~\cite{jiang2025mmsearch} & 44.76 & 62.86 & 65.71 & 36.95  &   57.97 &  69.52 & 60.29 & 72.73 & 75.60  & 44.58 & 62.78  & 70.68 \\
\midrule
VisRAG~\cite{yu2025visragvisionbasedretrievalaugmentedgeneration} & 40.95 & 64.76 & 68.57 &  39.03   & 45.96    & 50.12   & 60.77 & 67.46   &  70.33   & 45.38 & 54.62    & 58.37  \\

\bottomrule
\end{tabular}}
\vspace{-2mm}
\end{table*}

\section{Future Directions}
Table~\ref{tab:future} reports preliminary  experiments with advanced retrieval pipelines~\cite{yu2025visragvisionbasedretrievalaugmentedgeneration, jiang2025mmsearch}. 
While these pipelines introduce query rewriting, iterative verification, and re-ranking, they deliver only modest improvements over naive retrieval and remain far below the single-modality upper bounds in \cref{sec:exper}. 
Under the E5-V cross-modal retrieval setting, the persistent gap suggests that lightweight query refinement and agentic loops alone are insufficient, and that the dominant challenges lie in cross-modal grounding: (i) misaligned embedding spaces across modalities, (ii) loss of fine-grained spatial/textual cues when evidence is compressed or standardized into item-level representations, and (iii) difficulty in aggregating heterogeneous evidence with temporal or numerical structure.

These results point to several research opportunities. 
First, more expressive and modality-aware representations are needed to preserve document layout regularities, video temporal consistency, and localized visual semantics, while remaining comparable for retrieval. 
Second, retrieval and reasoning should be more tightly coupled, e.g., allowing intermediate reasoning states to condition query rewriting, adaptive re-ranking, and evidence expansion, rather than relying on a largely static retrieve-then-read pipeline. 
Finally, systematic failures in contextual disambiguation, statistical reasoning, and fine-grained grounding motivate architectures that incorporate structural priors and calibrated uncertainty to guide evidence selection and aggregation. 
By making these bottlenecks explicit and measurable, MultiHaystack provides a diagnostic foundation for evaluating and developing next-generation cross-modal RAG systems.

\section{Conclusion}

We introduced \textbf{MultiHaystack}, a large-scale benchmark for evaluating Multimodal Large Language Models under realistic cross-modal retrieval and reasoning settings. 
With over 46,000 images, videos, and documents paired with evidence-grounded questions, it systematically reveals limitations that single-modality evaluations conceal. Our results highlight the need for retrieval-aware reasoning and modality-agnostic architectures. We hope MultiHaystack will provide a useful testbed for studying multimodal retrieval and reasoning at scale and support the development of more robust multimodal systems.

% \section*{Acknowledgements}
% Please insert your acknowledgments here.

\bibliographystyle{splncs04}
\bibliography{main}

\begin{thebibliography}{10}
\providecommand{\url}[1]{\texttt{#1}}
\providecommand{\urlprefix}{URL }
\providecommand{\doi}[1]{https://doi.org/#1}

\bibitem{google2024gemini}
AI, G.: Gemini: Google's multimodal ai model. Google AI Research  (2024), \url{https://fireflies.ai/blog/gemini-vs-gpt-4}

\bibitem{mint1t}
Awadalla, A., Xue, L., Lo, O., Shu, M., Lee, H., Guha, E.K., Jordan, M., Shen, S., Awadalla, M., Savarese, S., Xiong, C., Xu, R., Choi, Y., Schmidt, L.: Mint-1t: Scaling open-source multimodal data by 10x: A multimodal dataset with one trillion tokens (2024), \url{https://arxiv.org/abs/2406.11271}

\bibitem{chang2025wearvqavisualquestionanswering}
Chang, E., Huang, Z., Liao, Y., Bhavsar, S.R., Param, A., Stark, T., Ahmadyan, A., Yang, X., Wang, J., Abdullah, A., Nguyen, G., Iyer, A., Hall, D., Li, E., Moon, S., Scheffer, N., Ahmed, K., Damavandi, B., Wanga, R., Kumar, A., Patel, R., Dong, X.L.: Wearvqa: A visual question answering benchmark for wearables in egocentric authentic real-world scenarios (2025), \url{https://arxiv.org/abs/2511.22154}

\bibitem{webvqa2023}
Chang, Y., Narang, M., Suzuki, H., Cao, G., Gao, J., Bisk, Y.: Webqa: Multihop and multimodal qa (2022), \url{https://arxiv.org/abs/2109.00590}

\bibitem{chen2024document}
Chen, J., Xu, D., Fei, J., Feng, C.M., Elhoseiny, M.: Document haystacks: Vision-language reasoning over piles of 1000+ documents (2024), \url{https://arxiv.org/abs/2411.16740}

\bibitem{chen2023minigpt}
Chen, J., Zhu, D., Shen, X., Li, X., Liu, Z., Zhang, P., Krishnamoorthi, R., Chandra, V., Xiong, Y., Elhoseiny, M.: Minigpt-v2: large language model as a unified interface for vision-language multi-task learning (2023), \url{https://arxiv.org/abs/2310.09478}

\bibitem{chen2023pretrainedvisionlanguagemodels}
Chen, Y., Hu, H., Luan, Y., Sun, H., Changpinyo, S., Ritter, A., Chang, M.W.: Can pre-trained vision and language models answer visual information-seeking questions? (2023), \url{https://arxiv.org/abs/2302.11713}

\bibitem{Cherti2022ReproducibleSL}
Cherti, M., Beaumont, R., Wightman, R., Wortsman, M., Ilharco, G., Gordon, C., Schuhmann, C., Schmidt, L., Jitsev, J.: Reproducible scaling laws for contrastive language-image learning. In: 2023 IEEE/CVF Conference on Computer Vision and Pattern Recognition (CVPR). p. 2818–2829. IEEE (Jun 2023). \doi{10.1109/cvpr52729.2023.00276}, \url{http://dx.doi.org/10.1109/CVPR52729.2023.00276}

\bibitem{fang2024mmbenchvideolongformmultishotbenchmark}
Fang, X., Mao, K., Duan, H., Zhao, X., Li, Y., Lin, D., Chen, K.: Mmbench-video: A long-form multi-shot benchmark for holistic video understanding (2024), \url{https://arxiv.org/abs/2406.14515}

\bibitem{Farre2024FineVideo}
Farré, M., Marafioti, A., Tunstall, L., Von~Werra, L., Wolf, T.: Finevideo. \url{https://huggingface.co/datasets/HuggingFaceFV/finevideo} (2024)

\bibitem{ging2024openendedvqabenchmarkingvisionlanguage}
Ging, S., Bravo, M.A., Brox, T.: Open-ended vqa benchmarking of vision-language models by exploiting classification datasets and their semantic hierarchy (2024), \url{https://arxiv.org/abs/2402.07270}

\bibitem{DBLP:journals/ijcv/GoyalKASBP19/VQA-Matter}
Goyal, Y., Khot, T., Summers-Stay, D., Batra, D., Parikh, D.: Making the v in vqa matter: Elevating the role of image understanding in visual question answering (2017), \url{https://arxiv.org/abs/1612.00837}

\bibitem{evermemos}
Hu, C., Gao, X., Zhou, Z., Xu, D., Bai, Y., Li, X., Zhang, H., Li, T., Zhang, C., Bing, L., Deng, Y.: Evermemos: A self-organizing memory operating system for structured long-horizon reasoning (2026), \url{https://arxiv.org/abs/2601.02163}

\bibitem{evermembench}
Hu, C., Li, T., Gao, X., Chen, H., Bai, Y., Xu, D., Lin, T., Zhao, X., Li, X., Han, Y., Pei, J., Deng, Y.: Evermembench: Benchmarking long-term interactive memory in large language models (2026), \url{https://arxiv.org/abs/2602.01313}

\bibitem{jiang2025mmsearch}
Jiang, D., Zhang, R., Guo, Z., Wu, Y., Lei, J., Qiu, P., Lu, P., Chen, Z., Fu, C., Song, G., Gao, P., Liu, Y., Li, C., Li, H.: Mmsearch: Benchmarking the potential of large models as multi-modal search engines (2024), \url{https://arxiv.org/abs/2409.12959}

\bibitem{jiang2024e5vuniversalembeddingsmultimodal}
Jiang, T., Song, M., Zhang, Z., Huang, H., Deng, W., Sun, F., Zhang, Q., Wang, D., Zhuang, F.: E5-v: Universal embeddings with multimodal large language models (2024), \url{https://arxiv.org/abs/2407.12580}

\bibitem{jiang2026pixelsfactspix2factbenchmarking}
Jiang, Y., Zhang, C., Zhang, B., Yang, Y., Wang, B., Ong, Y.S.: From pixels to facts (pix2fact): Benchmarking multi-hop reasoning for fine-grained visual fact checking (2026), \url{https://arxiv.org/abs/2602.00593}

\bibitem{2405.20204}
Koukounas, A., Mastrapas, G., Günther, M., Wang, B., Martens, S., Mohr, I., Sturua, S., Akram, M.K., Martínez, J.F., Ognawala, S., Guzman, S., Werk, M., Wang, N., Xiao, H.: Jina clip: Your clip model is also your text retriever (2024), \url{https://arxiv.org/abs/2405.20204}

\bibitem{koukounas2024jinaclipv2multilingualmultimodalembeddings}
Koukounas, A., Mastrapas, G., Wang, B., Akram, M.K., Eslami, S., Günther, M., Mohr, I., Sturua, S., Martens, S., Wang, N., Xiao, H.: jina-clip-v2: Multilingual multimodal embeddings for text and images (2024), \url{https://arxiv.org/abs/2412.08802}

\bibitem{li2024mvbenchcomprehensivemultimodalvideo}
Li, K., Wang, Y., He, Y., Li, Y., Wang, Y., Liu, Y., Wang, Z., Xu, J., Chen, G., Luo, P., Wang, L., Qiao, Y.: Mvbench: A comprehensive multi-modal video understanding benchmark (2024), \url{https://arxiv.org/abs/2311.17005}

\bibitem{li2025omnibenchfutureuniversalomnilanguage}
Li, Y., Zhang, G., Ma, Y., Yuan, R., Zhu, K., Guo, H., Liang, Y., Liu, J., Wang, Z., Yang, J., Wu, S., Qu, X., Shi, J., Zhang, X., Yang, Z., Wang, X., Zhang, Z., Liu, Z., Benetos, E., Huang, W., Lin, C.: Omnibench: Towards the future of universal omni-language models (2025), \url{https://arxiv.org/abs/2409.15272}

\bibitem{li2024videovista}
Li, Y., Chen, X., Hu, B., Wang, L., Shi, H., Zhang, M.: Videovista: A versatile benchmark for video understanding and reasoning (2024), \url{https://arxiv.org/abs/2406.11303}

\bibitem{lin2025mmembeduniversalmultimodalretrieval}
Lin, S.C., Lee, C., Shoeybi, M., Lin, J., Catanzaro, B., Ping, W.: Mm-embed: Universal multimodal retrieval with multimodal llms (2025), \url{https://arxiv.org/abs/2411.02571}

\bibitem{lin2015microsoftcococommonobjects}
Lin, T.Y., Maire, M., Belongie, S., Bourdev, L., Girshick, R., Hays, J., Perona, P., Ramanan, D., Zitnick, C.L., Dollár, P.: Microsoft coco: Common objects in context (2015), \url{https://arxiv.org/abs/1405.0312}

\bibitem{liu2024mmbench}
Liu, Y., Duan, H., Zhang, Y., Li, B., Zhang, S., Zhao, W., Yuan, Y., Wang, J., He, C., Liu, Z., Chen, K., Lin, D.: Mmbench: Is your multi-modal model an all-around player? (2024), \url{https://arxiv.org/abs/2307.06281}

\bibitem{liu2025ola}
Liu, Z., Dong, Y., Wang, J., Liu, Z., Hu, W., Lu, J., Rao, Y.: Ola: Pushing the frontiers of omni-modal language model (2025), \url{https://arxiv.org/abs/2502.04328}

\bibitem{luo2025globalretrievalaugmentedgeneration}
Luo, Q., Li, X., Fan, T., Chen, X., Qiu, X.: Towards global retrieval augmented generation: A benchmark for corpus-level reasoning (2025), \url{https://arxiv.org/abs/2510.26205}

\bibitem{ma2024mmlongbenchdoc}
Ma, Y., Zang, Y., Chen, L., Chen, M., Jiao, Y., Li, X., Lu, X., Liu, Z., Ma, Y., Dong, X., Zhang, P., Pan, L., Jiang, Y.G., Wang, J., Cao, Y., Sun, A.: Mmlongbench-doc: Benchmarking long-context document understanding with visualizations (2024), \url{https://arxiv.org/abs/2407.01523}

\bibitem{mangalam2023egoschemadiagnosticbenchmarklongform}
Mangalam, K., Akshulakov, R., Malik, J.: Egoschema: A diagnostic benchmark for very long-form video language understanding (2023), \url{https://arxiv.org/abs/2308.09126}

\bibitem{marino2019okvqavisualquestionanswering}
Marino, K., Rastegari, M., Farhadi, A., Mottaghi, R.: Ok-vqa: A visual question answering benchmark requiring external knowledge (2019), \url{https://arxiv.org/abs/1906.00067}

\bibitem{Mathew2020DocVQAAD}
Mathew, M., Karatzas, D., Jawahar, C.V.: Docvqa: A dataset for vqa on document images (2021), \url{https://arxiv.org/abs/2007.00398}

\bibitem{meng2025mmiu}
Meng, F., Wang, J., Li, C., Lu, Q., Tian, H., Liao, J., Zhu, X., Dai, J., Qiao, Y., Luo, P., Zhang, K., Shao, W.: Mmiu: Multimodal multi-image understanding for evaluating large vision-language models (2024), \url{https://arxiv.org/abs/2408.02718}

\bibitem{nussbaum2024nomic}
Nussbaum, Z., Duderstadt, B., Mulyar, A.: Nomic embed vision: Expanding the latent space (2024), \url{https://arxiv.org/abs/2406.18587}

\bibitem{openai2025-gpt5}
{OpenAI}: Introducing {GPT-5} (2025)

\bibitem{retvqa}
Penamakuri, A.S., Gupta, M., Gupta, M.D., Mishra, A.: Answer mining from a pool of images: Towards retrieval-based visual question answering (2023), \url{https://arxiv.org/abs/2306.16713}

\bibitem{peng2026unidocbenchunifiedbenchmarkdocumentcentric}
Peng, X., Qin, C., Chen, Z., Xu, R., Xiong, C., Wu, C.S.: Unidoc-bench: A unified benchmark for document-centric multimodal rag (2026), \url{https://arxiv.org/abs/2510.03663}

\bibitem{piergiovanni2022videoquestionansweringiterative}
Piergiovanni, A., Morton, K., Kuo, W., Ryoo, M.S., Angelova, A.: Video question answering with iterative video-text co-tokenization (2022), \url{https://arxiv.org/abs/2208.00934}

\bibitem{plummer2016flickr30kentitiescollectingregiontophrase}
Plummer, B.A., Wang, L., Cervantes, C.M., Caicedo, J.C., Hockenmaier, J., Lazebnik, S.: Flickr30k entities: Collecting region-to-phrase correspondences for richer image-to-sentence models (2016), \url{https://arxiv.org/abs/1505.04870}

\bibitem{radford2021learning}
Radford, A., Kim, J.W., Hallacy, C., Ramesh, A., Goh, G., Agarwal, S., Sastry, G., Askell, A., Mishkin, P., Clark, J., Krueger, G., Sutskever, I.: Learning transferable visual models from natural language supervision (2021), \url{https://arxiv.org/abs/2103.00020}

\bibitem{schwenk2022okvqa}
Schwenk, D., Khandelwal, A., Clark, C., Marino, K., Mottaghi, R.: A-okvqa: A benchmark for visual question answering using world knowledge (2022), \url{https://arxiv.org/abs/2206.01718}

\bibitem{shen2025rightwayassessingdocument}
Shen, W., Wang, M., Wang, Y., Chen, D., Yang, J., Wan, Y., Lin, W.: Are we on the right way for assessing document retrieval-augmented generation? (2025), \url{https://arxiv.org/abs/2508.03644}

\bibitem{singh2019towards}
Singh, A., Natarajan, V., Shah, M., Jiang, Y., Chen, X., Batra, D., Parikh, D., Rohrbach, M.: Towards vqa models that can read (2019), \url{https://arxiv.org/abs/1904.08920}

\bibitem{tschannen2025siglip2multilingualvisionlanguage}
Tschannen, M., Gritsenko, A., Wang, X., Naeem, M.F., Alabdulmohsin, I., Parthasarathy, N., Evans, T., Beyer, L., Xia, Y., Mustafa, B., Hénaff, O., Harmsen, J., Steiner, A., Zhai, X.: Siglip 2: Multilingual vision-language encoders with improved semantic understanding, localization, and dense features (2025), \url{https://arxiv.org/abs/2502.14786}

\bibitem{wang2024mmneedle}
Wang, H., Shi, H., Tan, S., Qin, W., Wang, W., Zhang, T., Nambi, A., Ganu, T., Wang, H.: Multimodal needle in a haystack: Benchmarking long-context capability of multimodal large language models (2025), \url{https://arxiv.org/abs/2406.11230}

\bibitem{Qwen2VL}
Wang, P., Bai, S., Tan, S., Wang, S., Fan, Z., Bai, J., Chen, K., Liu, X., Wang, J., Ge, W., Fan, Y., Dang, K., Du, M., Ren, X., Men, R., Liu, D., Zhou, C., Zhou, J., Lin, J.: Qwen2-vl: Enhancing vision-language model's perception of the world at any resolution (2024), \url{https://arxiv.org/abs/2409.12191}

\bibitem{wang2024needle}
Wang, W., Zhang, S., Ren, Y., Duan, Y., Li, T., Liu, S., Hu, M., Chen, Z., Zhang, K., Lu, L., Zhu, X., Luo, P., Qiao, Y., Dai, J., Shao, W., Wang, W.: Needle in a multimodal haystack (2024), \url{https://arxiv.org/abs/2406.07230}

\bibitem{wu2024longvideobenchbenchmarklongcontextinterleaved}
Wu, H., Li, D., Chen, B., Li, J.: Longvideobench: A benchmark for long-context interleaved video-language understanding (2024), \url{https://arxiv.org/abs/2407.15754}

\bibitem{yang2022learninganswervisualquestions}
Yang, A., Miech, A., Sivic, J., Laptev, I., Schmid, C.: Learning to answer visual questions from web videos (2022), \url{https://arxiv.org/abs/2205.05019}

\bibitem{wikiautogen}
Yang, Z., Chen, J., Xu, D., Fei, J., Shen, X., Zhao, L., Feng, C.M., Elhoseiny, M.: Wikiautogen: Towards multi-modal wikipedia-style article generation (2025), \url{https://arxiv.org/abs/2503.19065}

\bibitem{XR}
Yang, Z., Pang, W., Yuan, Y.: Xr: Cross-modal agents for composed image retrieval. ArXiv  \textbf{abs/2601.14245} (2026), \url{https://api.semanticscholar.org/CorpusID:284912133}

\bibitem{mermaid}
Yang, Z., Song, J., Song, S., Pang, W., Yuan, Y.: {MERMAID}: Multi-perspective self-reflective agents with generative augmentation for emotion recognition. In: Christodoulopoulos, C., Chakraborty, T., Rose, C., Peng, V. (eds.) Proceedings of the 2025 Conference on Empirical Methods in Natural Language Processing. pp. 24639--24655. Association for Computational Linguistics, Suzhou, China (Nov 2025). \doi{10.18653/v1/2025.emnlp-main.1252}, \url{https://aclanthology.org/2025.emnlp-main.1252/}

\bibitem{inex}
Yang, Z., Yuan, Y., Jiang, X., An, B., Pang, W.: Inex: Hallucination mitigation via introspection and cross-modal multi-agent collaboration (2025), \url{https://arxiv.org/abs/2512.02981}

\bibitem{yang2025longvt}
Yang, Z., Wang, S., Zhang, K., Wu, K., Leng, S., Zhang, Y., Li, B., Qin, C., Lu, S., Li, X., Bing, L.: Longvt: Incentivizing "thinking with long videos" via native tool calling (2025), \url{https://arxiv.org/abs/2511.20785}

\bibitem{ying2026retrievalinfusedreasoningsandboxbenchmark}
Ying, S., Wang, Z., Peng, Y., Chen, J., Wu, Y., Lin, H., He, D., Liu, S., Yu, G., Piao, Y., Wu, Y., Gui, X., Peng, Z., Li, X., Du, X., Qin, L., Cao, Y., Zhang, G., Huang, S.: Retrieval-infused reasoning sandbox: A benchmark for decoupling retrieval and reasoning capabilities (2026), \url{https://arxiv.org/abs/2601.21937}

\bibitem{yu2025visragvisionbasedretrievalaugmentedgeneration}
Yu, S., Tang, C., Xu, B., Cui, J., Ran, J., Yan, Y., Liu, Z., Wang, S., Han, X., Liu, Z., Sun, M.: Visrag: Vision-based retrieval-augmented generation on multi-modality documents (2025), \url{https://arxiv.org/abs/2410.10594}

\bibitem{zhai2023sigmoid}
Zhai, X., Mustafa, B., Kolesnikov, A., Beyer, L.: Sigmoid loss for language image pre-training (2023), \url{https://arxiv.org/abs/2303.15343}

\bibitem{zhang2025openmmreasoner}
Zhang, K., Wu, K., Yang, Z., Li, B., Hu, K., Wang, B., Liu, Z., Li, X., Bing, L.: Openmmreasoner: Pushing the frontiers for multimodal reasoning with an open and general recipe (2025), \url{https://arxiv.org/abs/2511.16334}

\bibitem{Zhu2025InternVL3EA}
Zhu, J., Wang, W., Chen, Z., Liu, Z., Ye, S., Gu, L., Tian, H., Duan, Y., Su, W., Shao, J., Gao, Z., Cui, E., Wang, X., Cao, Y., Liu, Y., Wei, X., Zhang, H., Wang, H., Xu, W., Li, H., Wang, J., Deng, N., Li, S., He, Y., Jiang, T., Luo, J., Wang, Y., He, C., Shi, B., Zhang, X., Shao, W., He, J., Xiong, Y., Qu, W., Sun, P., Jiao, P., Lv, H., Wu, L., Zhang, K., Deng, H., Ge, J., Chen, K., Wang, L., Dou, M., Lu, L., Zhu, X., Lu, T., Lin, D., Qiao, Y., Dai, J., Wang, W.: Internvl3: Exploring advanced training and test-time recipes for open-source multimodal models (2025), \url{https://arxiv.org/abs/2504.10479}

\end{thebibliography}

\newpage
\appendix
\section*{Appendix}
% \tableofcontents
% \clearpage

\renewcommand{\thefigure}{A.\arabic{figure}}
\renewcommand{\thetable}{A.\arabic{table}}
\setcounter{figure}{0}
\setcounter{table}{0}
\section{Statistics}
\label{sec:stat}
\begin{figure}[htbp]
    \centering
    \includegraphics[width=\linewidth]{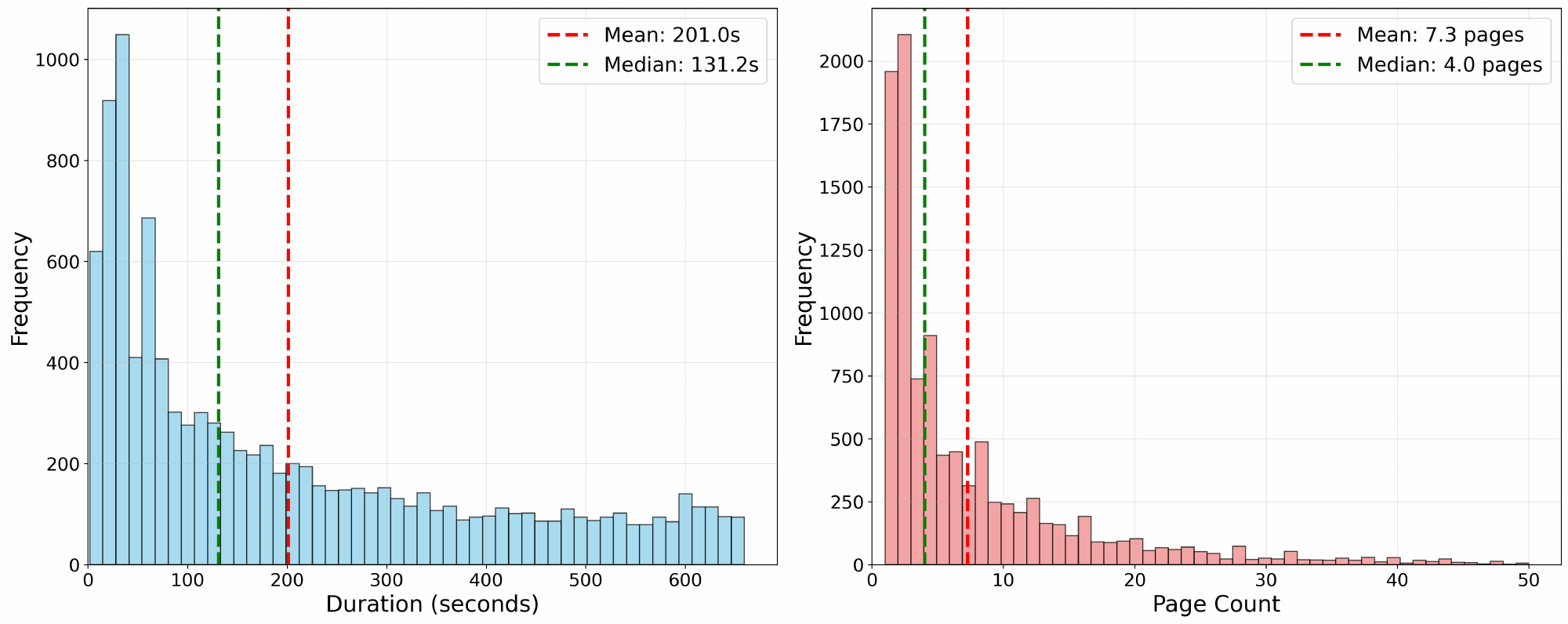}
    \caption{\textbf{Video–Document Distribution Overview.} Distributions of video duration (left) and document page count (right), with red dashed lines indicating means and green dashed lines indicating medians.}
    \label{fig:video_pdf_sta}
\end{figure}

\Cref{fig:video_pdf_sta} provides a corpus-level overview of sample lengths for two modalities in our benchmark: video (seconds) and document (pages), with means and medians annotated for reference. Both distributions are distinctly right-skewed, with many short items and a non-trivial long tail—statistics that mirror real-world multimedia collections and that are particularly relevant for retrieval under variable context sizes. This heterogeneity ensures that systems are evaluated on both rapid evidence localization in concise items and robust reasoning over extended content. The image modality comprises atomic, single-frame items and therefore has no analogous length measure. We report these statistics to characterize the benchmark and to contextualize evaluation difficulty, facilitating reproducibility and fair comparison across methods.

\renewcommand{\thefigure}{B.\arabic{figure}}
\renewcommand{\thetable}{B.\arabic{table}}
\setcounter{figure}{0}
\setcounter{table}{0}
\section{Examples from MultiHaystack}
\label{sec:appendix_example}

To illustrate the diverse and complex nature of the MultiHaystack benchmark, we present representative examples across video, image, and document modalities, including data-enriched cases. Each instance is designed for retrieval-augmented reasoning at scale, emphasizing both modality-specific understanding and fine-grained grounding.

These examples demonstrate that MultiHaystack provides a comprehensive and rigorous benchmark for cross-modal retrieval and reasoning, capturing both perceptual diversity and semantic nuance under realistic large-scale conditions.

\subsection{Modality Examples}

\subsubsection{Video}
\begin{figure}[htbp]
    \centering
    \includegraphics[width=0.8\linewidth]{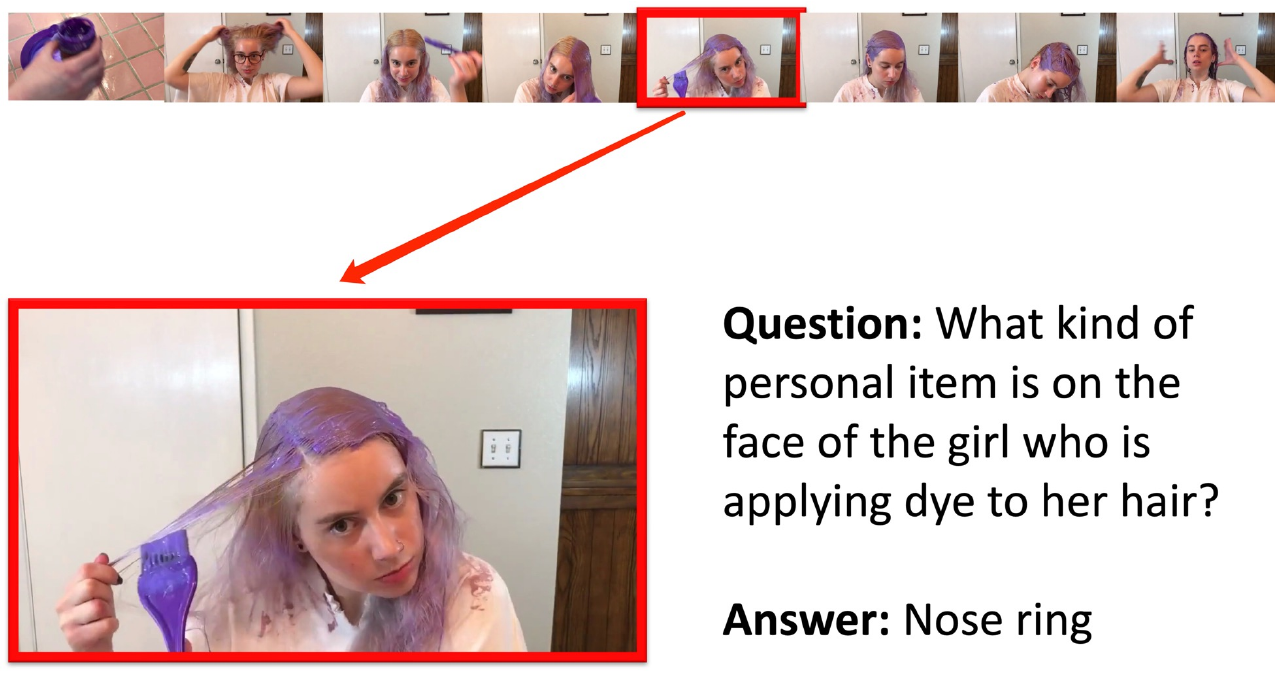}
    \caption{\textbf{Video Example 1.}}
    \label{fig:video1}
\end{figure}

\begin{figure}[htbp]
    \centering
    \includegraphics[width=0.8\linewidth]{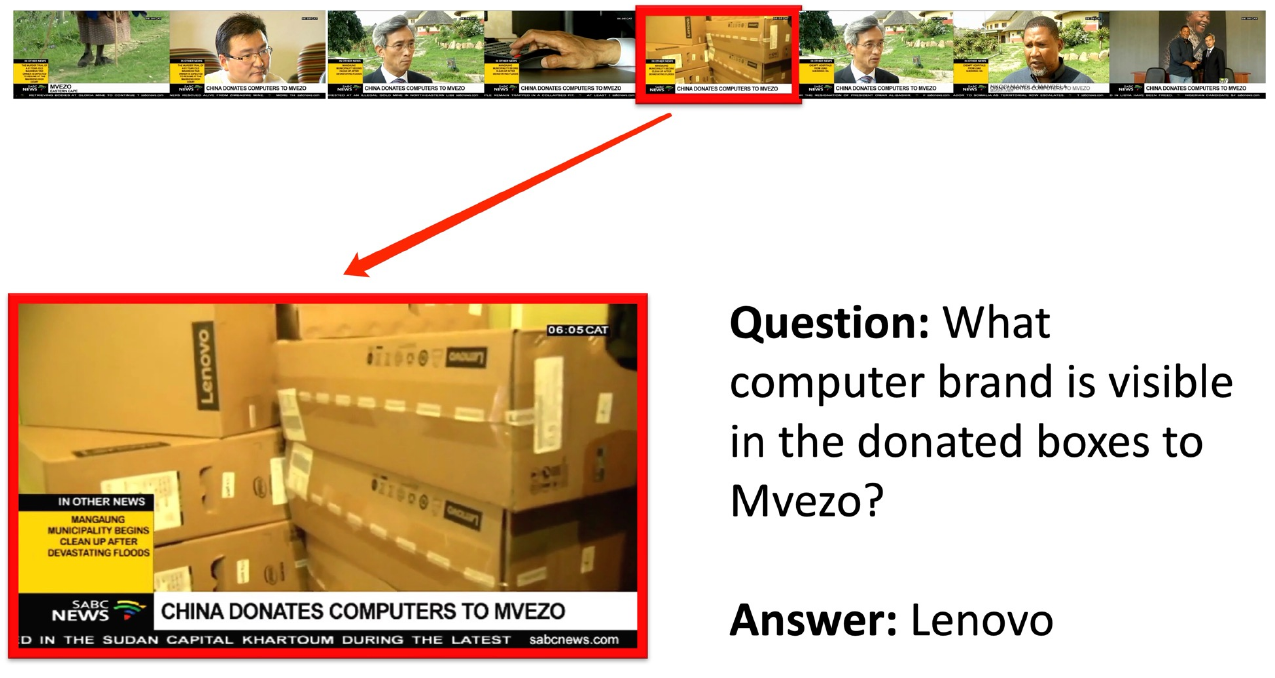}
    \caption{\textbf{Video Example 2.}}
    \label{fig:video2}
\end{figure}

\begin{figure}[htbp]
    \centering
    \includegraphics[width=0.8\linewidth]{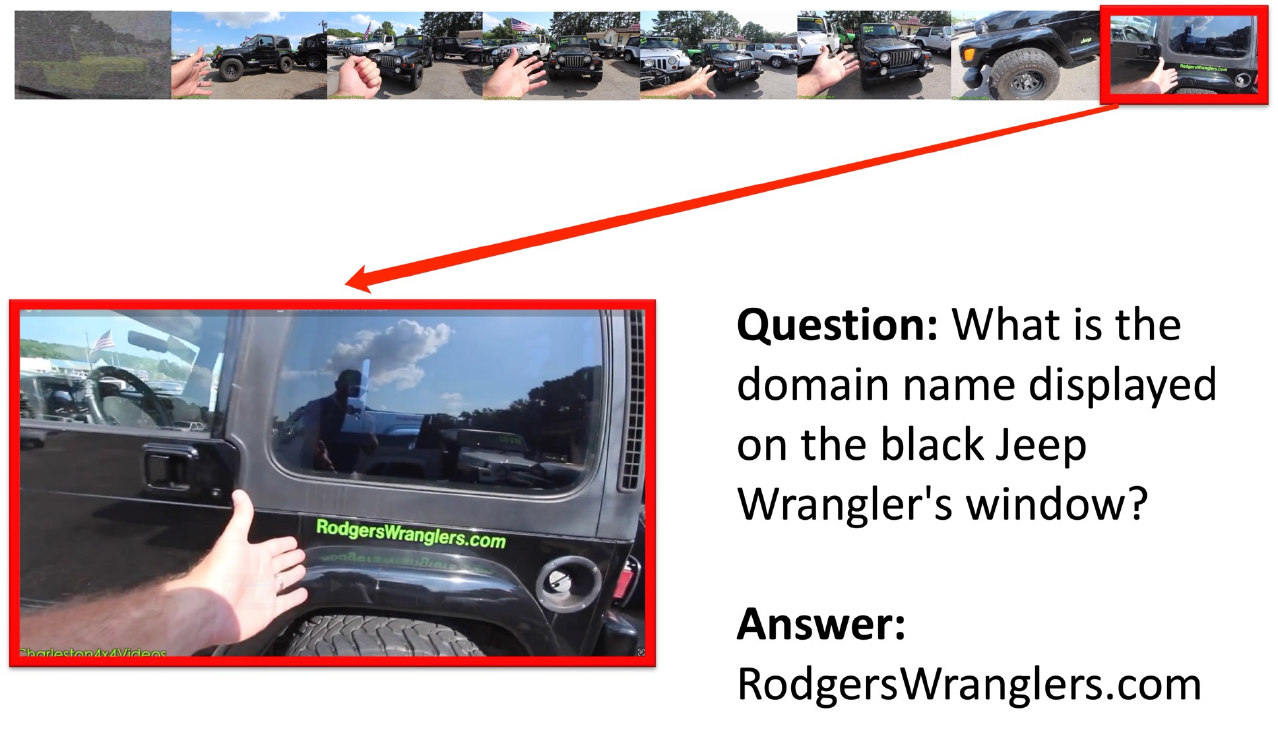}
    \caption{\textbf{Video Example 3.}}
    \label{fig:video3}
\end{figure}

\textbf{Video-based QA} often requires modeling temporal dynamics, capturing frame-level details, and leveraging embedded textual cues. 
For instance, in \cref{fig:video1}, the model must detect a small facial accessory (a nose ring) while the subject applies hair dye, illustrating the need for fine-grained perceptual grounding under distracting context. 
\Cref{fig:video2} demands recognition of a brand logo in a low-resolution news segment, testing robustness to visual degradation. 
\Cref{fig:video3} evaluates domain-level inference from motion-blurred frames, where temporal context must compensate for reduced visual clarity. 
Together, these tasks highlight the dual challenges of temporal sensitivity and perceptual precision in video retrieval.

\subsubsection{Image}
\begin{figure}[htbp]
    \centering
    \includegraphics[width=0.8\linewidth]{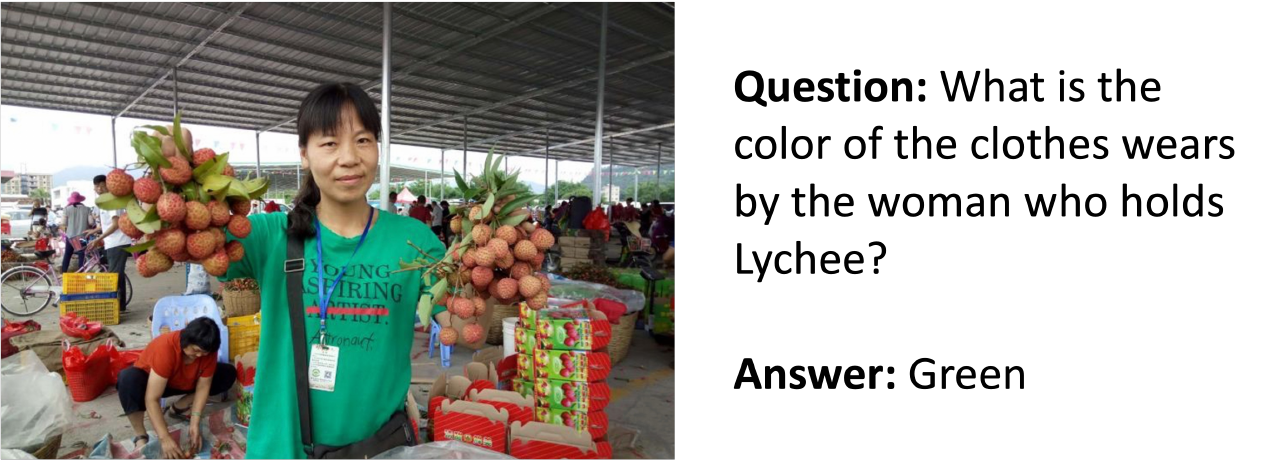}
    \caption{\textbf{Image Example 1.}}
    \label{fig:image1}
\end{figure}

\begin{figure}[htbp]
    \centering
    \includegraphics[width=0.8\linewidth]{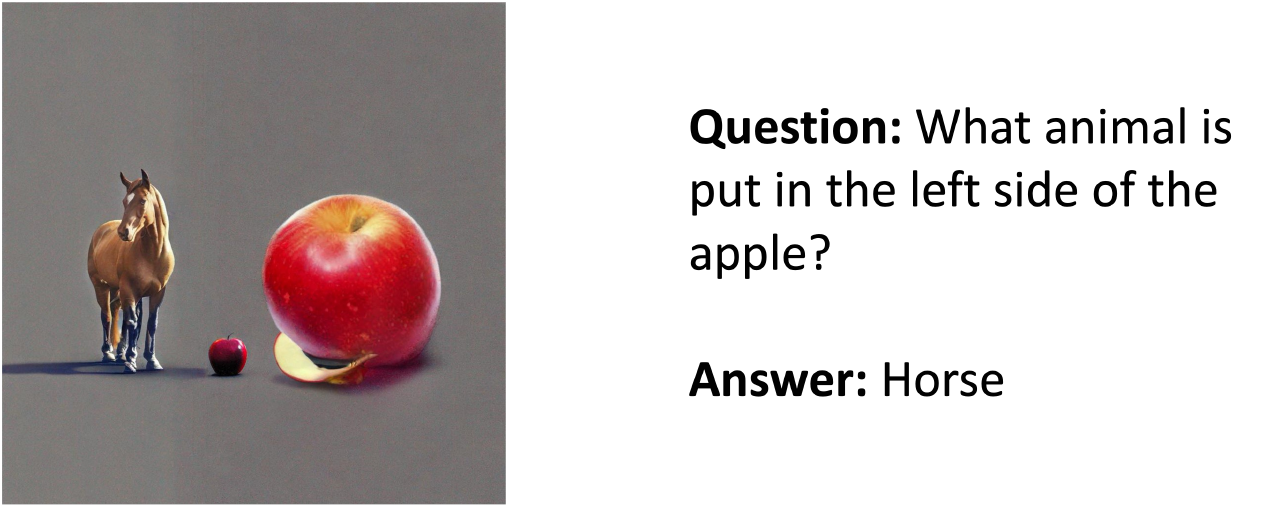}
    \caption{\textbf{Image Example 2.}}
    \label{fig:image2}
\end{figure}

\begin{figure}[htbp]
    \centering
    \includegraphics[width=0.8\linewidth]{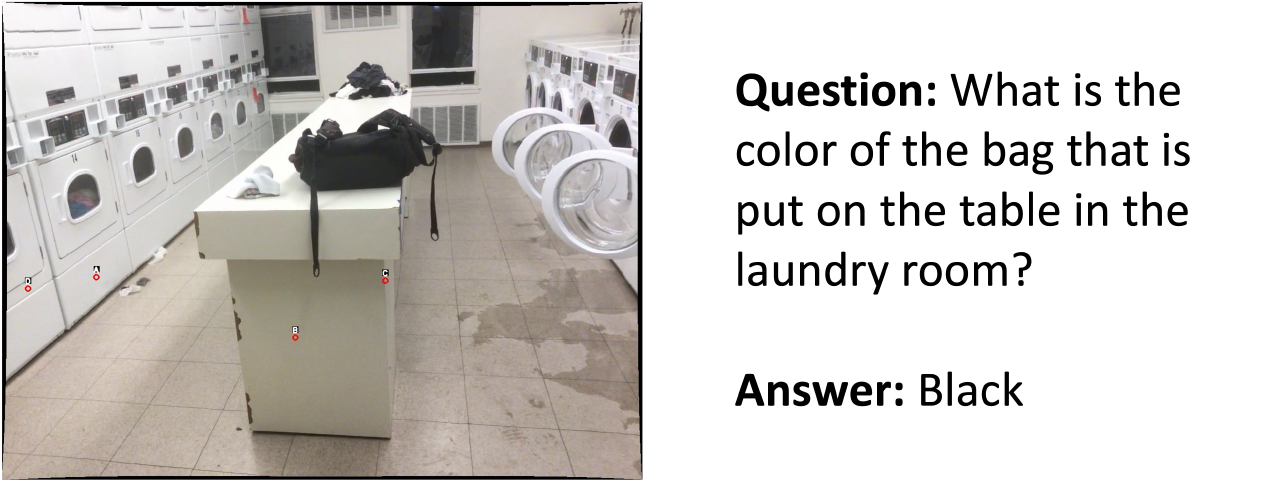}
    \caption{\textbf{Image Example 3.}}
    \label{fig:image3}
\end{figure}

\textbf{Image-based QA} emphasizes spatial understanding and localized recognition. As shown in \cref{fig:image1}, the model must infer color attributes from real-world marketplace settings. \Cref{fig:image2} requires recognizing small object co-occurrence (a horse next to an apple), while \cref{fig:image3} focuses on identifying object properties (a black bag in a laundry room). These examples test the model's capability to reason over everyday scenes with high visual clutter and subtle semantic cues.

\newpage

\subsubsection{Document}

\begin{figure}[htbp]
    \centering
    \includegraphics[width=0.9\linewidth]{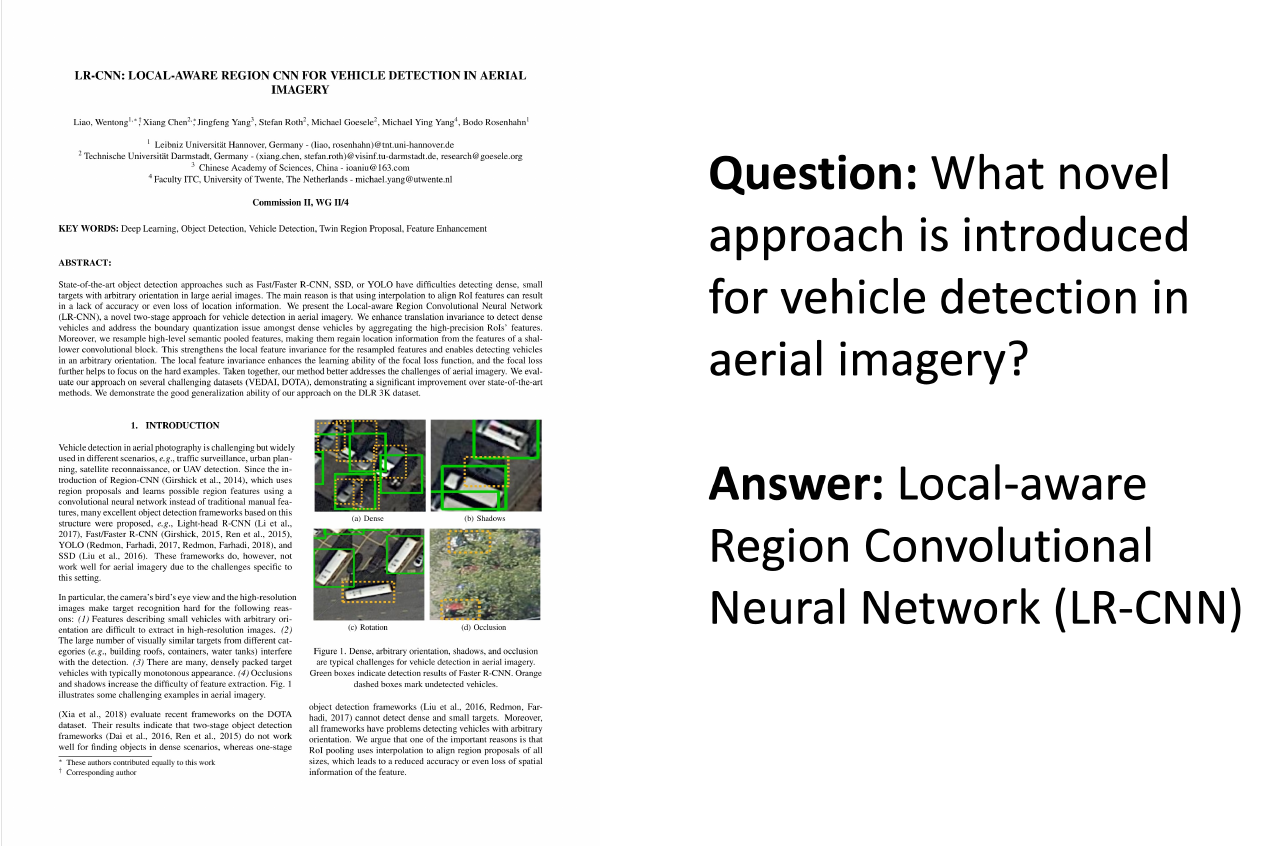}
    \caption{\textbf{Document Example 1.}}
    \label{fig:doc1}
\end{figure}

\begin{figure}[htbp]
    \centering
    \includegraphics[width=0.9\linewidth]{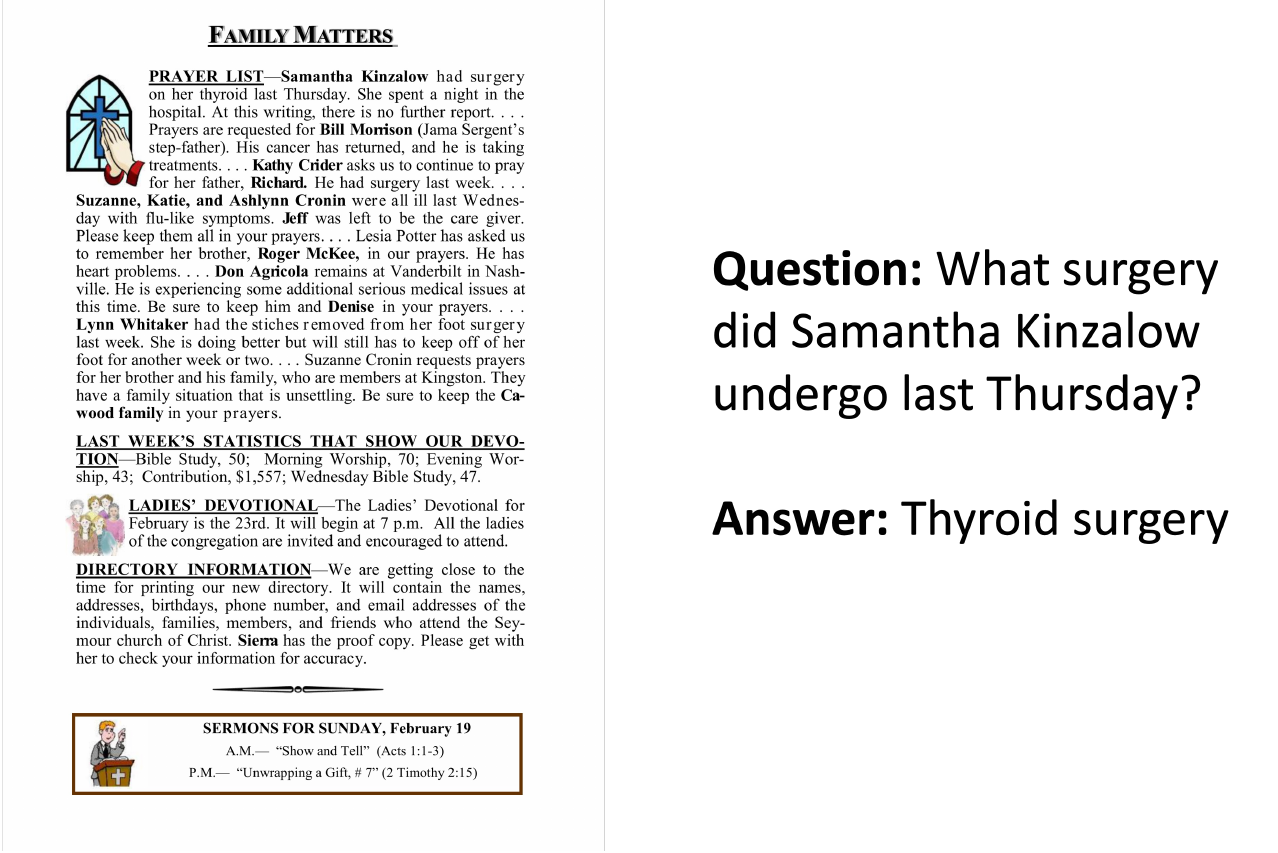}
    \caption{\textbf{Document Example 2.}}
    \label{fig:doc2}
\end{figure}

\begin{figure}[htbp]
    \centering
    \includegraphics[width=\linewidth]{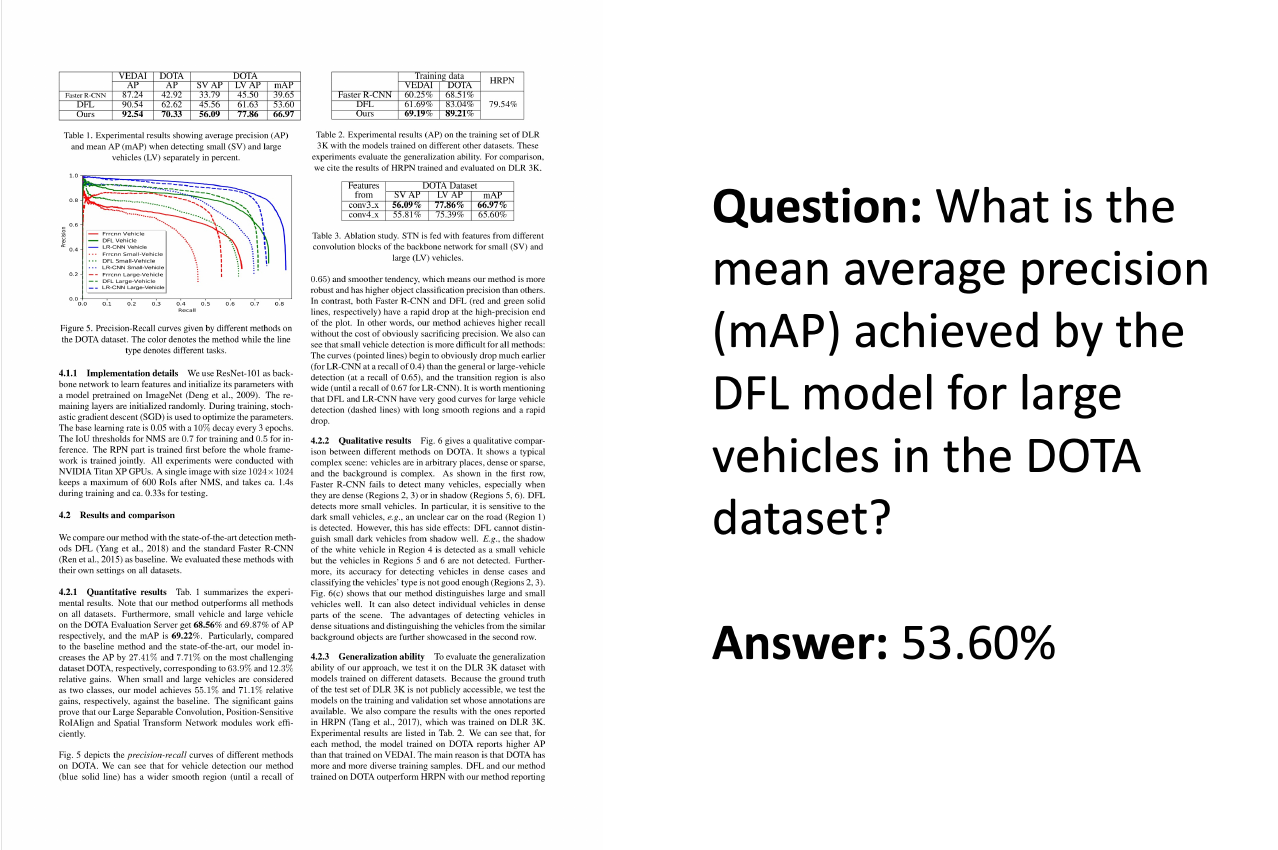}
    \caption{\textbf{Document Example 3.}}
    \label{fig:doc3}
\end{figure}

\textbf{Document-based QA} requires both visual–textual alignment and structured content reasoning. 
In \cref{fig:doc1}, the model must locate and integrate a technical concept introduced jointly in scientific text and figures, demanding precise cross-modal grounding. 
\Cref{fig:doc2} involves extracting factual content from narrative passages, testing robustness to linguistic variability and contextual dependencies. 
\Cref{fig:doc3} requires retrieving quantitative results (e.g., mean average precision) from densely packed tables, highlighting the difficulty of parsing layout-dependent numerical data. 
Together, these tasks illustrate the need for accurate text extraction, layout-aware reasoning, and fine-grained multimodal understanding in document QA.

\newpage

\subsection{Data Enrichment Examples}

\begin{figure}[htbp]
    \centering
    \includegraphics[width=0.8\linewidth]{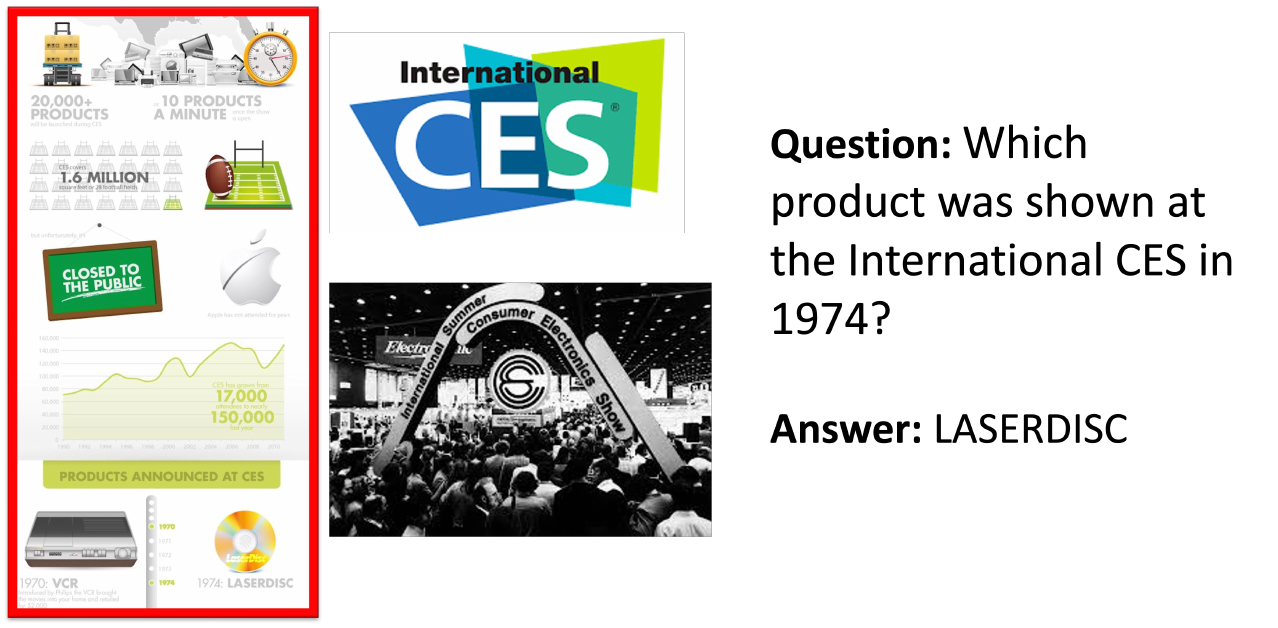}
    \caption{\textbf{Data Enrichment Example 1.}}
    \label{fig:distractor1}
\end{figure}

\begin{figure}[htbp]
    \centering
    \includegraphics[width=0.8\linewidth]{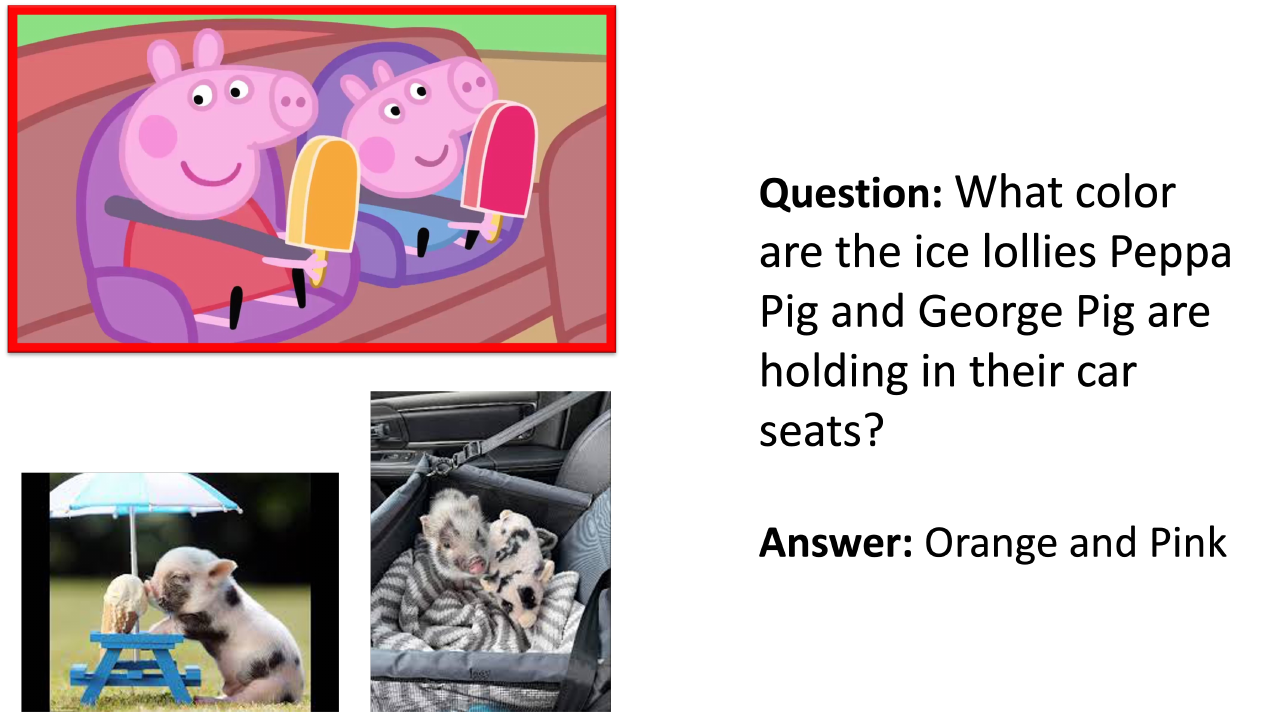}
    \caption{\textbf{Data Enrichment Example 2.}}
    \label{fig:distractor2}
\end{figure}

\begin{figure}[htbp]
    \centering
    \includegraphics[width=\linewidth]{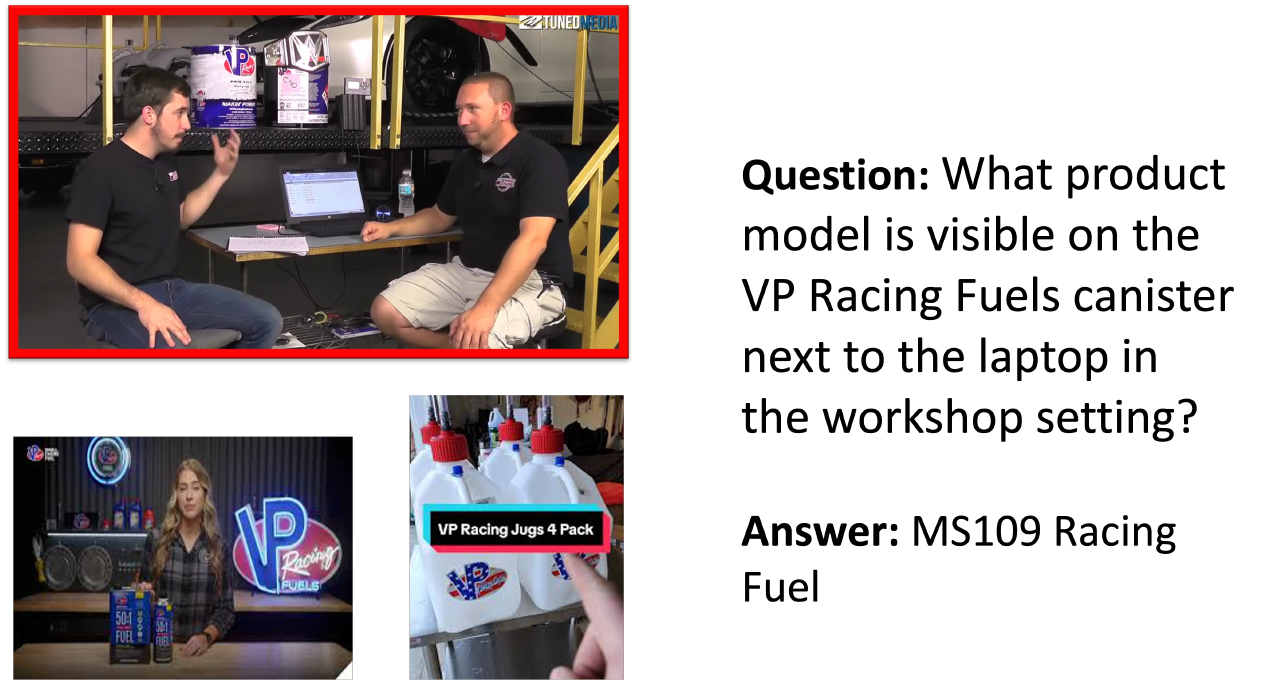}
    \caption{\textbf{Data Enrichment Example 3.}}
    \label{fig:distractor3}
\end{figure}

In addition to ground-truth sources, MultiHaystack incorporates \textbf{data-enriched contrastive examples} that bear strong semantic or visual similarity to the correct content but do not contain the target answer. The inclusion of these examples is motivated by the need to reflect the inherent ambiguity present in real-world retrieval scenarios, where multiple plausible candidates often appear contextually relevant despite being incorrect. Rather than artificially introducing noise, these examples are carefully selected based on contextual coherence and fine-grained resemblance, ensuring that they remain informative and challenging.
As illustrated in ~\cref{fig:distractor1}--\ref{fig:distractor3}, these contrastive examples are constructed to simulate realistic retrieval confusion without relying on synthetic perturbations. For instance, \cref{fig:distractor1} presents an electronics-related scene that is temporally close to the target reference but does not include the specified CES product. \Cref{fig:distractor2} shows a visually similar cartoon frame that lacks the queried object state. \Cref{fig:distractor3} depicts a relevant industrial setting, yet omits the specific label required by the question.

This design aligns closely with the task types defined in \cref{fig:task-examples} in the main text, particularly those requiring contextual understanding, visual parsing, and metadata identification. In these tasks, distinguishing semantically proximate yet incomplete candidates is essential for accurate reasoning. Moreover, such contrastive examples mirror the uncertainty faced in open-domain QA systems, where models must search over large corpora containing numerous partially relevant documents. 
By introducing semantically aligned but unanswerable instances, MultiHaystack encourages precise grounding and discourages superficial similarity matching, thereby offering a more faithful evaluation of retrieval and reasoning capabilities in realistic multimodal settings.

\newpage
\renewcommand{\thefigure}{C.\arabic{figure}}
\renewcommand{\thetable}{C.\arabic{table}}
\setcounter{figure}{0}
\setcounter{table}{0}
\section{Prompts}
\label{sec:prompt}
In this section, we present the prompts for data construction and evaluation: one enforces precise, unambiguous QA generation, and the other defines a binary protocol for judging predictions, together ensuring reliable and reproducible assessment.

\begin{ttcolorbox}[QA Generation Prompt]
\texttt{You are an expert at generating specific and precisely targeted questions based on a series of sequentially provided images. These images have been sampled in sequence from a video or a document and typically contain information-rich visuals.}\\
\texttt{Your Task: \\
Carefully analyze all provided images.\\
Generate exactly 30 distinct, highly-specific questions, each accompanied by its correct answer explicitly based on information visible in these images. \\
Strict Requirements for Question Generation: \\
DO NOT reference the images themselves or use vague positional indicators such as: \\
'in the image,' 'in the first/second/third image,' 'in the provided picture,' 'at the top/bottom,' 'this slide,' 'the table above/below,' etc. \\
Instead, always clearly and explicitly include specific details directly from the visuals, such as: \\
Exact names (persons, companies, products) \\
Precise numerical values (financial figures, percentages) \\
Exact dates or years explicitly mentioned \\
Specific places, titles, labels, captions, or identifiable entities clearly visible and named in the visuals. \\
Each question must stand alone as a fully self-contained, specific question that would allow someone seeing it later to precisely identify and locate the correct information within related textual documents or sources, without needing visual context. \\
Clear Examples of Correct Question Format (follow exactly this style of specificity): \\
"What is the 'net earnings' of Johnson \& Johnson and subsidiaries in the year 2009?" \\
"What was the 'gross profit' reported by Johnson \& Johnson and subsidiaries for the fiscal year 2010?" \\
Explicitly Prohibited Example (Do NOT do this): \\
Incorrect: "What are the two cats interacting with on the wooden floor in the second image?" (Reason: includes prohibited phrase 'in the second image') \\
Corrected: "What object are the two cats interacting with on the wooden floor next to the white sofa?" (Reason: explicitly references visual details, removing vague positional indicators.) \\
Final Output Requirements: \\
Generate exactly 30 questions with their correct, concise answers based explicitly on the details shown in the provided series of images. \\
Ensure every question strictly follows the specificity rules described above, completely eliminating unspecific or ambiguous references to images or their positions.}
\end{ttcolorbox}

\begin{ttcolorbox}[LLM Judgement Prompt]
\texttt{You are an evaluator. Compare the Predicted Answer with the True Answer and determine if the Predicted Answer is Correct or Incorrect.}\\
\texttt{Instructions: \\
1. If the Predicted Answer provides the same information or a reasonable interpretation of the True Answer, respond with 'Correct.' \\
2. If the Predicted Answer does not match or does not reasonably interpret the True Answer, respond with 'Incorrect.' \\
Important: Answer only with 'Correct' or 'Incorrect' - no explanations.}
\end{ttcolorbox}

\renewcommand{\thefigure}{D.\arabic{figure}}
\renewcommand{\thetable}{D.\arabic{table}}
\setcounter{figure}{0}
\setcounter{table}{0}
\section{Reproducibility}
\label{sec:reproduce}

\subsection{Implementation details} 
\noindent \textbf{Parameter settings}
Across all experiments, the language model temperature was fixed at 0.4. 
During data enrichment, we employed CLIP-based filtering of web-retrieved images, retaining an image as a candidate distractor only if its cosine similarity with the corresponding query exceeded 0.2. 
For the VQA experiments, retrieval used a fixed top-k setting with k=5. In addition, the VQA pipeline implemented an automatic retry mechanism to improve robustness: if an error occurred at any stage, the procedure was retried up to three times before being marked as failed.

\noindent \textbf{Implementation Environment}
All experiments were executed on a single NVIDIA H100 GPU (80 GB HBM3). The software stack comprised Python 3.12, PyTorch 2.6.0, and Hugging Face Transformers 4.51.0. Unless otherwise specified, inference was performed in bfloat16 (bf16) precision. These version details are reported to facilitate reproducibility.

\noindent \textbf{Task Distribution.} 
The task distribution in Figure~2 is deliberately designed rather than sampled from real-world frequencies. Our core motivation is to build a diagnostic benchmark: real-world distributions are long-tailed and dominated by easy perceptual queries, which severely under-represent harder reasoning types such as statistical analysis or metadata identification. If we followed such organic distributions, aggregate benchmark scores would largely reflect surface-level perception skills while masking weaknesses in deeper reasoning, thereby limiting the benchmark’s value for research. To address this, we enforce a balanced coverage across six categories: (i) Visual Parsing and Positioning, targeting spatial localization and object layout; (ii) Contextual Understanding, focusing on embedded text and local semantics; (iii) Video Temporal Reasoning, requiring comprehension of motion and temporal order; (iv) Statistical Reasoning, evaluating quantitative analysis of tables and charts; (v) Metadata Identification, stressing recognition of affiliations, timestamps, and sources; and (vi) Factual Knowledge Retrieval, ensuring grounding in corpus-level factual evidence. These categories were carefully chosen to span perceptual and analytical dimensions, covering the dominant reasoning skills demanded in real-world multimodal applications. By balancing across them, the benchmark ensures fair and reproducible evaluation, highlights fine-grained strengths and weaknesses of models, and provides a controlled yet realistic setting to stress-test multimodal retrieval and reasoning capabilities.

\subsection{Usage Benchmarks}

\begin{itemize}
  \item \textbf{VideoVista} ~\cite{li2024videovista} is a comprehensive video question answering benchmark with 24,906 multiple choice questions built from 3,402 YouTube videos across 14 categories, spanning a few seconds to over 10 minutes and covering 27 task types for understanding and reasoning. It is constructed via an automated pipeline that uses GPT-4o with video splitting, object segmentation, tracking, OCR, and ASR, followed by targeted human checks to ensure quality. Evaluations show persistent challenges in fine-grained temporal localization, anomaly detection, and relational and logical reasoning.

  \item \textbf{MMBench-Video} ~\cite{fang2024mmbenchvideolongformmultishotbenchmark} is a long-form, multi-shot VideoQA benchmark designed to holistically assess LVLMs’ spatial and temporal understanding across real-world web videos. It comprises 609 YouTube clips (30s–6min) spanning 16 categories and 1,998 human-authored, free-form QAs annotated under a 3-level taxonomy covering 26 fine-grained capabilities, with deliberate emphasis on temporal indispensability. The benchmark pairs open-ended evaluation with a GPT-4–based judging scheme to improve robustness and alignment with human preferences, and we report comprehensive comparisons of open-source and proprietary models. Code and evaluation are integrated into VLMEvalKit, providing a practical, scalable resource for advancing video understanding research.

  \item \textbf{FineVideo} ~\cite{Farre2024FineVideo} is a large-scale dataset for multimodal video understanding that targets the hard problems of mood analysis, narrative structure, and media editing. Spanning 43,751 YouTube videos (~3,425 hours; avg. 4.7 minutes) across 122 categories, it couples raw video with time-coded speech-to-text and rich, scene-level annotations—characters, activities, props, editing cues, audiovisual correlation, narrative progression, and emotional trajectories. This fine-grained supervision enables both pretraining and task-specific fine-tuning for context-savvy video models.

  \item \textbf{MVBench} ~\cite{li2024mvbenchcomprehensivemultimodalvideo} is a comprehensive benchmark for temporal video understanding in MLLMs, defining 20 temporally grounded tasks by transforming static image tasks into their dynamic video counterparts. Multiple-choice questions are automatically generated from annotations across 11 public video datasets to ensure objective, reproducible scoring. Initial evaluations reveal considerable headroom for temporal reasoning, with the VideoChat2 baseline substantially outperforming prior models, establishing MVBench as a standardized, motion-aware testbed spanning perception through cognition.

  \item \textbf{DocHaystack} ~\cite{chen2024document} is the large-scale benchmark for vision language reasoning that pairs each question with up to 1000 visual documents and requires a single document-grounded answer. Built from DocVQA and InfographicVQA using a pipeline that combines LLM filtering, human review, and removal of generic knowledge questions, they better reflect real retrieval needs at scale. The suite offers 100, 200, and 1000 document settings for joint evaluation of retrieval and VQA, with Recall at k used to assess retrieval quality. 

    \item \textbf{MMIU} ~\cite{meng2025mmiu} is a comprehensive multi-image benchmark for evaluating large vision–language models, spanning 7 inter-image relationship types and 52 tasks built over 77,659 images and 11,698 carefully curated multiple-choice questions across five modalities, with an explicit unanswerable set for robustness analysis. Designed via a top-down hierarchy inspired by cognitive psychology, MMIU supports fine-grained diagnosis of semantic, temporal, and spatial reasoning, offers task-map analyses to distinguish in- vs. out-of-domain skills, and provides SFT-based difficulty estimates to guide model and data improvement.

    \item \textbf{A-OKVQA} ~\cite{schwenk2022okvqa} is a knowledge-intensive VQA benchmark built on COCO-2017 that comprises 24,903 question–answer–rationale triplets with train/val/test splits preserved, targeting reasoning that combines visual understanding with commonsense, factual, and physical world knowledge rather than simple lookup. Each item includes multiple-choice options and ten free-form answers, enabling both MC and Direct Answer evaluation, while human-written rationales (three per question) support training and analysis of explainable models. Compared with prior knowledge-based VQA datasets (e.g., OK-VQA), A-OKVQA is larger and uniquely provides sentence-level rationales, yielding a more diverse and challenging testbed for multimodal reasoning.

   \item \textbf{MINT1T} ~\cite{mint1t} is a large-scale open source multimodal interleaved dataset that preserves image and text order, assembled from HTML, PDFs, and arXiv at trillion token and billion image scale. It uses targeted quality filtering, NSFW screening, limited PII redaction, and extensive deduplication across text and images to improve cleanliness and diversity. Compared to OBELICS, it provides broader coverage with longer and more image-dense documents, and models trained on it achieve competitive or improved results on multimodal benchmarks.  
\end{itemize}

\subsection{Evaluation Models}

\begin{itemize}
  \item \textbf{CLIP} ~\cite{radford2021learning} is a dual-encoder vision–language model that aligns images and text in a shared embedding space via a symmetric contrastive objective over large batches. Trained on hundreds of millions of image–text pairs, it enables zero-shot recognition by turning class names or descriptions into text prompts that act as a classifier. This design yields strong, scalable performance across diverse benchmarks without task-specific fine-tuning.

  \item \textbf{SigLIP2} ~\cite{tschannen2025siglip2multilingualvisionlanguage} is a multilingual vision and language encoder family that remains architecture-compatible with SigLIP and uses a unified training recipe combining a sigmoid image-text objective, a decoder for captioning and localization, and self-distillation with masked prediction to strengthen dense and spatial features; a NaFlex variant supports native aspect ratios and multiple resolutions, and the models deliver strong zero-shot classification and retrieval alongside improved localization and dense prediction.

  \item \textbf{OpenCLIP} ~\cite{Cherti2022ReproducibleSL} is an open source CLIP training and evaluation stack built on LAION data that enables fully reproducible studies of scaling laws; trained on billions of image text pairs, it releases the largest public CLIP models and shows that the training distribution drives task-dependent scaling, with OpenCLIP improving more on zero-shot retrieval while OpenAI CLIP improves more on zero-shot classification, alongside strong results on ImageNet, VTAB plus, and COCO retrieval.

  \item \textbf{Jina-CLIP-V1} ~\cite{2405.20204} is a unified contrastive language–image model that also serves as a strong text retriever: using EVA02 ViT-B/16 as the image encoder and JinaBERT v2 as the text encoder in a staged training pipeline, it jointly optimizes image–text and text–text objectives.

  \item \textbf{Jina-CLIP-V2} ~\cite{koukounas2024jinaclipv2multilingualmultimodalembeddings} is a multilingual dual-encoder vision–language model (XLM-RoBERTa text tower + EVA02-L/14 vision tower; ~865M params) trained with multi-task contrastive objectives over text–text, image–text, and hard-negative triplets. It employs Matryoshka representations for flexible embedding sizes and higher-resolution training for document images, yielding strong retrieval performance in English and across ~30 languages (including ViDoRe), while remaining openly available for reproducible research.

    \item \textbf{NEV} ~\cite{nussbaum2024nomic} is an open weights image embedding model that shares a unified latent space with nomic embed text via a LiT style recipe that freezes the text encoder while adapting an EVA02 ViT B/16 vision tower. Trained on a large curated web corpus for multiple epochs, it targets strong zero shot classification and cross modal retrieval, reporting gains over CLIP baselines across ImageNet, DataComp, and MTEB style evaluations and providing a practical unified embedding space for vision, language, and multimodal tasks.

    \item \textbf{E5-V} ~\cite{jiang2024e5vuniversalembeddingsmultimodal} is a multimodal embedding model that maps images, text, and interleaved inputs into a single semantic space using a prompt-based representation (for example, summarizing content in one word), which bridges the modality gap without multimodal fine-tuning. Trained only on text pairs with a contrastive objective while removing the visual pathway during training for major efficiency gains, it transfers at inference to image and mixed modality inputs and delivers strong zero-shot results on text and image retrieval, composed image retrieval, image to image retrieval with rendered text, and standard sentence similarity benchmarks.

   \item \textbf{MM-Embed} ~\cite{lin2025mmembeduniversalmultimodalretrieval} is a universal multimodal retriever built on MLLMs that unifies text, images, and interleaved inputs; it introduces modality-aware hard negative mining and continuous fine-tuning to curb MLLM modality bias and bolster text retrieval, achieving state-of-the-art results on M-BEIR and surpassing NV-Embed-v1 on MTEB.

   \item \textbf{Ola} ~\cite{liu2025ola} is an omnimodal language model for unified image, video, and audio understanding that uses native resolution visual encoding with a Local Global Attention Pooling layer for efficient token reduction, integrates a dual audio encoder with Whisper v3 for speech and BEATs for music along with simple MLP connectors to project all modalities into a shared token space, and emphasizes cross modal alignment by treating video as the central bridge within a progressive training schedule to balance modalities. 

   \item \textbf{Qwen2-VL} ~\cite{Qwen2VL} is a family of open-weight vision–language models (2B/8B/72B) that replaces fixed-resolution pipelines with Naive Dynamic Resolution and fuses multimodal positions via M-RoPE, achieving state-of-the-art perception across images and long videos and results comparable to GPT-4o and Claude 3.5 on key benchmarks.

   \item \textbf{InternVL-3} ~\cite{Zhu2025InternVL3EA} is an open-source multimodal large language model that natively unifies vision and language via a single pretraining stage, avoiding post-hoc adapters and alignment. Built on a ViT–MLP–LLM stack with Variable Visual Position Encoding for long-context perception, it delivers state-of-the-art open-source results across diverse multimodal benchmarks.

   \item \textbf{Gemini-2.5-Flash} ~\cite{google2024gemini} is a multimodal, low-latency model optimized for fast, cost-efficient inference across text, code, vision, and audio. It supports streaming generation, tool use, and extended context, making it a strong choice for interactive agents and high-throughput production systems where responsiveness is prioritized over peak accuracy.

   \item \textbf{GPT5} ~\cite{openai2025-gpt5} is a next-generation generative pre-trained transformer that advances reliability, reasoning, and multimodal understanding. It integrates longer-context modeling with robust tool use (e.g., function calling and retrieval) and a safety-focused post-training pipeline to improve calibration and control. Together, these capabilities make GPT-5 a practical foundation for research and applications requiring dependable, grounded generation.
  
\end{itemize}

\subsection{Experimental Code}

To promote transparency and ensure the reproducibility of our work, we will release all experimental code, datasets, and detailed tutorials necessary for replicating our experiments. Our goal is to make it straightforward for researchers and practitioners to reproduce our results, regardless of their technical background. Additionally, by providing comprehensive documentation and clear guidelines, we aim to facilitate the extension of our method to other models and architectures, enabling the broader research community to explore its potential applications and improvements. We believe that open and reproducible research is essential for advancing the field and fostering collaboration.

\renewcommand{\thefigure}{E.\arabic{figure}}
\renewcommand{\thetable}{E.\arabic{table}}
\setcounter{figure}{0}
\setcounter{table}{0}
\section{Context-Window Limitation Analysis}

A natural alternative to retrieval is to directly encode all items into the long context of frontier MLLMs (e.g., GPT, Gemini) and then perform end-to-end reasoning. However, this strategy is computationally prohibitive due to the quadratic growth of input tokens across heterogeneous modalities.  

\paragraph{Tokenization cost.}  
Based on Gemini’s official tokenization rules, the total token budget is
\[
T_{\mathrm{total}}
= 258 \sum_{m=1}^{M}\Big\lceil \tfrac{w_m}{768}\Big\rceil \Big\lceil \tfrac{h_m}{768}\Big\rceil
 +  263 \sum_{n=1}^{N} L_n
 +  T_{\mathrm{text}},
\]
where $(w_m,h_m)$ are image dimensions in pixels, $L_n$ is the duration of the $n$-th video in seconds, and $T_{\mathrm{text}}$ is the number of textual tokens. Each $768{\times}768$ image patch costs approximately 258 tokens, while each second of video costs about 263 tokens.  

Our benchmark contains 46,260 multimodal items (images, videos, and documents). Even under conservative assumptions—rescaling images to a single patch and compressing videos to low frame rates—the total budget reaches nearly 200M tokens. 
This exceeds the largest publicly available context window (1M tokens) by more than two orders of magnitude. In practice, many items are larger than a single patch or longer than a few seconds, which pushes the requirement even higher.  

This analysis highlights a fundamental limitation: even with million-token context windows, brute-force ingestion cannot approximate real-world conditions. Without targeted evidence selection, the input size scales linearly with corpus size but quadratically with attention, making end-to-end encoding infeasible. Therefore, MultiHaystack plays a critical role by providing a realistic evaluation setting where retrieval, rather than ever-larger context windows, is the decisive factor for scalable multimodal reasoning.

% \newpage
\renewcommand{\thefigure}{F.\arabic{figure}}
\renewcommand{\thetable}{F.\arabic{table}}
\setcounter{figure}{0}
\setcounter{table}{0}
\section{More analysis}
\begin{wraptable}[8]{r}{0.26\textwidth}
\vspace{-44pt}
\renewcommand{\arraystretch}{1}
\begin{center}
  \centering
\caption{\textbf{VQA performance in zero context.}}
  \label{tab:zero}
  \resizebox{0.25\textwidth}{!}{%
    \setlength{\tabcolsep}{6pt}%
    \renewcommand{\arraystretch}{1.0}%
    \begin{tabular}{l c}
      \toprule
      \textbf{Model} & \textbf{Overall} \\
      \midrule
      Ola      
      & 0.54 \\

      InternVL-3 
       & 0.80 \\

      Qwen2-VL  
       & 0.67 \\

      Gemini-2.5-Flash
       & 1.07 \\

      GPT-5   
       & 4.28 \\
      \bottomrule
    \end{tabular}%
  }
\end{center}
\end{wraptable}

\subsection{Zero-context Visual Question Answering}
One might suggest evaluating a zero-context baseline ($k=0$), where the model answers questions without any retrieved content. 
However, this setting is fundamentally incompatible with MultiHaystack. 
During construction, we explicitly apply a retrieval-independence filter to remove any question that could be answered from prior knowledge or common sense alone. 
As shown in \cref{tab:zero}, we evaluate several models under the zero-context setting, and find that even the strongest one (GPT-5) achieves only 4.28\% accuracy, confirming that virtually all queries require retrieval to be solvable.    
Consequently, reporting $k=0$ is not meaningful and would only obscure the purpose of the benchmark, which is to disentangle retrieval and reasoning in multimodal contexts. 
Instead, we provide \emph{Gold in Top-1/5} results (\cref{fig:video_image_doc_radar}), where the ground-truth item is guaranteed to be retrieved. 
These serve as a principled upper bound, directly isolating reasoning ability under perfect retrieval, and thus provide a far more informative diagnostic than an artificial zero-context baseline.

\subsection{Parameter Selection for Data Enrichment.}
A central challenge in data enrichment is to retain informative positives while suppressing noisy distractors. 
To this end, we first apply a coarse CLIP threshold to discard obviously unrelated candidates. 
We then compute the mean CLIP similarity for positive pairs ($\approx$0.74) and select a principled interval around it. 
\begin{wrapfigure}[13]{r}{0.55\textwidth}
    \centering
    \includegraphics[width=0.55\textwidth]{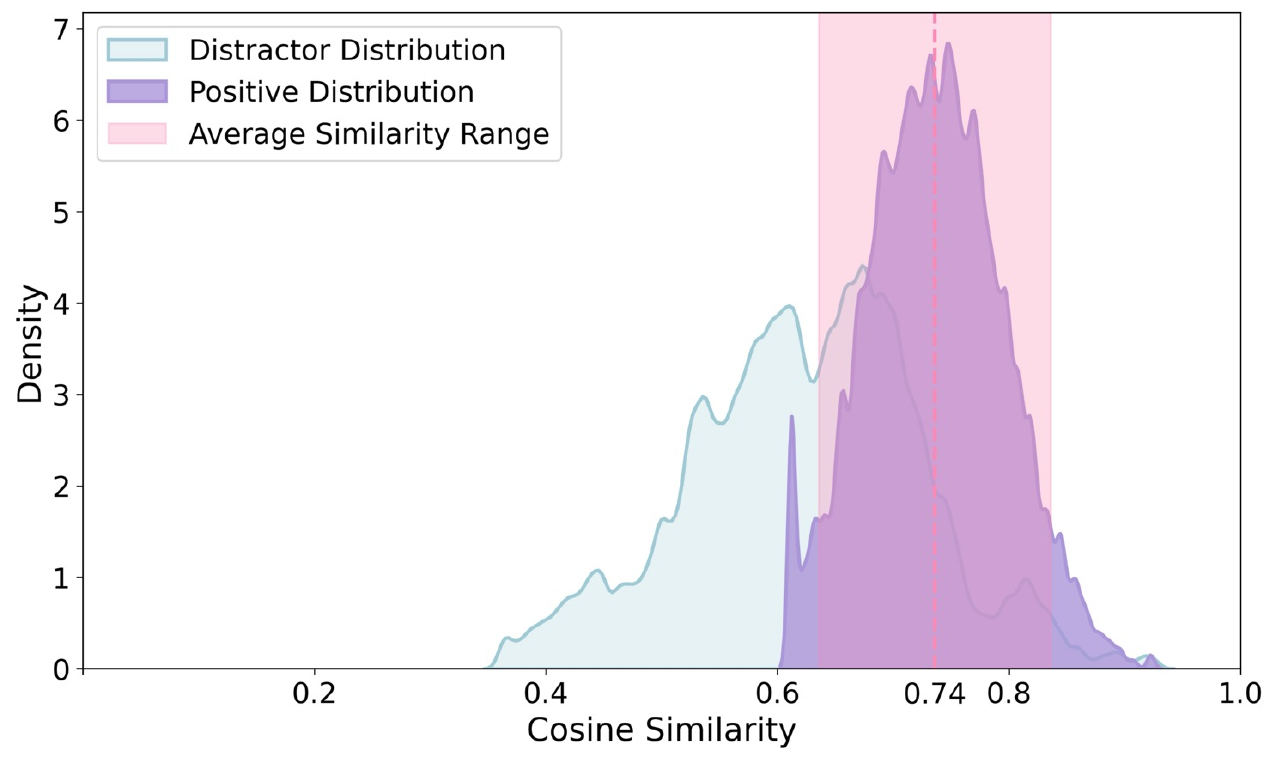}
    \caption{\textbf{Distribution of cosine similarity scores.} }
    \label{fig:data_enrich_distribution}
\end{wrapfigure}%
In our dataset, this corresponds to [0.64, 0.84], which is broad enough to preserve the majority of true positives while excluding distractors with artificially high similarity or positives with abnormally low similarity. 
\Cref{fig:data_enrich_distribution} highlights this separation: the purple curve shows CLIP-based similarities between each question and its ground-truth positive image, peaking near 0.74, while the green curve shows \texttt{vidore/colqwen2-v0.1}-based similarities between each question and a large pool of candidate distractors, concentrated at lower values. 
By explicitly grounding the threshold in the empirical distributions of CLIP positives and \texttt{vidore/colqwen2-v0.1} distractors, this procedure yields a cleaner candidate pool, mitigating CLIP-only bias and reducing noise propagation, ultimately stabilizing downstream training.

\begin{wraptable}[5]{r}{0.5\textwidth}
\vspace{-48pt}
\centering
% \caption{\textbf{VQA performance.} Each model answers questions using top-5 items retrieved from cross-modality inputs; {\color{gray}gray} numbers show single-modality Recall@5 for reference.}
% \caption{\textbf{VQA performance.} Models generate answers based on the top-5 cross-modality retrieved items. For comparison, single-modality Recall@5 scores are shown in {\color{gray}gray}.}
\caption{\textbf{VQA performance.} Results are conditioned on top-5 cross-modal retrieval context. Baseline single-modality Recall@5 scores are indicated in {\color{gray}gray}.}
  \resizebox{0.5\textwidth}{!}{%
    \setlength{\tabcolsep}{6pt}%
    \renewcommand{\arraystretch}{1.0}%
    \begin{tabular}{l c c c c}
      \toprule
      \textbf{Model} & \textbf{Video} & \textbf{Image} & \textbf{Document} & \textbf{Overall} \\
      \midrule

      InternVL-3.5 
      & 20.95 {\color{gray}\scriptsize(26.67)} 
      & 30.95 {\color{gray}\scriptsize(39.72)} 
      & 51.20 {\color{gray}\scriptsize(54.55)} 
      & 35.21 {\color{gray}\scriptsize(42.03)} \\

      Qwen2.5-VL  
      & 19.05 {\color{gray}\scriptsize(21.90)} 
      & 22.17 {\color{gray}\scriptsize(28.41)} 
      & 31.10 {\color{gray}\scriptsize(34.45)} 
      & 24.23 {\color{gray}\scriptsize(29.18)} \\
      
      \bottomrule
    \end{tabular}%
  }
  \label{tab:rebuttal_vqa}
\end{wraptable}

\subsection{Advanced models}
We further evaluated the newer models InternVL-3.5 and Qwen2.5-VL (see \cref{tab:rebuttal_vqa}), and both remain well below their single-modality upper bounds once retrieval is involved, confirming that even the latest models show the same limitations.

\renewcommand{\thefigure}{G.\arabic{figure}}
\renewcommand{\thetable}{G.\arabic{table}}
\setcounter{figure}{0}
\setcounter{table}{0}
\section{Error Analysis}
\label{sec:error}

\begin{figure}[!ht]
    \centering
    \includegraphics[width=0.95\linewidth]{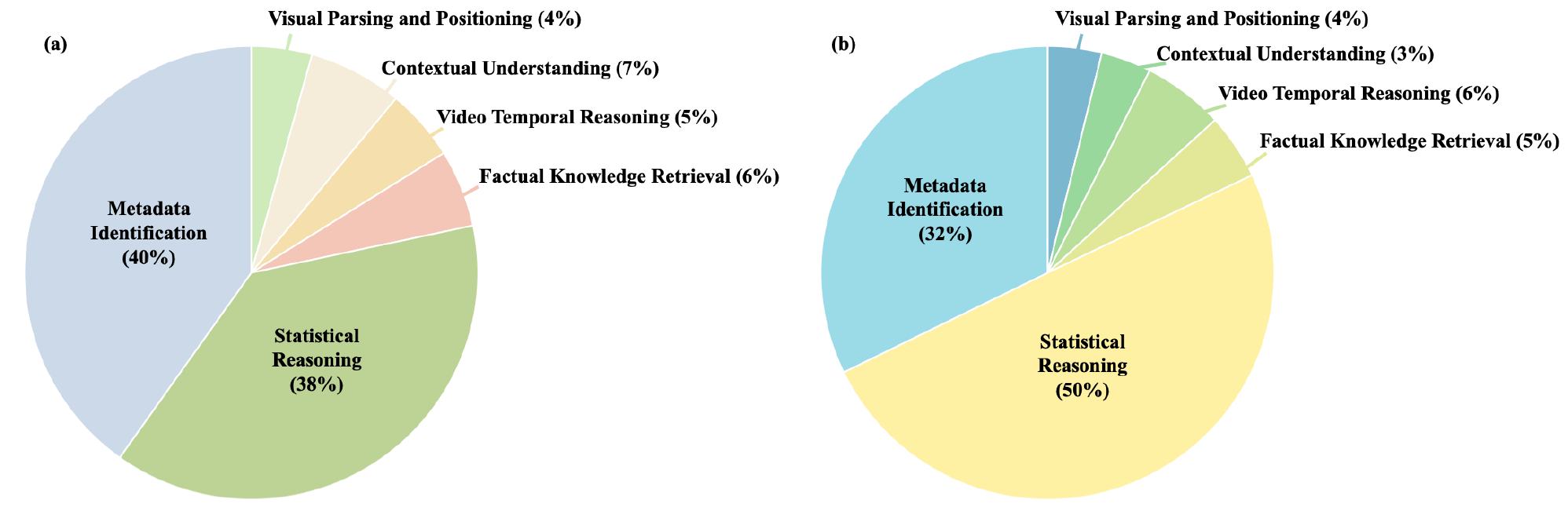}
    \caption{\textbf{Distribution of error types.} Panels: (a) retrieval error distribution and (b) reasoning error distribution. Retrieval errors are quantified by Recall@5 with the strongest retriever (E5-V), while reasoning errors are evaluated using VQA with the strongest reasoning model (GPT-5). Retrieval errors dominate across tasks, though reasoning errors remain substantial.}
    \label{fig:error_dist}
\end{figure}

To gain deeper insights into the failure of the current VLMs and MLLMs, we further perform a qualitative error analysis. We first compute the statistical distribution of the two major error categories: \textit{retrieval errors} and \textit{reasoning errors}. 
As shown in \cref{fig:error_dist}, retrieval errors account for a larger proportion overall, reflecting the difficulty of grounding queries in subtle but decisive evidence. Reasoning errors, though fewer, remain substantial, highlighting that even with correct retrieval, models frequently fail to extract or align fine-grained content. This distribution underscores that progress in both retrieval and reasoning is necessary to reduce failure rates.

\noindent Building on this distributional view, we next examine representative cases across six tasks: Contextual Understanding, Factual Knowledge Retrieval, Metadata Identification, Statistical Reasoning, Video Temporal Reasoning, and Visual Parsing and Positioning. \Cref{fig:cu_error}–\ref{fig:vpp_error} illustrate typical examples. Retrieval errors commonly arise when models are biased toward salient but irrelevant signals (e.g., league logos, headlines, colorful infographics), overlooking subtle yet decisive cues such as timestamps or spatial relations. Reasoning errors, on the other hand, often stem from shallow associative processing, where the system outputs plausible but incorrect answers (e.g., predicting ``State Farm'' instead of the correct sponsor, misreporting 2,743 instead of 2,740, or confusing the spatial relation between characters). These examples reveal a consistent bottleneck: current models struggle with sensitivity to fine-grained task-relevant details, both at the retrieval and reasoning stages.

\subsection{Contextual Understanding (CU)}

\begin{figure}[!ht]
    \centering
    \includegraphics[width=\linewidth]{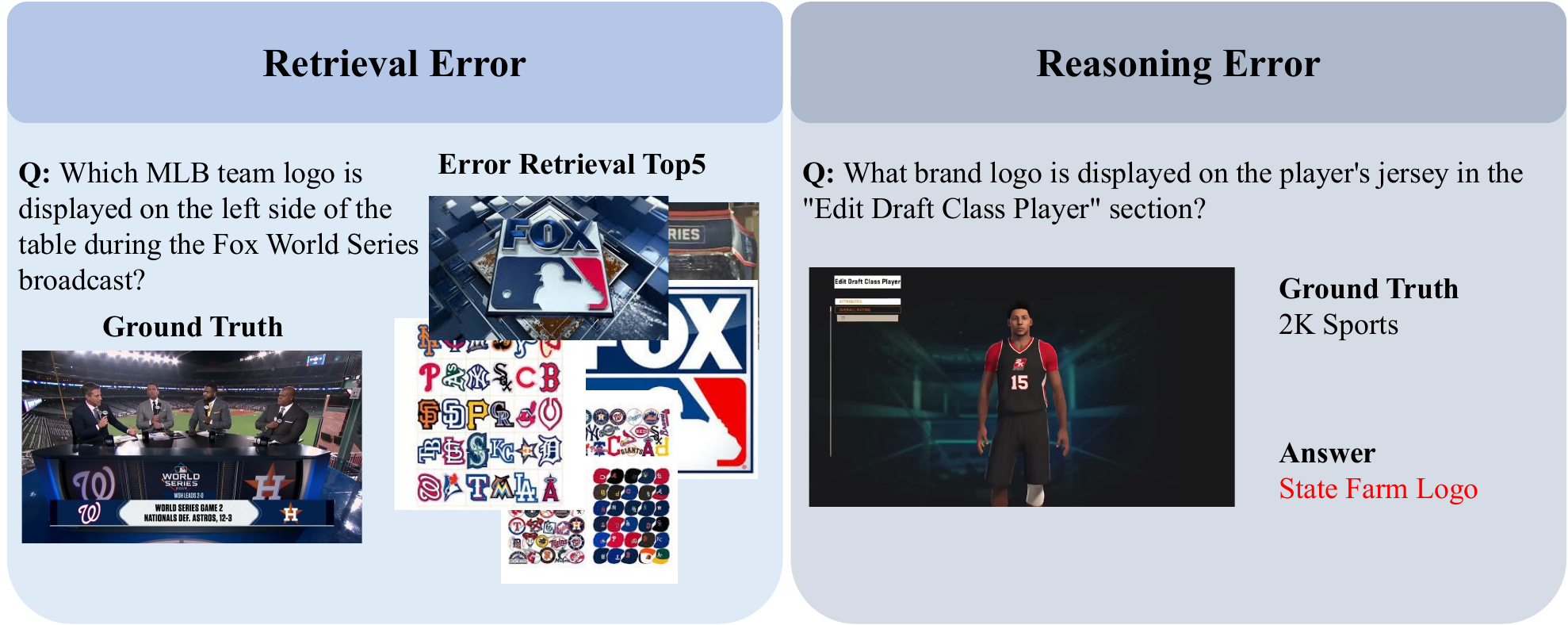}
    \caption{\textbf{Contextual Understanding representative error cases.}}
    \label{fig:cu_error}
\end{figure}

\noindent \textbf{Retrieval error.}  
Contextual understanding requires models to attend to subtle textual or symbolic signals embedded in a scene. As shown in \cref{fig:cu_error}, the retriever frequently selects broadcast frames with prominent Fox or MLB league logos, while failing to prioritize the smaller team emblem on the desk that is key to answering the query. This reveals a systematic bias toward globally salient elements and insufficient sensitivity to localized cues that define context.  

\noindent \textbf{Reasoning error.}  
Even when the relevant frame is retrieved, models often fail to identify the intended target. In the jersey example, the system outputs ``State Farm'', a frequent sponsor in sports scenes, instead of the actual ``2K Sports'' logo. This demonstrates shallow associative reasoning, where models rely on prior familiarity with common patterns rather than aligning their predictions with the fine-grained evidence present in the scene.

\subsection{Factual Knowledge Retrieval (FKR)}

\begin{figure}[!ht]
    \centering
    \includegraphics[width=\linewidth]{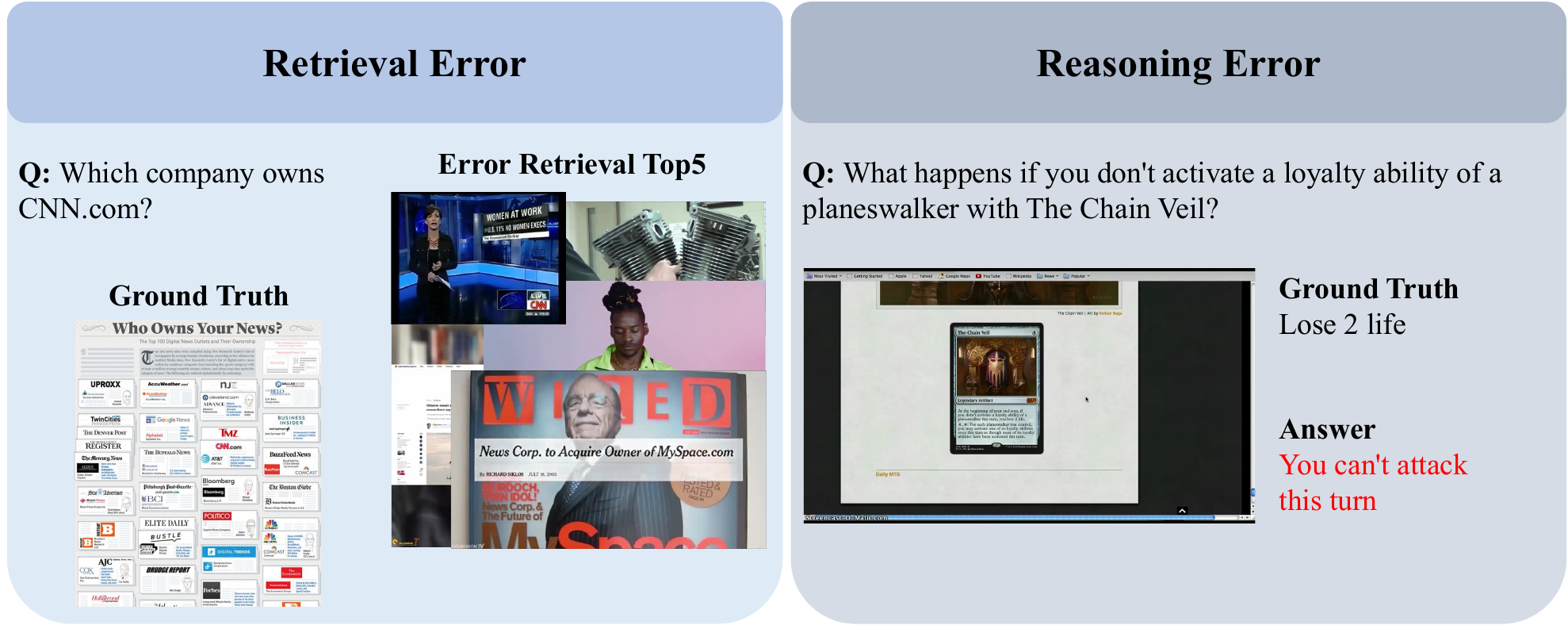}
    \caption{\textbf{Factual Knowledge Retrieval representative error cases.}}
    \label{fig:fkr_error}
\end{figure}

\noindent \textbf{Retrieval error.}  
Factual knowledge retrieval tasks demand grounding in specific factual sources rather than surface similarity. \Cref{fig:fkr_error} shows that retrievers often select generic news articles with overlapping topics, while missing the ownership chart that directly encodes the required fact. This indicates difficulty in filtering out visually or lexically similar distractors that lack factual relevance.  

\noindent \textbf{Reasoning error.}  
When the correct evidence is retrieved, models may still produce factually incorrect outputs. In the card-game case, the system outputs ``You can't attack this turn'' instead of the precise rule ``Lose 2 life.'' Such errors reflect limited capacity to extract exact symbolic content when distractors are semantically close or when plausible but incorrect alternatives exist in the model’s training distribution.  

\newpage

\subsection{Metadata Identification (MI)}

\begin{figure}[!ht]
    \centering
    \includegraphics[width=\linewidth]{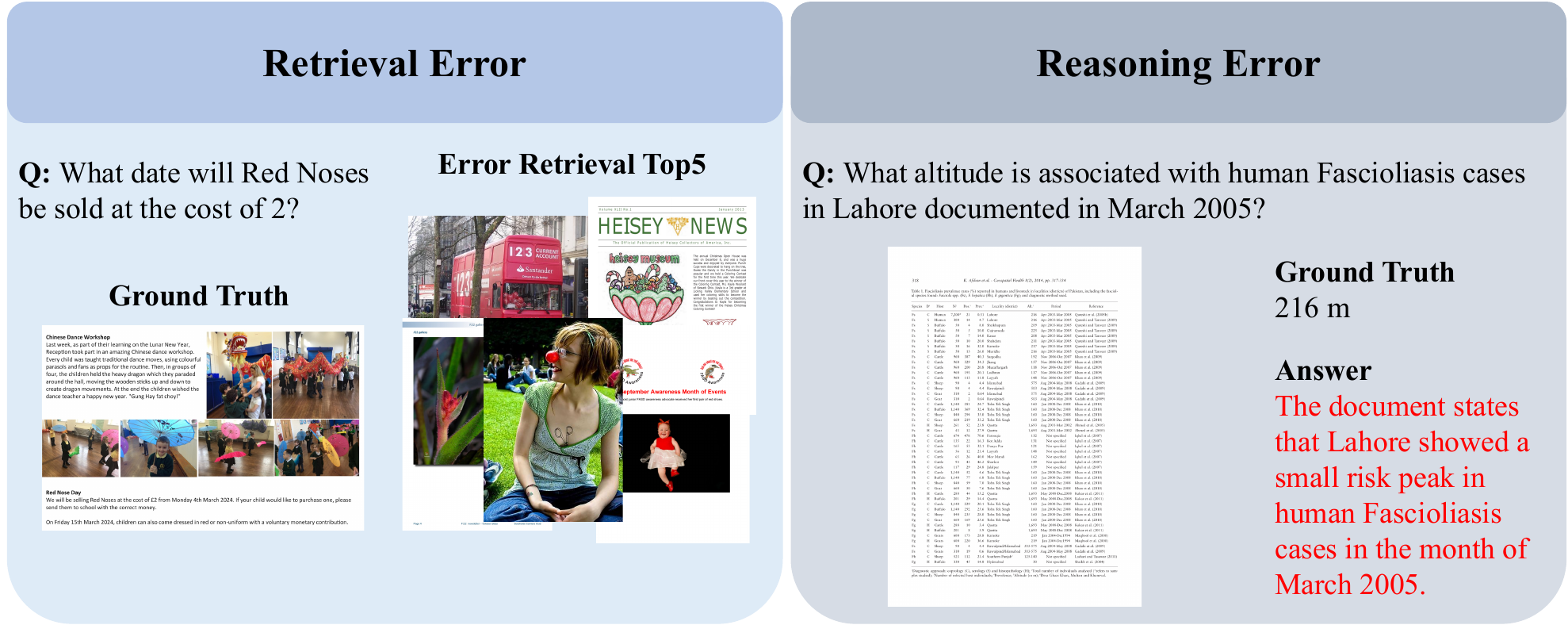}
    \caption{\textbf{Metadata Identification representative error cases.}}
    \label{fig:mi_error}
\end{figure}

\noindent \textbf{Retrieval error.}  
Metadata identification tasks emphasize peripheral information such as dates, publishers, or attribution details. As shown in \cref{fig:mi_error}, the retriever often selects documents with salient but irrelevant headlines (e.g., ``Heisey News''), while failing to identify the document that actually contains the event date. This suggests that subtle metadata cues are systematically underweighted during retrieval.  

\noindent \textbf{Reasoning error.}  
Even with the correct source, models may paraphrase broader contextual information instead of pinpointing the requested metadata. In the example, the system discusses risk levels but fails to extract the precise altitude value of 216 m. This highlights the difficulty in focusing on small but decisive details, especially when they appear in dense or noisy layouts.  

\subsection{Statistical Reasoning (SR)}

\begin{figure}[!ht]
    \centering
    \includegraphics[width=\linewidth]{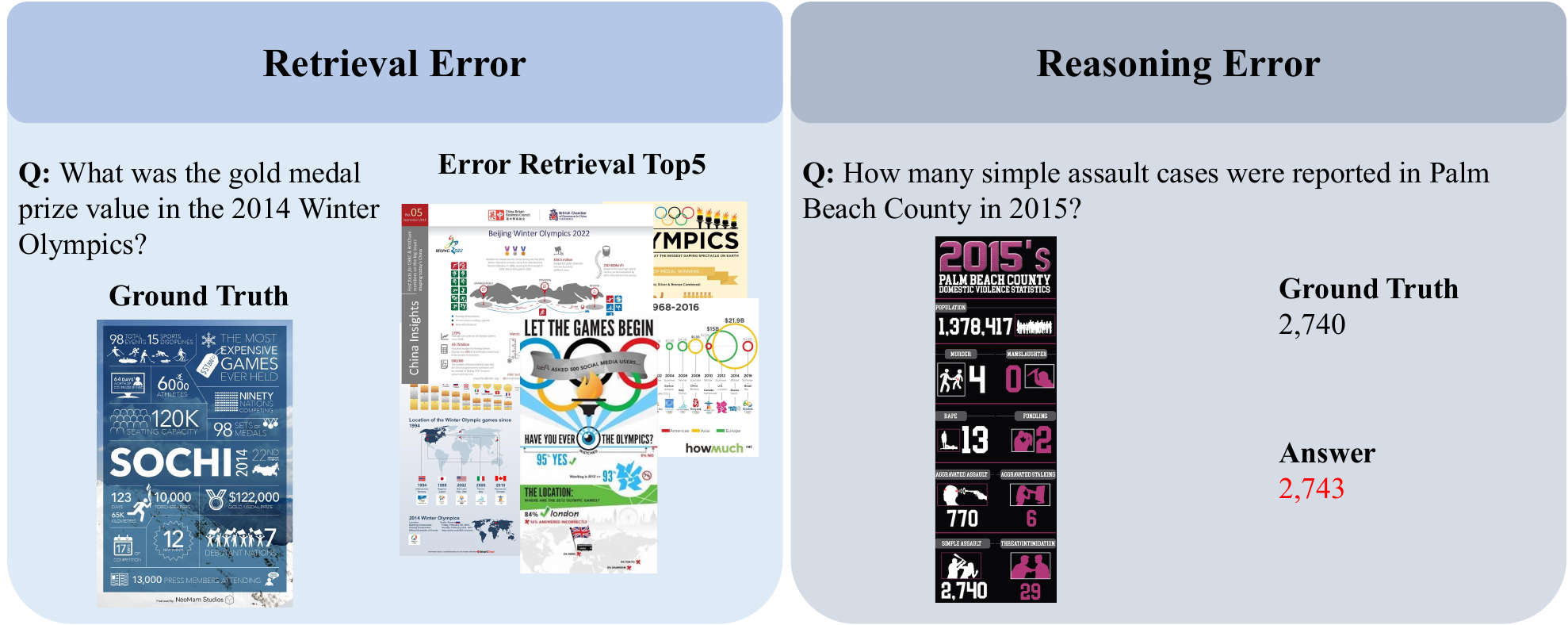}
    \caption{\textbf{Statistical Reasoning representative error cases.}}
    \label{fig:sr_error}
\end{figure}

\noindent \textbf{Retrieval error.}  
Statistical reasoning tasks hinge on retrieving charts or tables with exact quantitative relevance. \Cref{fig:sr_error} shows that retrievers sometimes surface colorful but semantically irrelevant infographics, prioritizing layout or style over the numerical semantics that matter for the query. This reveals a gap in embedding models’ ability to encode quantitative intent.  

\noindent \textbf{Reasoning error.}  
Once the correct chart is retrieved, errors often stem from fragile visual numeracy. The system may miscount bars, misalign values with axes, or confuse close numbers (e.g., reporting 2,743 instead of 2,740). Such mistakes indicate that while models perceive the chart, their mapping from visual encodings to precise numerical answers is brittle and error-prone.  

\newpage

\subsection{Video Temporal Reasoning (VTR)}

\begin{figure}[!ht]
    \centering
    \includegraphics[width=\linewidth]{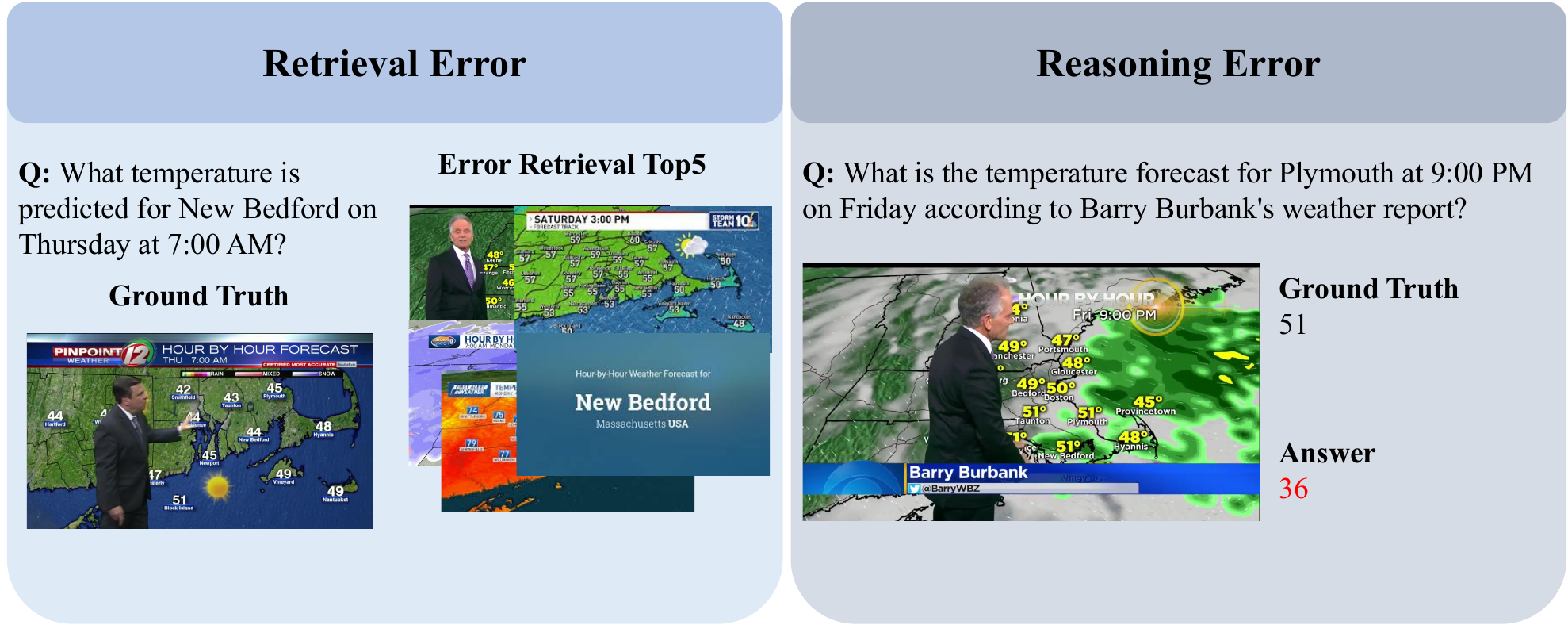}
    \caption{\textbf{Video Temporal Reasoning representative error cases.}}
    \label{fig:vtr_error}
\end{figure}

\noindent \textbf{Retrieval error.}  
Video temporal reasoning tasks require isolating evidence at the correct temporal point. As illustrated in \cref{fig:vtr_error}, the retriever often selects weather maps with similar layouts but corresponding to the wrong time or location, failing to encode temporal anchors. This points to the underrepresentation of sequential and time-sensitive features in retrieval embeddings.  

\noindent \textbf{Reasoning error.}  
Even when the correct video frame is retrieved, the model may misread numeric overlays or confuse temporal ordering, e.g., predicting ``36'' instead of ``51.'' These errors demonstrate the fragility of temporal–numerical reasoning, where minor OCR-like mistakes or misinterpretations of frame order propagate into incorrect conclusions.  

\subsection{Visual Parsing and Positioning (VPP)}

\begin{figure}[!ht]
    \centering
    \includegraphics[width=\linewidth]{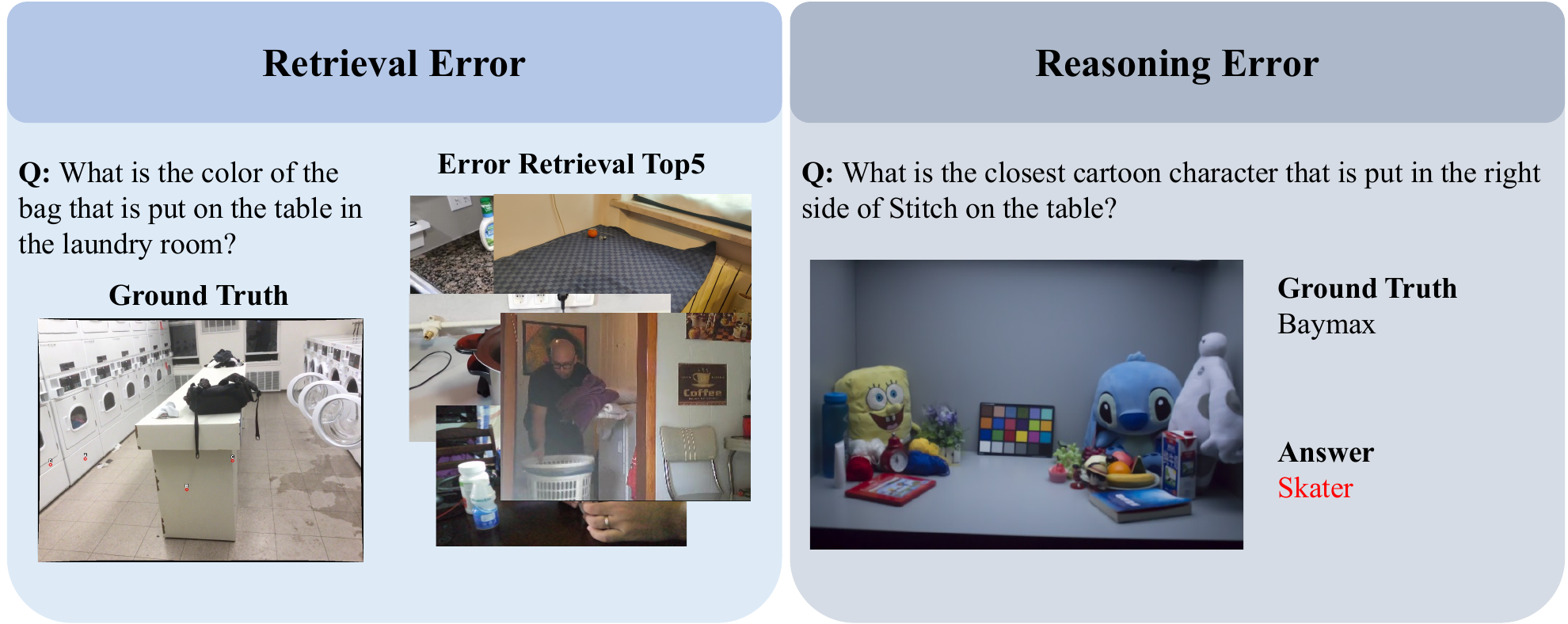}
    \caption{\textbf{Visual Parsing and Positioning representative error cases.}}
    \label{fig:vpp_error}
\end{figure}

\noindent \textbf{Retrieval error.}  
Visual parsing and positioning requires attention to spatial relationships rather than global scene similarity. \Cref{fig:vpp_error} shows retrieval returning indoor scenes with similar textures or objects (e.g., laundry baskets, storage rooms) instead of the specific bag-on-table instance. This reflects insufficient encoding of spatial layout information in the retrieval stage.  

\noindent \textbf{Reasoning error.}  
When the relevant scene is retrieved, reasoning errors arise from misinterpreting spatial relations. The model identifies the wrong character (``Skater'') instead of ``Baymax'' when asked about the figure to the right of Stitch, showing that relational parsing across entities remains a bottleneck even when object recognition is accurate.  

\subsection{Summary of Error Patterns}

Across the six tasks, two consistent tendencies emerge. Retrieval errors are predominantly driven by saliency bias: systems privilege visually prominent elements such as logos, headlines, or colorful charts while neglecting the subtle but decisive cues that ground context, such as timestamps, metadata, or spatial layouts. This suggests that current multimodal embeddings fail to adequately encode task-specific contextual signals that are less obvious but more critical.  

Reasoning errors, in contrast, often reflect shallow associative processing. Models default to frequent or plausible outputs, common sponsors in sports broadcasts, approximate numbers in charts, or generic spatial relations, instead of extracting the exact information encoded in the evidence. These patterns indicate that while retrieval and reasoning failures manifest differently, both are rooted in insufficient sensitivity to fine-grained, task-relevant details that determine correctness. Addressing this limitation will require embedding models that better capture subtle contextual cues and reasoning modules that enforce tighter alignment between queries and retrieved evidence.

\renewcommand{\thefigure}{H.\arabic{figure}}
\renewcommand{\thetable}{H.\arabic{table}}
\setcounter{figure}{0}
\setcounter{table}{0}

\section{Qualitative Case Studies of Model Performance}

\subsection{Contextual Understanding (CU)}

\begin{figure}[!ht]
    \centering
    \includegraphics[width=0.8\linewidth]{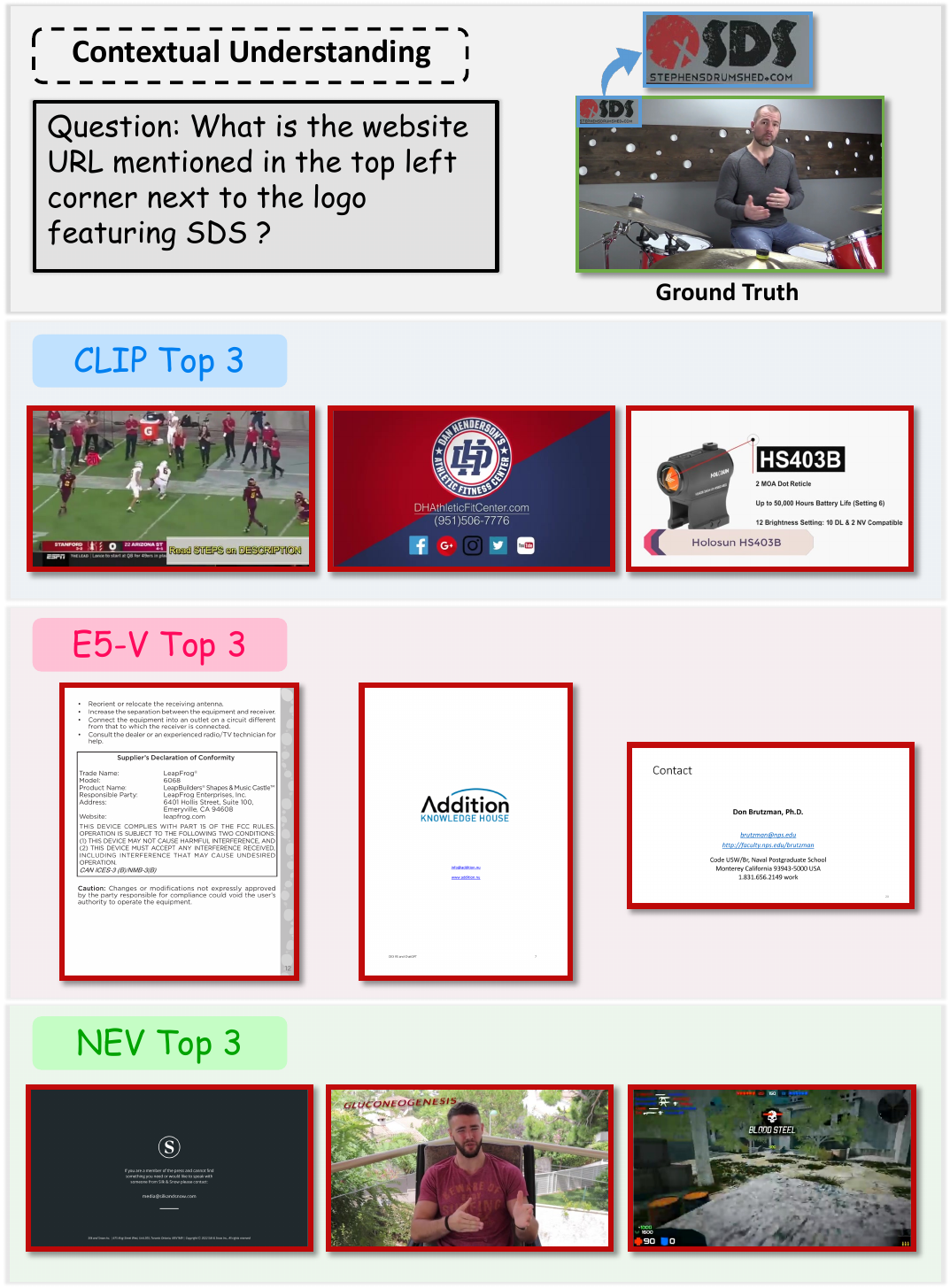}
    \caption{\textbf{Contextual Understanding Retrieval.}}
    \label{fig:case_retrieval_cu}
\end{figure}

\begin{figure}[!ht]
    \centering
    \includegraphics[width=0.8\linewidth]{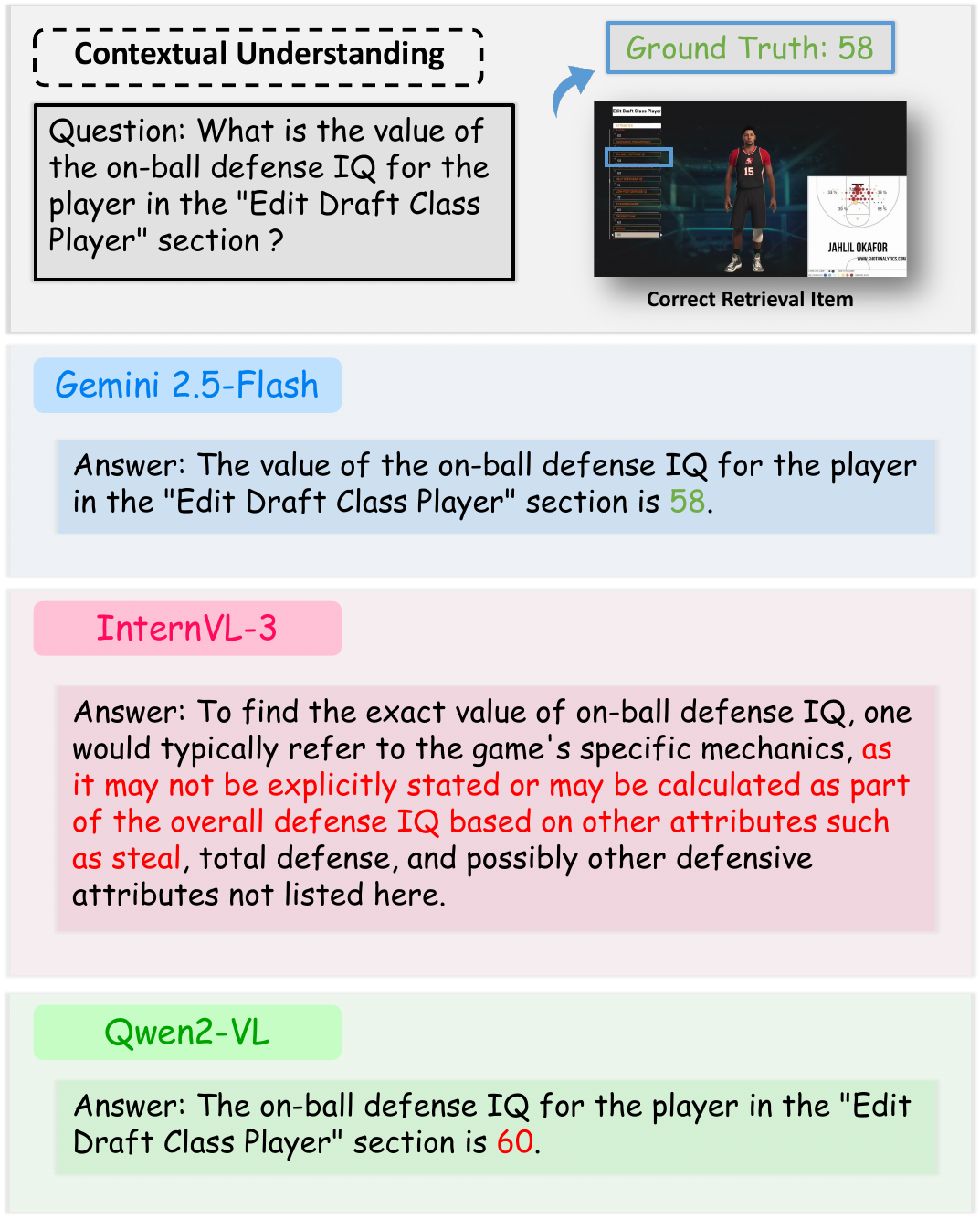}
    \caption{\textbf{Contextual Understanding VQA.}}
    \label{fig:case_vqa_cu}
\end{figure}

In the Contextual Understanding retrieval task (See \cref{fig:case_retrieval_cu}), the baseline models struggle to pinpoint the exact video frame containing a specific URL, often retrieving visually similar but entirely incorrect scenes or documents. For the VQA task (See \cref{fig:case_vqa_cu}), InternVL-3 and Qwen2-VL fail to accurately extract the specific "on-ball defense IQ" value from the dense, text-heavy game interface, outputting incorrect numbers or explanations.

\newpage

\subsection{Factual Knowledge Retrieval (FKR)}

\begin{figure}[!t]
    \centering
    \includegraphics[width=0.8\linewidth]{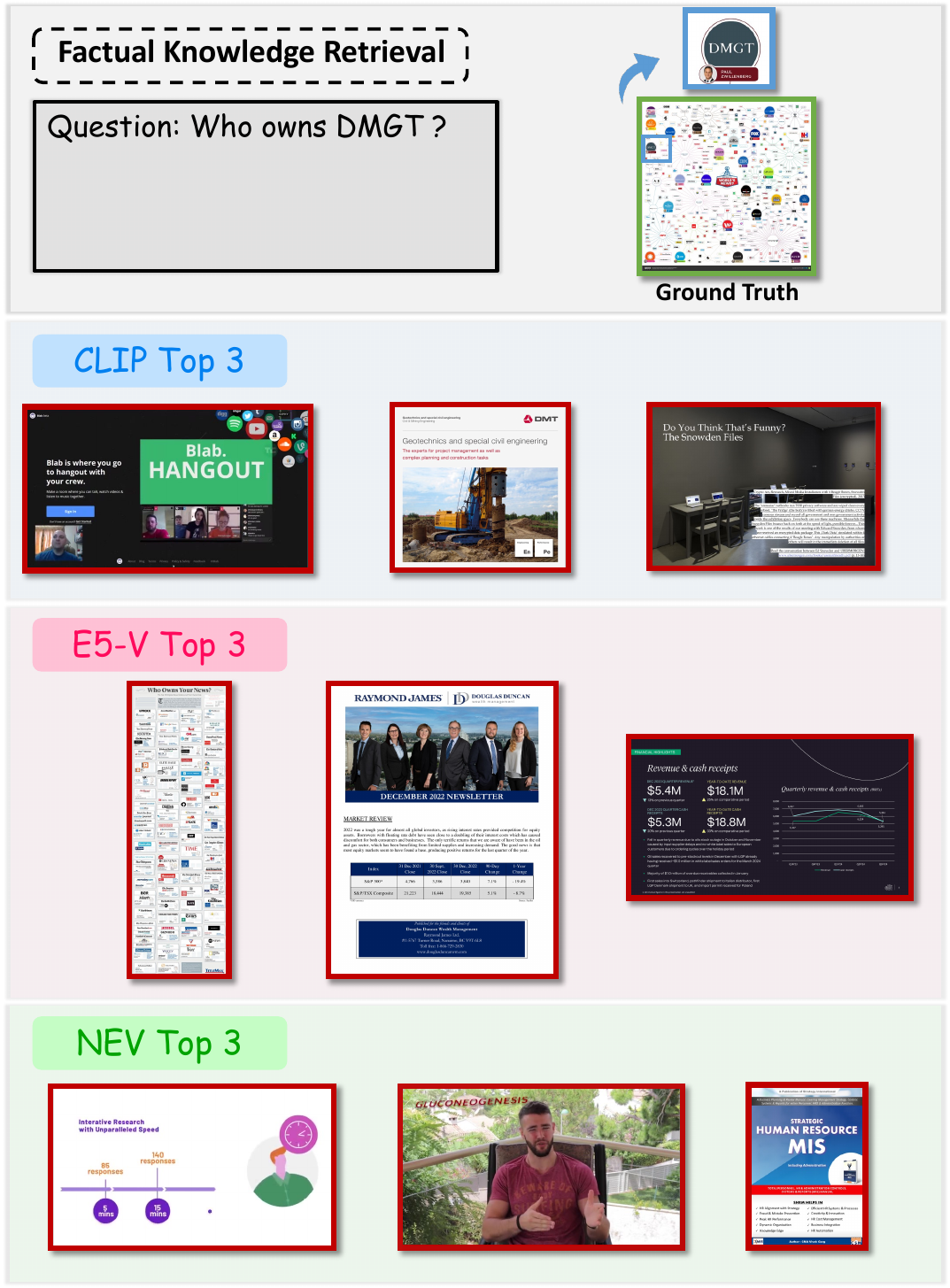}
    \caption{\textbf{Factual Knowledge Retrieval.}}
    \label{fig:case_retrieval_fkr}
\end{figure}

\begin{figure}[!t]
    \centering
    \includegraphics[width=0.8\linewidth]{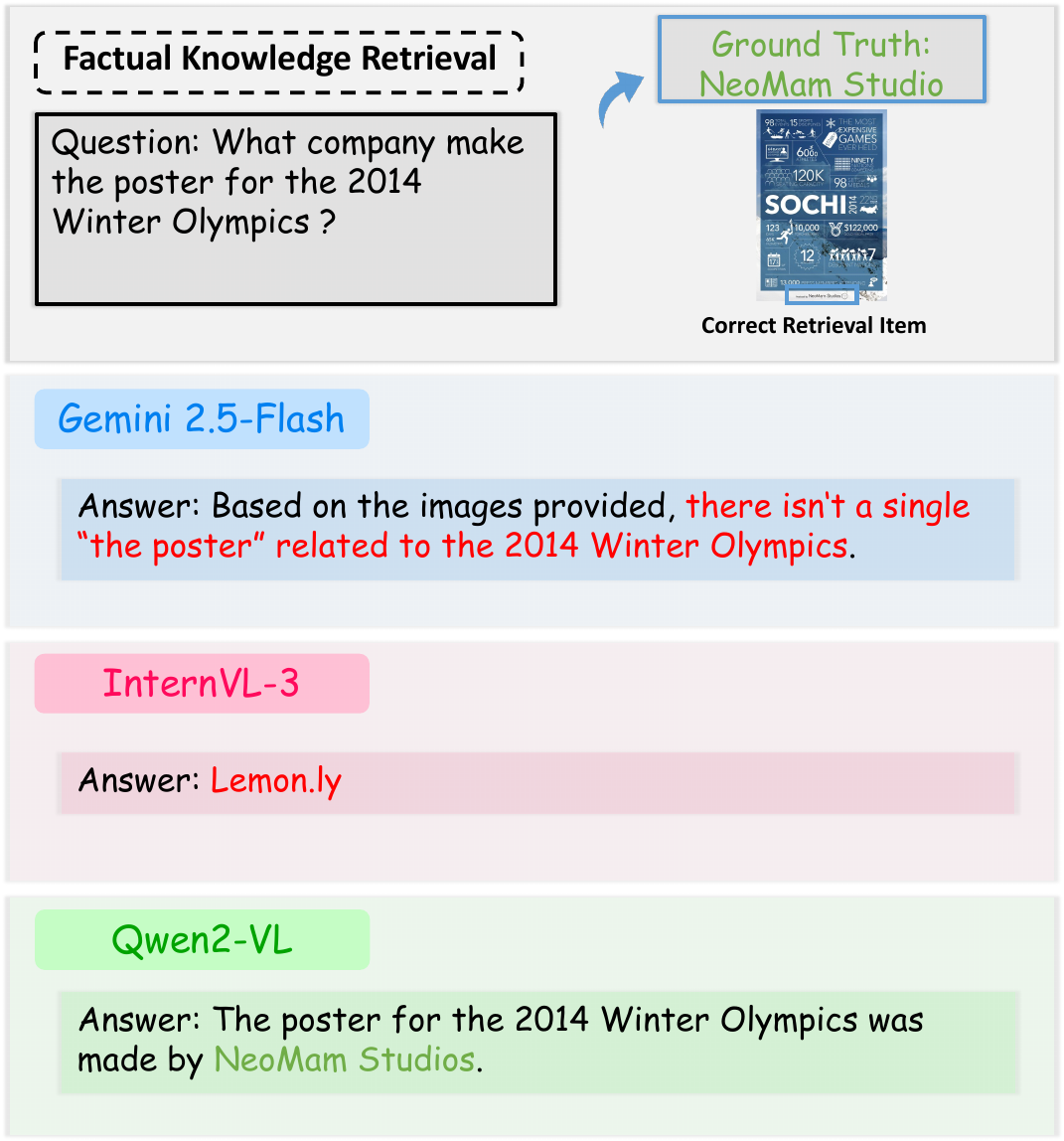}
    \caption{\textbf{Factual Knowledge VQA.}}
    \label{fig:case_vqa_fkr}
\end{figure}

During the Factual Knowledge Retrieval task (See \cref{fig:case_retrieval_fkr}), the models fail to isolate the specific dense infographic required to answer the query about "DMGT", instead returning structurally similar but irrelevant charts or slides. In the corresponding VQA example (See \cref{fig:case_vqa_fkr}), Gemini-2.5-Flash and InternVL-3 incorrectly identify the creator of the 2014 Winter Olympics poster, struggling to associate the correct attribution text within the complex graphic.

\FloatBarrier

\subsection{Metadata Identification (MI)}

\begin{figure}[!ht]
    \centering
    \includegraphics[width=0.8\linewidth]{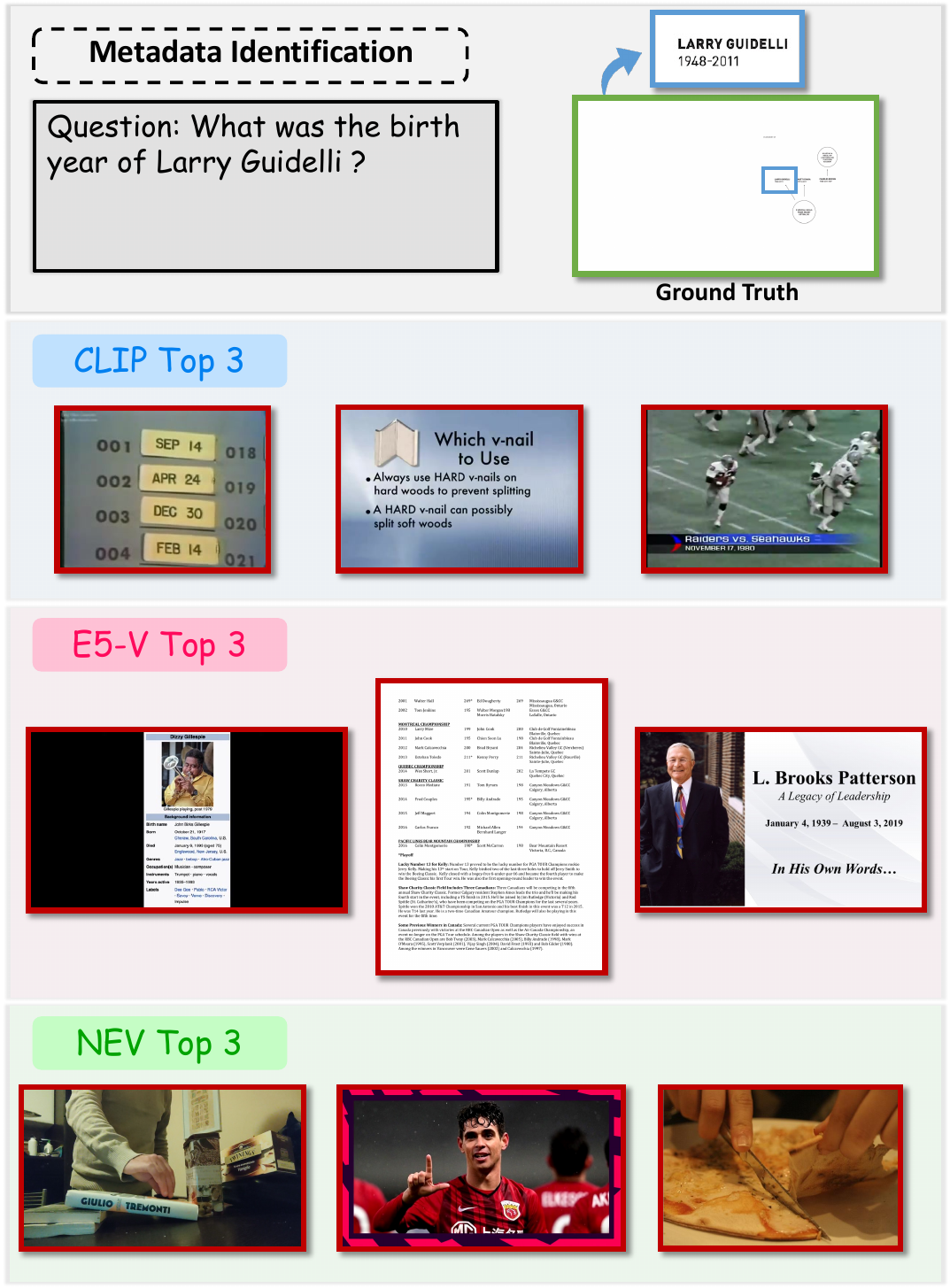}
    \caption{\textbf{Metadata Identification Retrieval.}}
    \label{fig:case_retrieval_mi}
\end{figure}

\begin{figure}[!ht]
    \centering
    \includegraphics[width=0.8\linewidth]{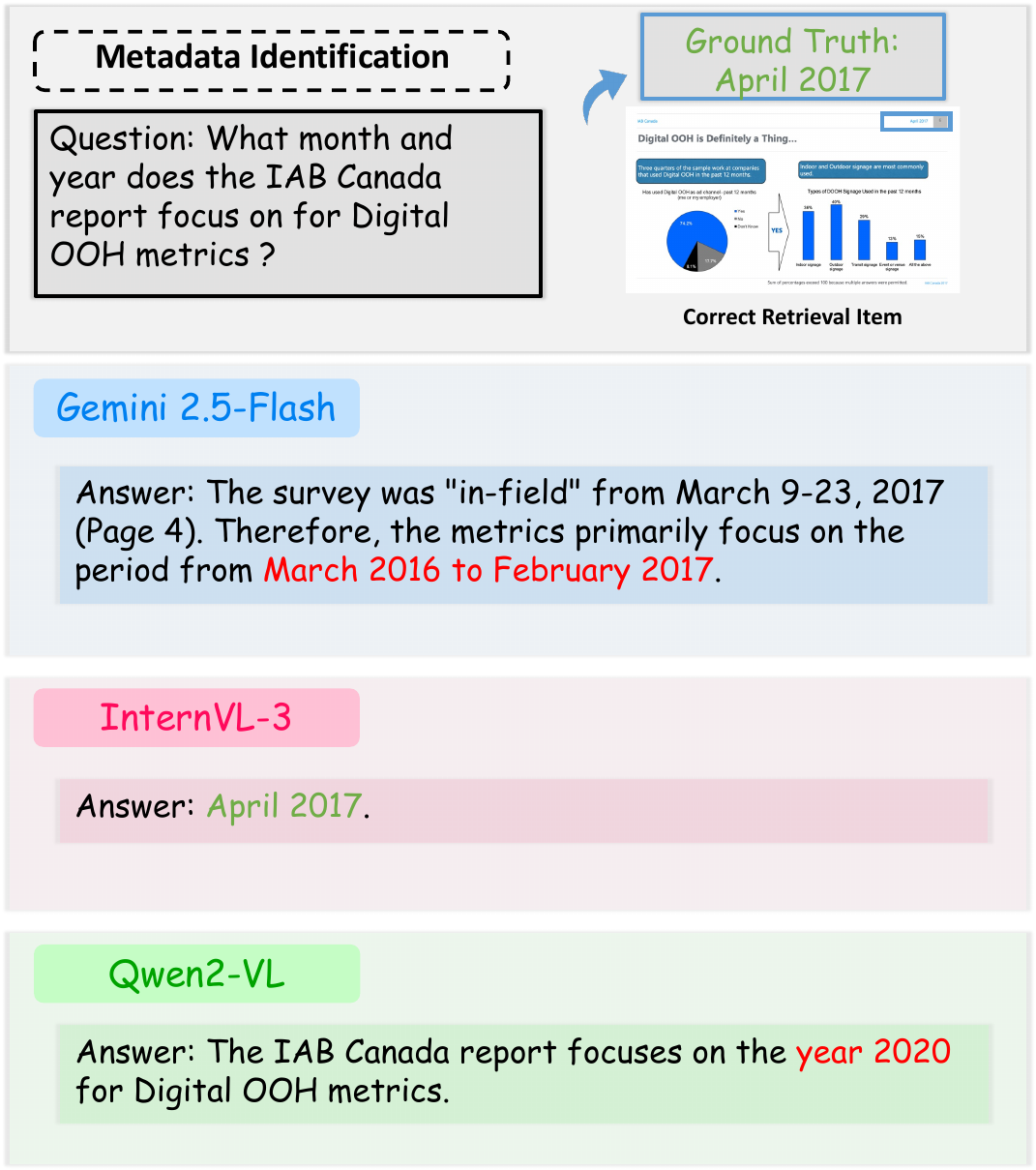}
    \caption{\textbf{Metadata Identification VQA.}}
    \label{fig:case_vqa_mi}
\end{figure}

For Metadata Identification retrieval (See \cref{fig:case_retrieval_mi}), the models miss the correct slide containing the subject's birth year, retrieving loosely related images that happen to feature dates or chronological formats. In the VQA case (See \cref{fig:case_vqa_mi}), while InternVL-3 successfully extracts the correct "April 2017" date from the document header, the other models output incorrect or adjacent date ranges found elsewhere in the text.

\newpage

\subsection{Statistical Reasoning (SR)}

\begin{figure}[!ht]
    \centering
    \includegraphics[width=0.8\linewidth]{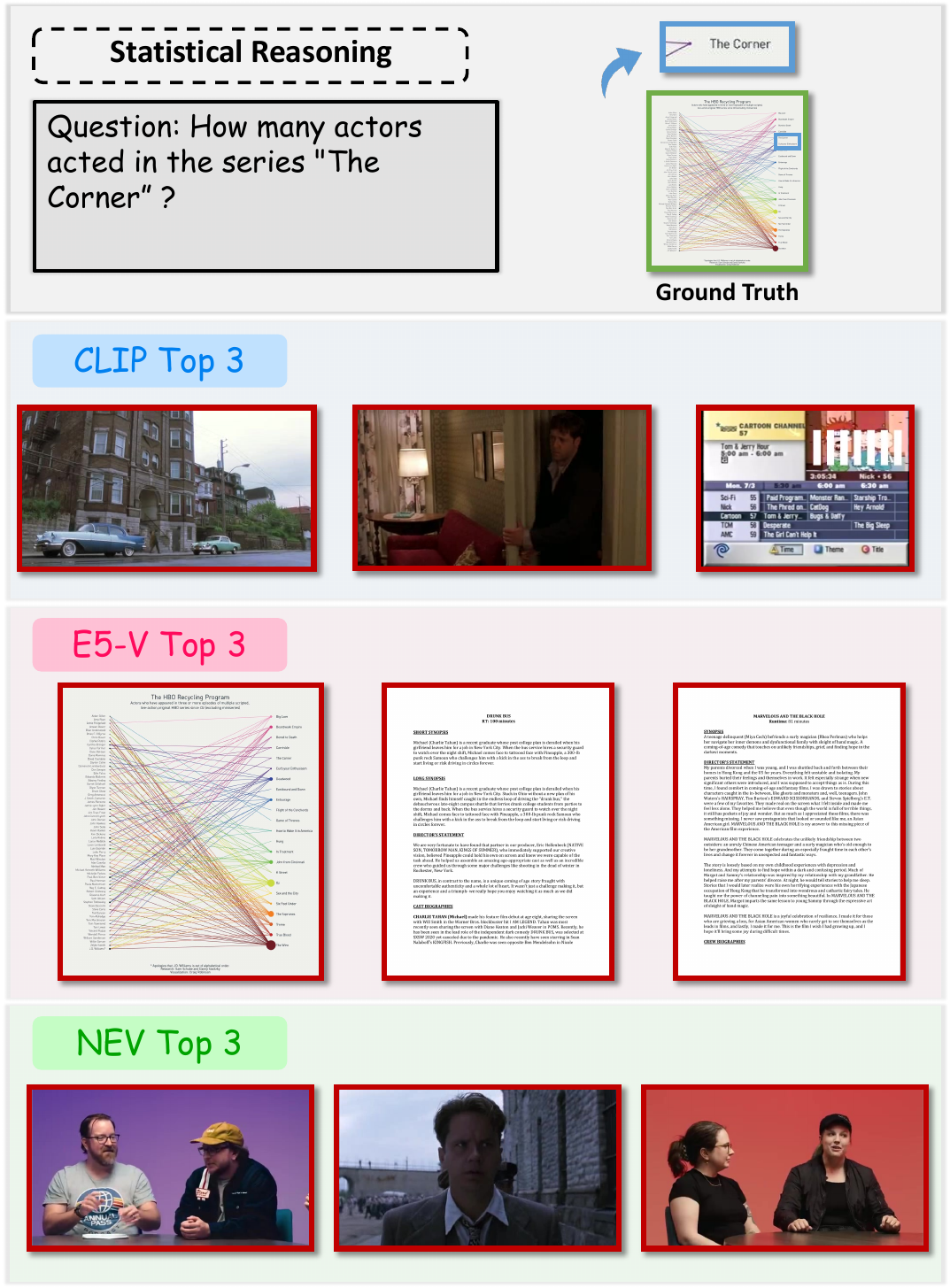}
    \caption{\textbf{Statistical Reasoning Retrieval.}}
    \label{fig:case_retrieval_sr}
\end{figure}

\begin{figure}[!ht]
    \centering
    \includegraphics[width=0.8\linewidth]{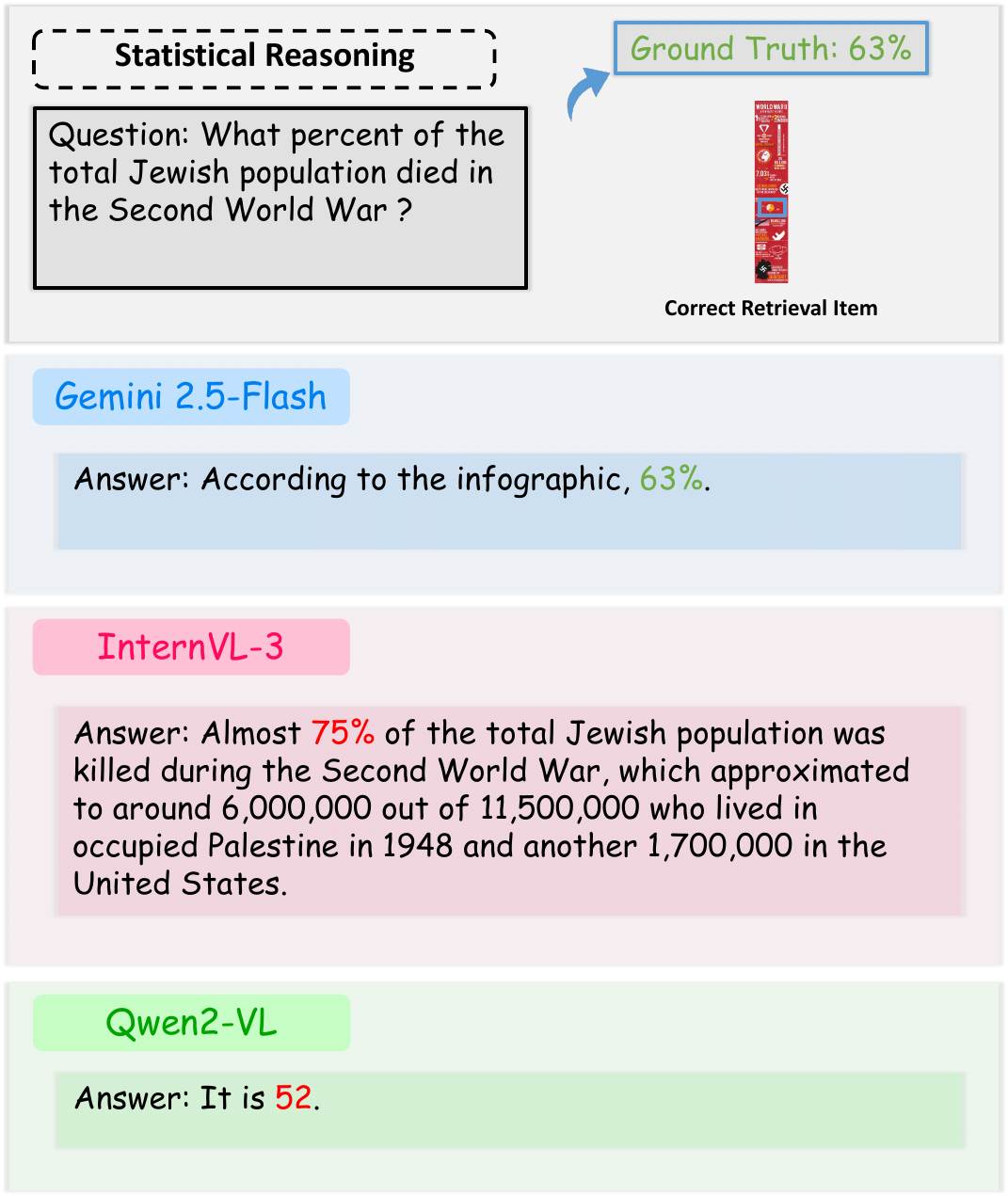}
    \caption{\textbf{Statistical Reasoning VQA.}}
    \label{fig:case_vqa_sr}
\end{figure}

In Statistical Reasoning, as shown in \cref{fig:case_retrieval_sr}, CLIP and NEV return generic text documents or unrelated video frames rather than the specific node-link chart required to count actor appearances. Similarly, in the VQA task (See \cref{fig:case_vqa_sr}), InternVL-3 and Qwen2-VL fail to correctly read the "63\%" statistic from the provided historical infographic, hallucinating different percentages.

\FloatBarrier

\subsection{Video Temporal Reasoning (VTR)}

\begin{figure}[!ht]
    \centering
    \includegraphics[width=0.8\linewidth]{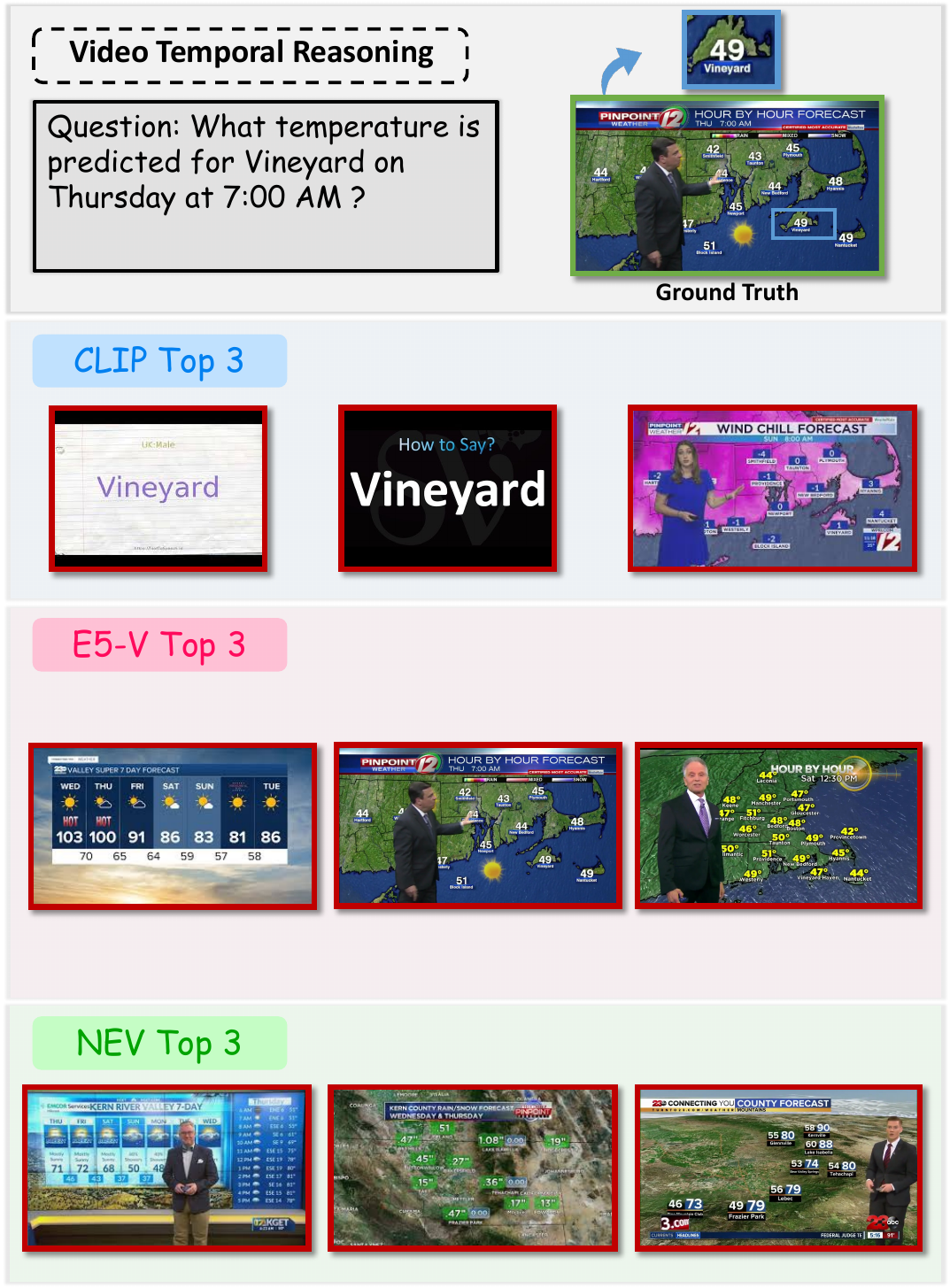}
    \caption{\textbf{Video Temporal Reasoning Retrieval.}}
    \label{fig:case_retrieval_vtr}
\end{figure}

\begin{figure}[!ht]
    \centering
    \includegraphics[width=0.8\linewidth]{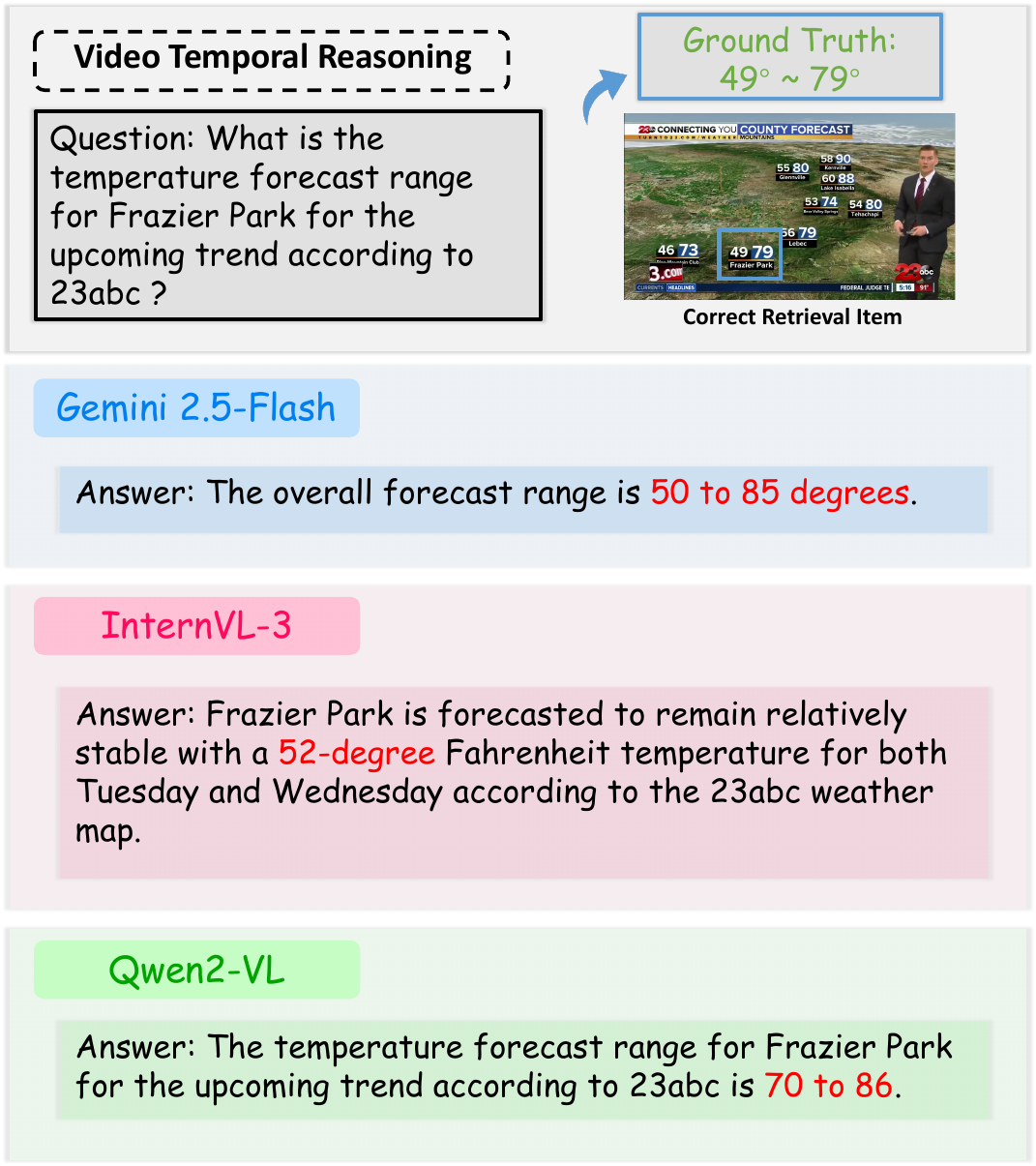}
    \caption{\textbf{Video Temporal Reasoning VQA.}}
    \label{fig:case_vqa_vtr}
\end{figure}

For Video Temporal Reasoning, as shown in \cref{fig:case_retrieval_vtr}, CLIP and NEV struggle to retrieve the exact weather forecast frame corresponding to a specific time and location, frequently defaulting to generic weather maps or text slides containing the keyword. In the VQA task (See \cref{fig:case_vqa_vtr}), all three models misread or hallucinate the temperature range for Frazier Park from the correct forecast frame, failing to output the highlighted "49° $\sim$ 79°".

\newpage

\subsection{Visual Parsing and Positioning (VPP)}

\begin{figure}[!ht]
    \centering
    \includegraphics[width=0.8\linewidth]{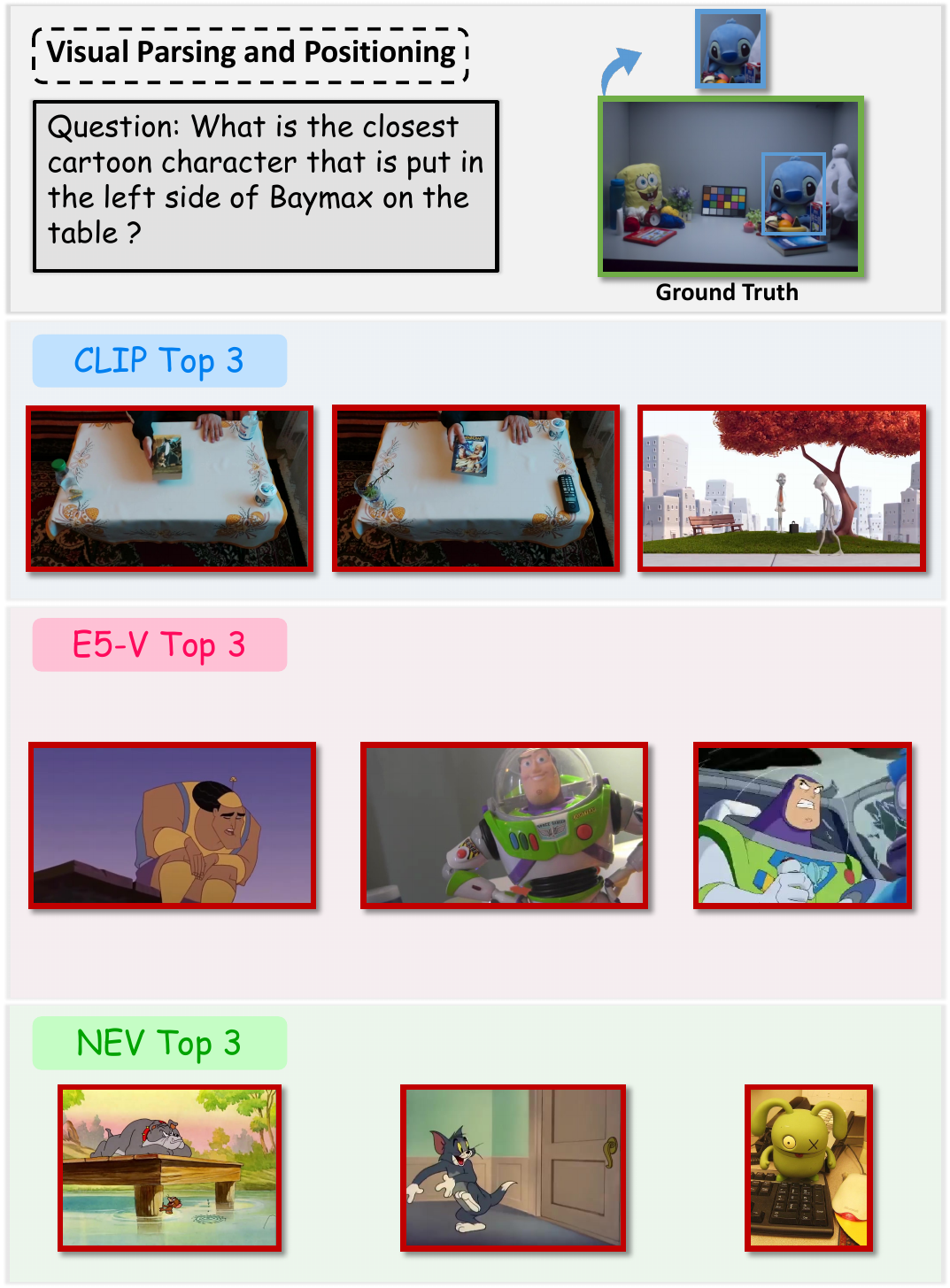}
    \caption{\textbf{Visual Parsing and Positioning Retrieval.}}
    \label{fig:case_retrieval_vpp}
\end{figure}

\begin{figure}[!ht]
    \centering
    \includegraphics[width=0.8\linewidth]{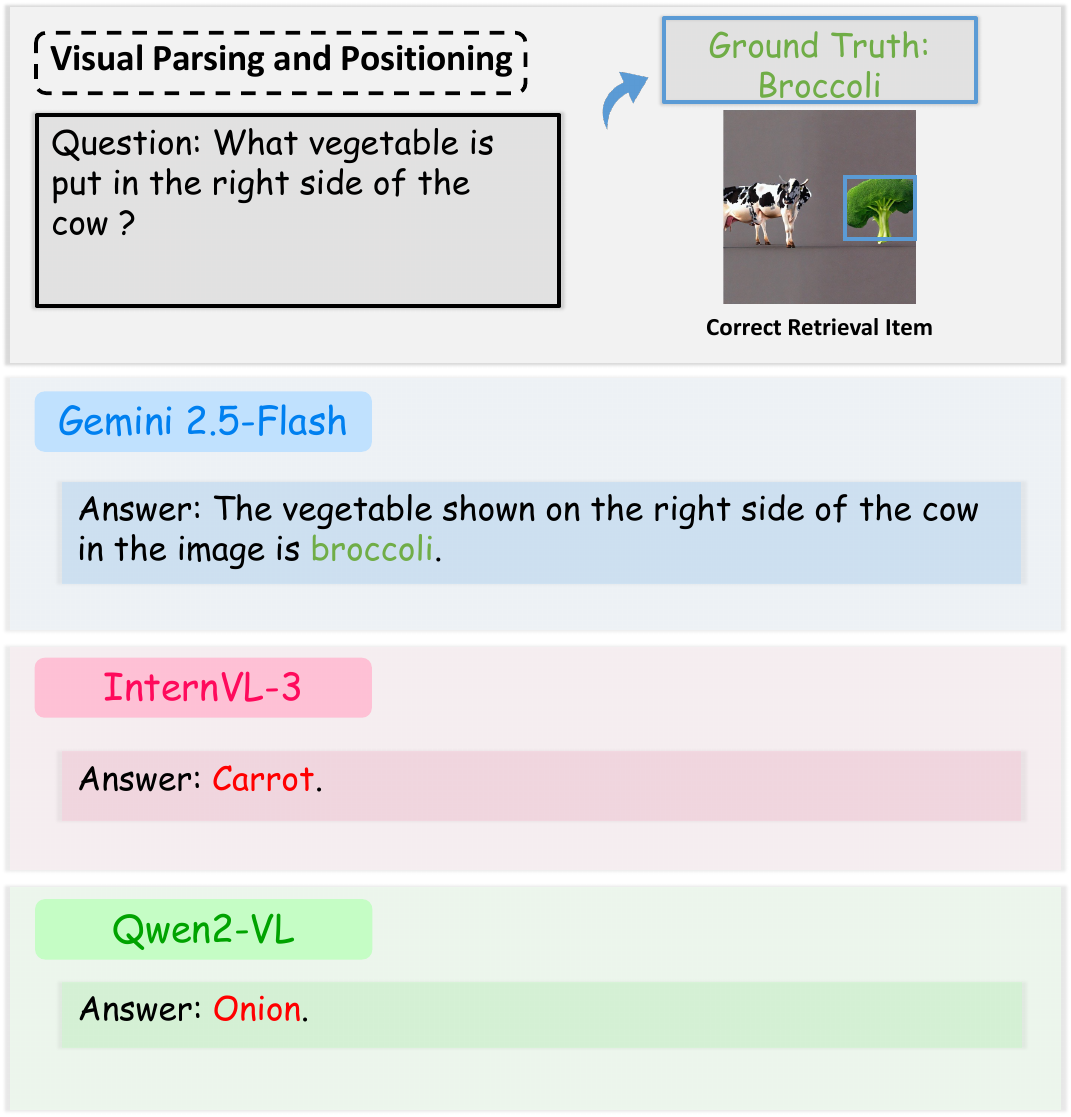}
    \caption{\textbf{Visual Parsing and Positioning VQA.}}
    \label{fig:case_vqa_vpp}
\end{figure}

In the Visual Parsing and Positioning retrieval task (See \cref{fig:case_retrieval_vpp}), all models struggle to comprehend specific spatial relationships between multiple objects, returning generic cartoon characters or unrelated tabletop scenes instead of the precise spatial configuration requested. For the VQA task (See \cref{fig:case_vqa_vpp}), while Gemini successfully identifies the "broccoli" positioned to the right of the cow, the other evaluated models fail to process the visual evidence and hallucinate incorrect vegetables entirely, such as a carrot or an onion.

\FloatBarrier

\section{Artifacts and licenses}

We report a list of licenses for all datasets and models used in our experiment in \cref{tab:license}.
We strictly follow all the model licenses and limit the scope of these models to academic research only. 

\begin{table}[ht]
\centering
\caption{License information for the scientific artifacts.} 
\label{tab:license}
    \scalebox{0.9}{
    \begingroup
    \setlength{\tabcolsep}{3pt}
    \hspace*{-6pt}
    \begin{tabular}{lll}
        \toprule
        \textbf{Data Sources}  & \textbf{URL} & \textbf{License}   \\ 
        \cmidrule(lr){1-1} \cmidrule(lr){2-3}
         VideoVista  &  \href{https://huggingface.co/datasets/Uni-MoE/VideoVista}{Link}  & Apache-2.0 \\
         MMBench-Video   & \href{https://huggingface.co/datasets/opencompass/MMBench-Video}{Link} & CC BY 4.0 \\
         FineVideo & \href{https://huggingface.co/datasets/HuggingFaceFV/finevideo}{Link} & CC BY 4.0 \\
         MVBench & \href{https://huggingface.co/datasets/OpenGVLab/MVBench}{Link} & MIT \\
         DocHaystack & \href{https://huggingface.co/datasets/DanielXu0208/Document_Haystacks}{Link} & MIT \\
         MMIU & \href{https://huggingface.co/datasets/FanqingM/MMIU-Benchmark}{Link} & CC BY 4.0 \\
         A-OKVQA & \href{https://huggingface.co/datasets/HuggingFaceM4/A-OKVQA}{Link} & Apache-2.0 \\
         MINT1T & \href{https://huggingface.co/datasets/mlfoundations/MINT-1T-PDF-CC-2024-18}{Link} & CC BY 4.0 \\
        \midrule
        \textbf{Software Code / Models} & \textbf{URL} & \textbf{License}   \\
        CLIP & \href{https://github.com/openai/CLIP}{Link} & MIT \\
        SigLIP2 & \href{https://huggingface.co/blog/siglip2}{Link} & Apache-2.0 \\
        OpenCLIP & \href{https://github.com/mlfoundations/open_clip}{Link} & MIT \\
        Jina-CLIP-V1 & \href{https://huggingface.co/jinaai/jina-clip-v1}{Link} & Apache-2.0 \\
        Jina-CLIP-V2 & \href{https://huggingface.co/jinaai/jina-clip-v2}{Link} & CC-BY-NC-4.0 \\
        NEV & \href{https://huggingface.co/nomic-ai/nomic-embed-vision-v1.5}{Link} & Apache-2.0 \\
        E5-V & \href{https://huggingface.co/royokong/e5-v}{Link} & Apache-2.0 \\
        MM-Embed & \href{https://huggingface.co/nvidia/MM-Embed}{Link} & CC-BY-NC-4.0 \\
        \cmidrule(lr){1-1} \cmidrule(lr){2-3}
        Ola & \href{https://github.com/Ola-Omni/Ola}{Link} &  Apache-2.0 \\
        Qwen2-VL &  \href{https://huggingface.co/Qwen/Qwen2-VL-7B-Instruct}{Link} & Apache-2.0 \\
        InternVL-3 &  \href{https://github.com/OpenGVLab/InternVL}{Link} & Apache-2.0 \\
        Gemini-2.5-Flash  & \href{https://deepmind.google/technologies/gemini/flash/}{Link} & \href{https://ai.google.dev/terms}{Google Terms of Use} \\
        GPT-5/4o-mini &  \href{https://openai.com/chatgpt}{Link} &  \href{https://openai.com/policies/terms-of-use/}{OpenAI Terms of Use} \\
        \bottomrule
    \end{tabular}
    \endgroup}
\end{table}

\paragraph{Practical usability.} 
In addition to licensing, we emphasize several practical aspects of dataset usability. 
First, the benchmark involves over 40K multimodal files (videos, images, and documents), which requires significant storage (on the order of terabytes) and compute resources for full-scale evaluation. 
Second, while all datasets are publicly hosted on Hugging Face under open licenses (Apache, MIT, CC BY), certain redistribution restrictions (e.g., CC-BY-NC) limit commercial use. 
Third, video corpora may present bandwidth challenges, and we recommend that academic users to selectively download subsets for targeted experiments. 
Finally, to ensure long-term accessibility, we will maintain mirrors for all datasets and scripts, together with versioned releases to facilitate reproducibility. 
These considerations ensure that our benchmark is both legally compliant and practically usable by the research community.

\section{Limitations and Future Work}
\label{sec:limitation}

Our study has several limitations. 
First, while MultiHaystack integrates text, images, and videos, it does not yet cover modalities such as audio or sensor signals. Extending to these would increase realism but also introduce challenges like temporal alignment and redundancy modeling.  
Second, benchmark construction relies on semi-automatic question generation and human verification. Although we enforce unique ground truths, annotation noise or bias may remain. Moreover, while GPT-4o is the backbone for both data construction and evaluation, we mitigate potential bias through multi-stage filtering, human checks, and consistency validation against human judgments, substantially reducing dependence on a single model. Future work could explore more scalable and diverse verification pipelines.  
Finally, current results are bounded by retriever quality: poor recall limits downstream reasoning regardless of model ability. Exploring retrieval-augmented training, adaptive candidate selection, or hybrid retrieval strategies may help overcome this bottleneck.

\section{Broader Impact}
\label{sec:impact}

By providing a large-scale multimodal benchmark, MultiHaystack can accelerate research on retrieval-augmented reasoning, enabling applications in search, education, healthcare, and scientific discovery. Improved systems may broaden access to complex multimodal information and support more reliable decision-making.  
At the same time, stronger retrieval and reasoning also raise risks, such as exposing sensitive information or amplifying misinformation. While our benchmark itself does not contain harmful content, responsible use of models evaluated on it requires privacy safeguards, robust verification, and appropriate policy frameworks.  
We hope MultiHaystack will guide both technical progress and responsible discourse on the societal impact of multimodal AI.

\section{LLM USAGE STATEMENT}
During dataset construction, we leveraged large language models as an auxiliary tool to suggest candidate question--answer pairs and to aid preliminary filtering. These outputs were then subjected to rigorous multi-stage manual verification to ensure both accuracy and diversity. For evaluation, we employed an automatic judging protocol, where an LLM was used to assist in assessing the correctness of answers from multiple VQA models. To validate the robustness of this approach, we performed direct comparisons against independent human annotations and confirmed high consistency. Finally, the manuscript underwent multiple rounds of refinement, combining careful manual revision with selective automated editing support to further improve clarity, coherence, and readability.

\end{document}